\documentclass[twoside,11pt]{article}
\usepackage{jair, theapa, rawfonts}

\usepackage{verbatim}
\usepackage{amsmath,amssymb,amsthm}
\usepackage{color, colortbl}
\usepackage[all,knot]{xy}
\usepackage{epsf,tikz}
\usepackage{enumitem}
\usepackage{multirow}

\newcommand{\del}[1]{} 

\newcommand{\todo}[1]{{
	}} 
\newcommand{\FAN}[1]{\textcolor{brown}{#1}}

\definecolor{Gray}{gray}{0.9}

\newcommand*{\DashedArrow}[1][]{\mathbin{\tikz [baseline=-0.25ex,-latex, dashed,#1] \draw [#1] (0pt,0.5ex) -- (1.3em,0.5ex);}}%

\def\flat{lean}
\def\Flat{Lean}
\def\dialogical{argumentative}
\def\Dialogical{Argumentative}
\def\dSpace{\ensuremath{\mathbb{S}}}
\def\oSpace{\ensuremath{\mathbb{Y}}}
\def\ofunc{\ensuremath{\mathbf{f}}}
\def\fSln{\mc{X}}
\def\yImg{\mc{Y}}

\def\LB{\ensuremath{\mc{L}_{\info}}}

\def\sent{\ensuremath{\beta}}

\def\sentset{\ensuremath{S}}

\newcommand{\lracFiLiG}[1]{\ensuremath{\langle\mathcal{L}^g_{#1},\mathcal{R}^g_{#1},\mathcal{A}^g_{#1},\mathcal{C}^g_{#1}\rangle}} 
 
\def\asmset{\ensuremath{\Delta}}
\def\asm{\ensuremath{\alpha}}

\def\argA{\ensuremath{{\tt A}}}
\def\argB{\ensuremath{{\tt B}}}
\def\argAs{\ensuremath{\tt As}}
\def\argBs{\ensuremath{\tt Bs}}

\newcommand{\atree}{\ensuremath{\mc{T}^a}}

\newcommand{\mtree}{\ensuremath{\mc{T}^{max}}}

\newcommand{\anode}[2]{\ensuremath{{#1}:{#2}}}

\def\node{\ensuremath{{\tt N}}}
\def\dNode{\ensuremath{\decisions}}
\def\iNode{\ensuremath{\node_{int}}}
\def\gNode{\ensuremath{\goals}}

\def\unreach{\ensuremath{\mu}}

\def\edge{\ensuremath{\tt E}}
\def\dEdge{\ensuremath{\edge_d}}
\def\sEdge{\ensuremath{\edge_s}}

\def\info{\ensuremath{{\tt B}}}

\def\dgPair{\ensuremath{\langle \node,\edge \rangle}}
\def\edgTuple{\ensuremath{\langle \node, \edge, \info \rangle}}
\def\edgTupleEmpty{\ensuremath{\langle \node, \edge, \emptyset \rangle}}
\def\edgTupleEmptyStrict{\ensuremath{\langle \node, \sEdge, \emptyset \rangle}}

\newcommand{\egPair}[2]{[#1 \rhd #2]} 
\newcommand{\myTag}[1]{t(#1)}

\newcommand{\wrt}{with respect to}
\newcommand{\ifaf}{if and only if}
\newcommand{\respectively}{respectively}
\newcommand{\st}{such that}

\newcommand{\bSet}{defeasible condition} 

\newcommand{\lracF}{\ensuremath{\langle\mathcal{L}, \mathcal{R}, 
 \mathcal{A}, \mathcal{C}\rangle}}

\newcommand{\lracFSD}{\ensuremath{\langle \mc{L}_{s}, \mc{R}_{s},
    \mc{A}_{s}, \mc{C}_{s} \rangle}}

\newcommand{\lracFD}{\ensuremath{\langle \mc{L}_{d}, \mc{R}_{d},
    \mc{A}_{d}, \mc{C}_{d} \rangle}}

\newcommand{\lracFWD}{\ensuremath{\langle \mc{L}_{w}, \mc{R}_{w},
    \mc{A}_{w}, \mc{C}_{w} \rangle}}

\newcommand{\lracFPG}{\ensuremath{\langle \mc{L}_{p}, \mc{R}_{p},
    \mc{A}_{p}, \mc{C}_{p} \rangle}}

\newcommand{\lracFcore}{\ensuremath{\langle\mathcal{L}_0,
    \mathcal{R}_0, \mathcal{A}_0, \mathcal{C}_0\rangle}} 

\newcommand{\lracFcoreG}{\ensuremath{\langle \LcoreG, \RcoreG,
    \AcoreG, \CcoreG \rangle}} 

\newcommand{\lracFcoreGD}{\ensuremath{\langle \LcoreGD, \RcoreGD,
    \AcoreGD, \CcoreGD \rangle}} 

\newcommand{\LcoreG}{\ensuremath{\mc{L}_g}}
\newcommand{\RcoreG}{\ensuremath{\mc{R}_g}}
\newcommand{\AcoreG}{\ensuremath{\mc{A}_g}}
\newcommand{\CcoreG}{\ensuremath{\mc{C}_g}}

\newcommand{\LcoreGD}{\ensuremath{\mc{L}_e}}
\newcommand{\RcoreGD}{\ensuremath{\mc{R}_e}}
\newcommand{\AcoreGD}{\ensuremath{\mc{A}_e}}
\newcommand{\CcoreGD}{\ensuremath{\mc{C}_e}}

\newcommand{\Rcore}{\ensuremath{\mc{R}_0}}
\newcommand{\Acore}{\ensuremath{\mc{A}_0}}
\newcommand{\Ccore}{\ensuremath{\mc{C}_0}}

\def\notPS{\ensuremath{notGP}}
\def\notMetSet{\ensuremath{notMetS}}
\def\metSet{\ensuremath{metS}}
\def\pS{\ensuremath{gP}}

\newcommand{\la}{\ensuremath{\leftarrow}}
\newcommand{\mc}[1]{\ensuremath{\mathcal{#1}}}
\newcommand{\mb}[1]{\ensuremath{\mathbf{#1}}}

\newcommand\pro{\ensuremath{\mathcal{\bf{P}}}}
\newcommand\opp{\ensuremath{\mathcal{\bf{O}}}}

\def\df{\ensuremath{F}}
\def\fa{\ensuremath{F_a}}
\def\fe{\ensuremath{F_e}}

\def\cToD{\ensuremath{\mb{M_1}}}

\def\dg{\ensuremath{G}}

\def\gp{\ensuremath{G_p}}

\newcommand{\argu}[2]{#1 \vdash #2} 
\def\arg{{\tt Arg}}

\newcommand{\Tdgi}[2]{\ensuremath{{\tt T_{DG}}[#1,#2]}}

\def\decisions{{\tt D}}

\def\cgset{{\tt S}}

\def\goals{{\tt G}}
\def\preferences{{\tt P}}

\def\preferenceSet{{\tt 
\leq_g}}
\def\notpreferenceSet{{\tt 
{\not\leq}_g}}
\def\preferenceSetstrict{{\tt 
<_g}}

\def\Tdg{\ensuremath{{\tt T_{DG}}}}

\newcommand{\nadf}{\ensuremath{\langle \decisions, \goals, \dmg \rangle}}
\newcommand{\adf}{\ensuremath{\langle \decisions, \goals, \Tdg \rangle}}
\newcommand{\paretoFramework}{\ensuremath{\langle \dSpace, \oSpace,
    \ofunc, \prec \rangle}}
 
\newcommand{\neadfps}{\ensuremath{\langle \decisions, \goals, \dmg, \preferenceSet \rangle}}
\newcommand{\eadfps}{\ensuremath{\langle \decisions, \goals, \Tdg, \preferenceSet \rangle}}

\def\CNJalternatives{\ensuremath{{\tt L}}}
\def\CNJrequirements{\ensuremath{{\tt R}}}
\def\CNJattributes{\ensuremath{{\tt I}}}
\def\boolFunc{\ensuremath{
	\mathcal{B}}}

\def\LexAlternatives{\ensuremath{{\tt A}}}
\def\LexAttributes{\ensuremath{{\tt X}}}

\newcommand{\cnjFmk}{\ensuremath{\langle \CNJalternatives,
    \CNJrequirements, \boolFunc \rangle}}

\def\dmg{\ensuremath{\gamma}}

\def\pfrGoal{g-preferred}

\def\pfrSet{
	g-preferred}
\def\PfrGoal{G-Preferred}
\def\PfrSet{
	G-Preferred}

\def\leastAsm{least-assumption}
\def\LeastAsm{Least-assumption}

\def\PDG{Preferential Decision Graph}
\def\pdgTuple{\ensuremath{\langle \node, \edge, \info, \preferenceSet \rangle}}

\newcommand{\expPair}[2]{\ensuremath{(#1,#2)}}

\newtheorem{preex}{Example}[section]
\newenvironment{example}
{\begin{preex} \begin{rm}}
{\end{rm} \end{preex}}

\newtheorem{predfn}{Definition}[section]
\newenvironment{definition}
{\begin{predfn} \begin{rm}}
{\end{rm} \end{predfn}}

\newtheorem{prethm}{Theorem}[section]
\newenvironment{theorem}
{\begin{prethm} \begin{rm}}
{\end{rm} \end{prethm}}

\newtheorem{precol}{Corollary}[section]

\newtheorem{prelem}{Lemma}[section]
\newenvironment{lemma}
{\begin{prelem} \begin{rm}}
{\end{rm} \end{prelem}}

\newtheorem{prepro}{Proposition}[section]
\newenvironment{proposition}
{\begin{prepro} \begin{rm}}
{\end{rm} \end{prepro}}

\newtheorem{prenot}{Notation}[section]

\ShortHeadings{Explainable Decision Making with 
\Flat\ and \Dialogical\ explanations 
}
{Fan and Toni}

\begin{document}

\title{
	Explainable Decision Making with \\
	\Flat\ and \Dialogical\ Explanations 
	}

\author{\name Xiuyi Fan \email xyfan@ntu.edu.sg \\
\addr School of Computer Science and Engineering, Nanyang Technological University, Singapore \\
        \name Francesca Toni \email ft@imperial.ac.uk \\
        \addr Department of Computing, Imperial College London, 
UK}
\maketitle

\begin{abstract}
It is widely acknowledged that transparency of automated decision
making is crucial for deployability of intelligent systems, and
explaining the reasons why some decisions are ``good'' and some are
not is a way to achieving this transparency. We consider two variants
of decision making, where ``good'' decisions amount to alternatives
(i) meeting ``most'' goals, and
(ii) meeting ``most preferred'' goals.
We then define, for each variant and notion of ``goodness''
(corresponding to a number of existing notions in the literature),
explanations in two formats, for justifying the selection of an
alternative to audiences with differing needs and competences: 
\emph{\flat} explanations, in terms of goals satisfied and, for some
notions of ``goodness'', alternative decisions, and
\emph{\dialogical} explanations, reflecting the decision process
leading to the selection, while corresponding to the
\flat\ explanations. To define \dialogical\ explanations, we use {\em
  assumption-based argumentation} (ABA), a well-known form of 
structured argumentation. Specifically, we define ABA frameworks  
\st{} ``good'' decisions are admissible ABA arguments and draw
\dialogical\ explanations from {\em dispute trees} sanctioning this
admissibility. Finally, we instantiate our overall framework for
explainable decision-making to accommodate connections between goals
and decisions in terms of {\em decision graphs} incorporating
defeasible and non-defeasible information.
\end{abstract}

\newpage
\tableofcontents{}

\newpage
\section{Introduction}
\label{sec:intro}

Decision making can be understood as a process of selecting ``good''
{\em decisions} amongst several alternatives, e.g. using {\em decision
  criteria}, based on relevant information available to the decision
maker and a representation of this information. For systems assisting
decision making process, it is important that these systems not only
generate ``good'' selections, but be able to explain their
selections to their users. Indeed, it is widely acknowledged that
transparency of automated decision making is crucial and explaining
the reasons why some decisions are ``good'' and some are not is a way
to achieving this transparency \cite{right2x}. It is less clear
however what these explanations may be but, given that the decisions
may need to be understood by different kinds of users, different forms
of explanations are desirable, tailored to the needs and abilities of
their users \cite{Sokol20,Sheh17,Phillips2020}.

In this paper we consider two different types of explanations for several forms of decision-making, based on different decision criteria. The first type of explanation is targeted at users who do not need or do not care to understand the mechanisms for selecting a decision as ``good'' (according to a chosen decision criterion): these explanations thus simply refer to the inputs for the decision-making process, namely the goals satisfied by the ``good'' decisions, and, in some settings, preferences over these goals and alternative decisions. We refer to  explanations of this first type as \emph{\flat}, 
given that they focus solely on the building-blocks of decision making and ignore the intricacies of the underlying decision-making criterion sanctioning the decision as ``good''. The second type of explanation is targeted at users who need and want to understand the underpinning mechanisms: these explanations bring up the reasoning behind the identification of the explanations. We refer to  explanations of this second type as \emph{\dialogical}, because we use \emph{dispute trees} \cite{Dung06b,Dung07,aij12-XDDs}, as understood in \emph{computational argumentation} (see e.g. \cite{Bench-Capon07,Besnard08,Rahwan09,arg-survey-recent}), to represent the reasoning leading to the identification of the explained decisions as ``good''. 
The two forms of explanations are related, as we show that \flat\ explanations can be obtained from \dialogical\ ones.

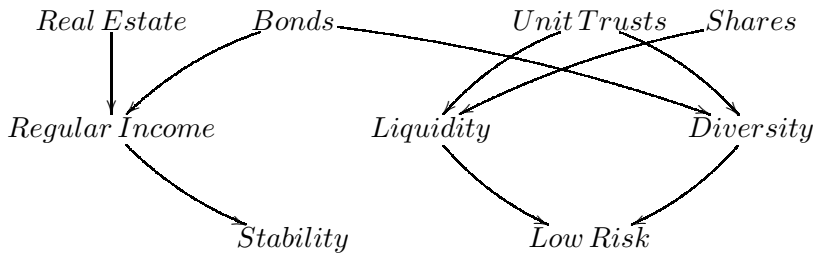
\begin{figure}[h]
\centerline{
  \xymatrix@1@=6pt@R=30pt{
    Real \, Estate \ar[d]    & Bonds \ar@/_/[ld] \ar@/^/[rrrd]      & & Unit \, Trusts\ar@/_/[ld] \ar@/^/[rd] & Shares \ar@/_/[lld]\\
    Regular \, Income \ar@/_/[rd]&           & Liquidity \ar@/_/[rd] & & Diversity \ar@/^/[ld] \\
                          & Stability &                  & Low \, Risk \\
}
}
\caption{Investment Product Example.\label{fig:introExp}}
\end{figure}

For illustration, consider the following scenario. A financial advisor
  is to make investment suggestions to their client. The decision
  alternatives are {\em Shares}, {\em Bonds}, {\em Unit Trusts} and
  {\em Real Estate}; these have attributes {\em Regular Income}, {\em
    Liquidity}, and {\em Diversity}; and the goals in consideration
  are {\em Stability} and {\em Low Risk}. The relations amongst these
  are depicted in Figure~\ref{fig:introExp} as a \emph{decision graph (DG)}. To explain why {\em Bonds}
  is a good investment choice, the advisor can either give the reason 
  ``{\em Bonds} meets goals {\em Stability} and {\em Low Risk}'', or go the long way saying something along the lines ``{\em
    Bonds} brings \emph{Regular Income}, so it brings \emph{Stability}; it also has good
  \emph{
	  Diversity} so it has \emph{Low Risk}''. Both explanations can be seen as \flat, as they refer exclusively to the building blocks of the decision problem (how the decisions fulfil the goals in the first case and how they bring intermediate attributes in turn determining the goals). However, neither explanations give any indications about the reasoning of the decision maker towards sanctioning the decisions as ``good''. An \dialogical\ explanation in this toy setting, unearthing the underpinning reasoning, may, for example, amount to a dispute between fictional proponent and opponent players whereby the proponent starts by arguing for the ``goodness'' of {\em Bonds}, followed by the opponent questioning its satisfaction of goals {\em Stability} and {\em Low Risk}, and ending with the proponent proving (by reasoning on the DG) that both goals are satisfied, which the opponent cannot object against. The underpinning \dialogical\ reasoning here is straightforward (given that the decision problem is simple and one alternative decision is obviously ``best'', meeting as it does both goals). For other settings, though, where more sophisticated decision criteria may be required to discriminate amongst alternatives, \dialogical\ explanations introduce more transparency into decision making. 

\Flat\ explanations are, in spirit, similar to model-agnostic explanations for machine learning models (e.g. the popular LIME \cite{LIME} and SHAP \cite{SHAP}) and model-specific explanations  (e.g. counterfactual explanations as in \cite{CF-silvestri,CF-wachter}): like  these existing explanation methods, our \flat\ explanations 
focus on explaining the outputs of AI systems solely in terms of their inputs.
Instead, \dialogical\ explanations are, in spirit, similar to work on visualisation of deep learning models \cite{Olah18} or explanations of their inner workings \cite{argflow,purin}: like these existing explanation methods, our \dialogical\ explanations attempt to shed light on the underpinning workings of the AI system towards obtaining outputs from inputs.  
Our \dialogical\ explanations also share with \cite{argflow,purin} (as well as with several other works, as overviewed in \cite{Cyras21}) the choice of computational argumentation as the underpinning methodology, and strives towards explanations that are in line with recommendations from the social sciences (e.g. as overviewd in \cite{miller}) and in particular with interpretations of human-oriented explanations as argumentative (e.g. see \cite{argExpl,mercier}). 

We consider two variants of decision making.
The first variant considers {\em Abstract Decision Frameworks (ADFs)} 
consisting of {\em decisions}, {\em goals} and a {\em table} 
giving a {\em decisions-meeting-goals} relation. In the context of
ADFs, we study three decision criteria, {\em strong dominance} (the most basic, which we illustrated in the earlier toy example),
  \emph{dominance}, and {\em weak dominance}, \respectively{} enforcing, as
``good'', decisions meeting all goals, meeting all goals that are met
by any decision, and meeting goals not met by any other decision. 
The second variant considers {\em Preferential Decision
  Frameworks (PDFs)}, which add to ADFs
  {\em preferences} over goals. In the context of PDFs, we study the
decision criterion (that we call \pfrSet)
enforcing, as ``good'', decisions meeting the most preferred
goals. 
The decision criteria tht we consider are strongly connected to 
other decision making works in the literature. 
In particular, we show that strong dominance corresponds to selected decisions using
the conjunctive method in standard multi-attribute decision making
\cite{Yoon95}, weak dominance corresponds to the well-known
decision criterion of Pareto optimality in decision theory
\cite{Emmerich06} and \pfrSet{} generalizes the lexicographic method in 
multi-attribute decision making \cite{Yoon95}.
For each of the proposed decision criteria, we
give \flat\ and \dialogical\ explanation for a decision candidate,
either justifying the selection or identifying conditions that
the decision fails to meet
.  
 We also show that the \flat\ explanations that we define (of various kinds, corresponding to the various decision criteria and decision framework) can be directly obtained from (the corresponding) \dialogical\ explanations.

In order to provide \dialogical\ explanations, for each of the proposed decision making frameworks (ADFs and PDFs) and
respective decision criteria, we give mappings into
assumption-based argumentation (ABA) frameworks
\cite{Bondarenko93,Bondarenko97,Dung09,Toni14}, a well known form of computational argumentation with provably correct computational mechanisms (dispute trees)
\wrt{} several semantics~\cite{Dung06b,Dung07,aij12-XDDs}, from which \dialogical\ explanations can be readily extracted. Specifically, we prove that ``good''
decisions correspond to \emph{admissible arguments}, namely arguments
belonging to admissible sets (which are conflict-free and can defend
themselves against all attacks); admissible
arguments can be computed through the construction of {\em admissible
  abstract dispute trees}: \dialogical\
explanations are then drawn from variants of these trees. 

Finally, we consider an instance of our overall framework for explainable decision making, using DGs (generalising that illustrated in Figure~\ref{fig:introExp}) 
for capturing more complex information about the relations between
decisions and goals, in ADFs and PDFs. In a DG, a decision leads to {\em intermediate
  goals}, which in turn may lead to additional intermediate goals and
eventually to goals. With DGs, we have the ability to model {\em
  defeasibility} in decision making, by expressing, essentially, 
that ``a decision normally meets a goal, unless some conditions hold''. 
In this instantiation, the argumentative counterpart of decision making giving \dialogical\ explanations can also be used to support reasoning  leading to ``good'' decisions.

The paper is organised as follows. 
We position our paper in the context of related work in
Section~\ref{sec:relatedWork}.
In Section~\ref{sec:new} we define ADFs and PDFs with their decision criteria (and in Appendix~\ref{sec:relation} show relations with existing decision-making methods). In Section~\ref{sec:explanation} we define our \flat\ explanations and in Section~\ref{sec:dial} we define our \dialogical\ explanations, after providing some necessary background on ABA (in Section~\ref{sec:bkg}) 
and after introducing (in Section~\ref{sec:pre}) variants of the {\em abstract dispute
trees} used in this work.
We introduce DGs and explainable
decision making therein, with and without defeasible information, in
Section~\ref{sec:dg}. 
We  conclude in
Section~\ref{sec:conclusion}. 

This paper builds upon and generalises existing prior work as follows.
ADFs, PDFs and DGs introduced in this work are abstractions of the
decision frameworks, preferential decisions frameworks and DGs
introduced in \cite{Fan13-TAFA}, \cite{Fan13-CLIMA,Fan14-aamas} and
\cite{Carstens15}, \respectively. In this paper, we have expanded the theoretical presentation of these various decision making frameworks (including relations with existing decision making methods)
and, crucially, focused on formulations of explanation, absent from this prior work. 

\section{Related Work}
\label{sec:relatedWork}

Even though most of the recent work on explainable AI focuses on
  explaining black-box models, there is also significant effort in
  explaining transparent methods, especially when non-expert users
  need to make sense of the outputs of these
  methods~\cite{magazzeni}. Specifically, \cite{xplanning} provide a
  survey of existing work on explainable planning as a form of
  decision making: our \dialogical\ explanations can be seen as forms
  of their ``algorithm-based explanations''. Also,
  \cite{scheduling,scheduling-demo} focus on  scheduling, another form
  of decision making: like us, they developing argumentation-based
  explanations, but based on abstract argumentation \cite{Dung95}
  rather than ABA. 

Our proposal of two types of (\flat\ and \dialogical) explanations is in line with the recent advocacy of multiple forms of explanations.
Specifically, \cite{Sheh17}, 
in the context of Human-Robot Interaction (HRI) research, introduce an
  explanation categorization containing five types: {\em teaching,
    introspective tracing, introspective informative, post-hoc} and
  {\em execution} for the purpose of ``matching machine learning
  capabilities with HRI requirements''. In the context of explaining
machine learning predictions, \cite{Sokol20} have looked at explaining
prediction models globally with {\em model visualisation} and {\em
  feature importance} and prediction instances locally with {\em
  decision rule, counterfactual} and {\em exemplar}. They suggest that
personalised explanations, which present different forms of
explanations to different users, are beneficial for achieving
interpretable machine learning. Moreover, a recent National
Institute of Standards and Technology report published by the
U.S. Department of Commerce \cite{Phillips2020} reports that ``the
audience will strongly influence the purpose of the explanation'' and
``the explanation's purpose will influence its style''. 

Our \dialogical\ explanations leverage on the field of computational
argumentation, as overviewed, for example, in
\cite{Bench-Capon07,Besnard08,Rahwan09,Modgil13,arg-survey-recent}. Computational
argumentation in many forms and shapes is widely used in explainable
AI (e.g. see the recent survey by \cite{Cyras21}). Our approach uses
ABA \cite{Bondarenko93,Bondarenko97,Dung09,Toni14,Cyras17},  a
specific form of computational argumentation, well-suited
for our purposes for a number of reasons. First, it is a general
purpose argumentation framework 
suited in principle to any application
(e.g. see~\cite{Dung09,Toni12,Toni14}), that has already proven useful 
to support decision
making~\cite{Matt09,Dung08,Fan13-CLIMA,Fan13-TAFA,Fan14-aamas,Zeng18,Carstens15}. ABA has strong theoretical
foundations~\cite{Bondarenko93,Bondarenko97}, with several formal
results readily available characterising the computational complexity 
of various tasks in ABA~\cite{complexity:aij02,Dunne09-aba,new-compl}.
Another essential feature of ABA, for the purposes of this paper, is
that it is equipped with provably correct computational mechanisms
(dispute trees) \wrt{} several
semantics~\cite{Dung06b,Dung07,aij12-XDDs}, which includes {\em
  admissibility}. We rely upon aspects of these mechanisms, as well as
their soundness, to define our \dialogical\ explanations.
Finally, 
ABA is a framework for \emph{structured} argumentation~\cite{tutor},
allowing a fine-grained representation of the various components in a
decision model. At the same time, the specific form of ABA we need
(referred to in the literature as \emph{flat} \cite{Toni14}), is an
instance of abstract argumentation~\cite{Dung07,Toni12,Toni14} and it
admits abstract argumentation as an
instance~\cite{Toni12,Toni14}. Moreover, (flat) ABA is an 
instance of ASPIC+~\cite{Prakken10}, another framework for structured
argumentation. Thus, our ABA formulations could serve as a starting point for other explanation styles, different from our \dialogical\ explanations, built using these other related argumentation frameworks. 

%
%
ABA has already been studied in XAI in the literature. For instance,
  \cite{Fan15-AAAI} introduce {\em related admissibility} to identify
  arguments and assumptions that explain argument acceptability in 
ABA; \cite{Wakaki17} model ABA with preference and
experiment with decision making; \cite{Fan18,Fan18a} study how ABA can be used to generate
explainable plans; and \cite{Zeng20} use ABA for explainable
diagnostics and prognostics of Alzheimer's Disease.
To the best of our knowledge, though, our work is unique for the variety of decision problems it considers and in showing how \dialogical\ explanations relate to \flat\ explanations, of a more conventional spirit, as they are directly defined in terms of the basic components of the decision problems.

\cite{Cyras21} categorise work on explainable AI based on computational
argumentation according to the relation between the explanations and 
the method being explained and according to the format of
argumentation-based explanations in particular. In their context, our
work can be categorised as ``post-hoc complete'', in that we provide a
one-to-one mapping between decisions and \dialogical\ explanations while also, in addition, formally relating the latter and \flat\ explanations.

We define abstract decision problems (with and without preferences over goals) as well as a specific instance thereof integrating information in the form of decision gaphs (possibly integrating defeasible knowledge). This instance can be seen as a form of (explainable) argumentation-based decision making, in the sense that it can be used to directly support the identification of ``good'' decisions and their explanations. 
Argumentation-based decision making in general has attracted a considerable
amount of research interest over the years (e.g. see
\cite{Fox93,Nawwab08,Amgoud09,Fox10,Muller12}). Argumentation-based
decision making is a form of {\em qualitative decision theory}
\cite{Doyle99}, understood as an alternative to classical 
\emph{quantitative decision theory} \cite{French86} when a decision
problem cannot be easily formulated in standard decision-theoretic
terms using decision tables, utility functions and probability
distributions. In decision-theoretic terms, argumentation can be used
to compute a utility function which is too complex to be given a
simple analytic expression in closed form, and/or when a transparent
justification of decisions is beneficial.
Existing approaches to decision making using argumentation are
predominantly \emph{descriptive}, in that they are inspired by 
what people actually do and aim at justifying decisions by equipping
them with a transparent explanation. For example,
\cite{Atkinson04} focuses on structuring arguments according to
well-defined argument schemes in order to justify and explain
decisions. In this work, we focus on {\em normative} decision 
theory, regulated by decision criteria fulfilling properties of
rationality. In other words, we first define the type of decision
making problems we want to solve, the decision criteria for solving
them and the types of explanations that justify the selection, then
we use argumentation as an alternative way to identify the solutions
and generate explanations.

\begin{table}
  \begin{center}
    \caption{Works on Argumentation-based Decision
      Making. \label{table:related}} 
    \begin{small}
    \begin{tabular}{|c|ccc|}
      \hline
      \multirow{2}{*}{Works}& Argumentation  & \multirow{2}{*} {Preference} & Explanation \\ 
                            & Frameworks     &                              & Type \\ 
      \hline
      Matt et al. 2009             & ABA                 & No       & None \\ 
      Fox et al.  2007             & ASPIC               & No       & Arguments \\ 
      Marreiros et al. 2007        & Logic Programming   & Decision & None \\ 
      Visser et al. 2012           & Structured          & Decision &  None \\ 
      Dung et al., 2008            & ABA                 & Goals \& States & None \\
      Fan \& Toni, 2013            & ABA                 & Goals & None \\
      Fan et al., 2013             & ABA                 & Goals & None \\
      Fan et al., 2014             & ABA                 & No & Dialogue \\
      Zeng et al., 2018            & ABA                 & No & Argument \& Context \\ 
      Zeng et al., 2020            & ABA                 & Yes & Dialogue \\ 
      Kakas and Mora\"\i tis, 2003 & Logic Programming   & No & None \\ 
      M{\"u}ller and Hunter, 2012  & ASPIC+              & Value & None \\
      Amgoud and Prade, 2006       & Structured          & Decision & Arguments \\ 
      Amgoud and Prade, 2009       & AA                  & Decisions & Arguments \\ 
      Teze et al., 2020            & DeLP                & Arguments & None \\ 
      Labreuche 2013               & Propositional Logic & Attributes & None \\ 
      \hline      
    \end{tabular}
    \end{small}    
  \end{center}
\end{table}

A summary of related work on argumentation-based decision making is
shown in Table~\ref{table:related}.
These existing works 
can use different argumentation frameworks, e.g. ABA~\cite{Toni14},
ASPIC+~\cite{Prakken10}, Logic 
Programming~\cite{Lloyd87}. Many of them
accommodate preference modelling for which preferences can be
expressed over various elements of their decision making
models, as indicated in the table. The majority of these works
\cite{Kakas03,Matt09,Marreiros07,Dung08,Muller12,Visser12,Fan13-TAFA,Fan13-CLIMA,Labreuche13,Teze20}
do not explicitly consider explanations but rely on the underlying
argumentation formalism to realise decision making interpretability. A
few works \cite{Fox07,Amgoud06,Amgoud09} explicitly construct and
label arguments for and against decision candidates as a form of
explanations. \cite{Fan14-aamas} and \cite{Zeng20} employ dialogues as
a form of explanation. Lastly, \cite{Zeng18} consider explanations in the
form of arguments and contexts.

\section{Decision Making Abstractions}
\label{sec:new}

In this section, we introduce Abstract Decision Frameworks (ADF) and
Preferential Decision Frameworks (PDF) to model decision making
problems involving decisions, goals and preferences. We introduce
several decision making criteria to identify ``good'' decisions in
these frameworks. Some of these notions correspond to notions in the literature
as shown in Appendix~\ref{sec:relation}.

\subsection{Abstract Decision Frameworks (ADFs)}

{\em Abstract Decision Frameworks} describe the relation between {\em
  decisions} and {\em goals} in a decision problem. 

\begin{definition}
\label{dfn:adf}

An {\em Abstract Decision Framework (ADF)} is a tuple \nadf{}, with

\begin{itemize}
\item
a (finite) set of \emph{decisions} $\decisions = \{d_1, \ldots, d_n\},
n > 0$; 

\item
a (finite) set of \emph{goals} $\goals = \{g_1, \ldots, g_l\}, l > 0$;

\item
  a mapping $\dmg: \decisions \rightarrow
  2^\goals$
  \st\ 
given a decision $d \in \decisions$, $\dmg(d) \subseteq \goals$
denotes the set of goals \emph{met by} $d$.
\end{itemize}
\end{definition}

The following simple example, adapted
from \cite{Matt09}, illustrates the notion of ADF.

\begin{example}
\label{exp:adf}

An agent is deciding on accommodation in London. The three candidate
decisions (\decisions) are John Howard Hotel ($jh$), Imperial College
Halls ($ic$), and Ritz ($ritz$). The agent deems two goals (\goals)
$cheap$ and $near$ as important, where $jh$ meets the goal $near$
($\dmg(jh) = \{near\}$), $ic$ meets both goals ($\dmg(ic) = \{near,
cheap\}$), and $ritz$ meets no goal ($\dmg(ritz) = \{\}$). 
\end{example}

In the remainder of this section, unless otherwise specified, we
assume as given a generic ADF $\nadf$. 

We define three decision criteria for ADFs, characterising different
notions of ``good'' decisions:
{\em strongly dominant} decisions, which select as ``good'' decisions
meeting {\em all} goals;
{\em dominant}  decisions, meeting all goals that are {\em ever met}
by any decision; and 
{\em weakly
  dominant} decisions, meeting a set of goals which is {\em not a
  subset} of the set of goals met by any other decision. Formally,

\begin{definition}
A decision $d \in \decisions$ (in \nadf) is
	\begin{itemize}
		\item 
\label{dfn:SD}
{\em strongly dominant} \ifaf\
$\dmg(d) = \goals$; 
\item 
\label{dfn:d}
{\em dominant} \ifaf\
there is no $g' \in \goals \setminus \dmg(d)$ with $g' \in \dmg(d')$
for some $d' \in \decisions \setminus \{d\}$;
\item 
\label{dfn:wd}
{\em weakly dominant}  \ifaf\
there is no $d' \in \decisions \setminus \{d\}$ with $\dmg(d) \subset
\dmg(d')$.
\end{itemize}
\end{definition}

In Example~\ref{exp:adf}, $ic$ is a strongly dominant decision as it
meets both $cheap$ and $near$, and there is no other strongly dominant
decision.
%
%
%
In 
the same example, $ic$ is the only dominant and the only weakly dominant decision.
The following examples illustrate, \respectively, the cases when dominant decisions exist,
but there is no strongly dominant decision, and when weakly dominant decisions exist, but there is not dominant or strongly dominant decision.

\begin{example}
\label{expD}

We again consider the problem of an agent deciding accommodation in
London, but represented by an ADF \nadf\ with 
$\decisions=\{jh,ritz\}$, $\goals=\{cheap,near\}$, $\dmg(jh)=\{near\}$, and $\dmg(ritz)=\{\}$. Now, differently from Example~\ref{exp:adf},
no decision meets both goals, $cheap$ and $near$. Nevertheless, $jh$
is a better decision than $ritz$ as it meets $near$ whereas $ritz$
meets no goal. Thus, although there is no strongly dominant decision,
	$jh$ is a dominant (as well as a weakly dominant) decision.

\end{example}
%
%
%
%

\begin{example}
\label{expWD}
We now consider the decision problem represented by the ADF $\nadf$ with 
$\decisions=\{jh,ic, ritz\}$, $\goals=\{cheap,near,clean\}$, $\dmg(jh)=\{near,clean\}$, $\dmg(ic)=\{cheap,clean\}$, and $\dmg(ritz)=\{clean\}$. 
Here, differently from Example~\ref{exp:adf},
there is a new goal $clean$ that is met by all decisions. However,
$ic$ no longer meets $near$, hence $ic$ is not strongly
dominant. Moreover, $ic$ does not meet $near$, which is met by $jh$,
hence $ic$ is not dominant. Similarly, $jh$  is neither strongly
dominant nor dominant. However, since $ic$ and $jh$ both meet goals
that are not met by the other, the are both weakly dominant. Finally,
$ritz$ meets no goal that is not met by any other decision
and is not even weakly dominant.
\end{example}

In the remainder of this section, we identify simple connections
amongst and properties of the three given decision criteria. First, it
is easy to see that all dominant decisions meet the same set of goals:
\begin{proposition}
\label{oneDominantSet}
For any $d,d' \in \decisions$, if $d,d'$ are dominant then $\dmg(d) =
\dmg(d')$. 
\end{proposition}

Moreover, dominant decisions are guaranteed to be weakly dominant and
strongly dominant decisions are guaranteed to be  dominant:
\begin{proposition}
\label{sdISd}
For any $d\in \decisions$:

\begin{itemize}
\item if $d$ is strongly dominant then it is dominant;

\label{dISwd}
\item if $d$ is dominant then it is weakly dominant.
\end{itemize}
\end{proposition}

Further, if there are strongly dominant decisions, then decisions are
(weakly) dominant \ifaf{} they are strongly dominant:

\begin{proposition}
\label{prop:allEqual}
Let 
$S_{s}=\{ d \in \decisions | d $ is strongly dominant$\}$,
$S_{d}=\{ d \in \decisions | d $ is dominant$\}$,
and
$S_{w}=\{ d \in \decisions | d $ is weakly dominant$\}$.
If 
$S_{s} \neq \{\}$, then $S_{s} = S_{d} = S_{w}$. 
\end{proposition}

Similarly, if there are dominant decisions, then a decision is weakly
dominant \ifaf{} it is dominant:

\begin{proposition}
\label{prop:dEQw}
Let $S_{d}=\{ d \in \decisions | d $ is dominant$\}$ and $S_{w}=\{ d
\in \decisions | d $ is weakly dominant$\}$. If $S_{d} \neq \{\}$,
then $S_{d} = S_{w}$. 
\end{proposition}

Proposition~\ref{oneDominantSet} indicates
that, when there are (strongly) dominant decisions for the given ADF,
then they are all equally ``good'' in meeting goals, since $\dmg(d) =
\goals$ for all strongly dominant decisions $d$, and $\dmg(d) =
\dmg(d')$ for all $d, d'$ dominant decisions. Instead, as illustrated
by Example~\ref{expWD}, if there are only weakly dominant decisions,
then there are multiple such decisions and they meet different sets
of goals. Formally:

\begin{theorem}
\label{twoSetWD}
Let 
$S_{d}=\{ d \in \decisions | d $ is dominant$\}$
and
$S_{w}=\{ d \in \decisions | d $ is weakly dominant$\}$.
If 
$S_d = \{\}$ and $S_{w} \neq \{\}$, then there exists $d, d' \in
S_{w}, d \neq d'$ and $\dmg(d) \neq \dmg(d')$.
\end{theorem}

Hence, whereas discriminating amongst (strongly) dominant decisions is
not an issue for the decision maker, selecting one weakly dominant
decision amongst all possible such decisions requires other measures
to discriminate amongst goals. We introduce {\em preferences} 
next to further identify suitable decisions.

\subsection{Abstract Decision Frameworks with Preferences (PDFs)}
Preference rankings over sets of goals could help us
further rank weakly dominant decisions. We incorporate these rankings
into ADFs as follows.

\begin{definition}
\label{dfn:preferentialDecisionFramework}
A {\em Preferential Decision Frameworks} (PDF) is a tuple \neadfps, in
which \nadf{} is an ADF and $\preferenceSet$ is a partial order over
$2^\goals$.
\end{definition}

In the remainder of this section, when we refer to decision frameworks
with preferences, we assume as given a generic PDF
\fe=\neadfps. Moreover, as conventional, we will use $s
\preferenceSetstrict s'$ to denote $s \preferenceSet s'$ and $s'
\notpreferenceSet s$.

To ease presentation, we define the notion of {\em comparable goal
  set}, namely the set of all goal sets appearing in the ranking, as
follows: 
\begin{definition}
\label{dfn:comparableGoalSet}
The {\em comparable goal set} (in $\fe$) is $\cgset \subseteq
2^{\goals}$ such that 
\begin{itemize}
\item
for every $s \in \cgset$, there is $s' \in \cgset, s \neq s'$, such
that either $s \preferenceSet s'$ or $s' \preferenceSet s$; 

\item
for every $s \in 2^{\goals} \setminus \cgset$, there is no $s' \in
2^{\goals}$, such that $s \preferenceSet s'$ or $s' \preferenceSet s$.
\end{itemize}
\end{definition}

Hence sets of goals in $\cgset$ are ``comparable'' (according to
$\preferenceSet$). We illustrate $\cgset$ with the following example.

\begin{example}
\label{cgsExp}
Let \goals{} be $\{g_1, g_2, g_3, g_4, g_5\}$ and $\preferenceSet$
be such that\footnote{Here and in the remainder of the paper, we
  sometimes give  $\preferenceSetstrict$ instead of 
$\preferenceSet$ and assume that $\preferenceSet$ is any 
partial order with the strict counterpart $\preferenceSetstrict$.} 

\begin{center}
$\{g_5\} \preferenceSetstrict \{g_4\}  \preferenceSetstrict \{g_3\}
  \preferenceSetstrict \{g_4, g_5\}  \preferenceSetstrict  \{g_2\}
  \preferenceSetstrict  \{g_1\}$. 
\end{center}
Then the comparable goal set is  $\cgset=\{\{g_1\}, \{g_2\}, \{g_3\},
\{g_4\}, \{g_5\}, \{g_4, g_5\}\}$.
\end{example}

\begin{definition}
\label{decFuncPSet}

A decision $d \in \decisions$ is {\em \pfrSet}  (in $\neadfps$) iff it
is weakly dominant in
\nadf{}
and for all weakly dominant decisions $d'
\in \decisions$ such that $d \neq d'$, for all $s \in \cgset$ (the
comparable goal set in $\neadfps$):  
\begin{itemize}
\item
  if $s \not \subseteq \dmg(d)$ and $s \subseteq \dmg(d')$, then there
  exists $s' \in \cgset$, \st{}

\hspace{20pt}$s' \preferenceSet s$, 
\hspace{40pt}$s' \subseteq \dmg(d)$, and 
\hspace{40pt}$s' \not\subseteq \dmg(d')$.
\end{itemize}

\end{definition}

Definition~\ref{decFuncPSet} states that to select a decision $d$,
firstly, we check that $d$ is weakly dominant; then, we check against
all other weakly dominant decisions $d'$ to ensure that for any member
$s$ of the comparable goal set, if $d'$ meets $s$ but $d$ does not,
then there exists some member $s'$ more preferred than, or as
preferred as, $s$ such that $d$, but not $d'$, meets (all goals in) $s'$.
Thus, the notion of \pfrSet\ decisions uses the preference ranking
only to discriminate amongst weakly dominant decisions; indeed, if a
decision $d$ is not weakly dominant in the first place, there exists
some other decision meeting all goals met by $d$ and more, and thus
``better''. 

\begin{example}
\label{DGExpS}
Consider $\neadfps$ with decisions $\decisions=\{d_1, d_2\}$,
goals $\goals=\{g_1, g_2, g_3, g_4,$ $g_5\}$, 
$\dmg(d_1)=\{g_2,g_4,g_5\}$, 
$\dmg(d_2)=\{g_2,g_3\}$, 
and $\preferenceSet$ as in Example~\ref{cgsExp}. 
Here, 
both decisions meet $g_2$, $d_1$ meets both $g_4$ and $g_5$ whereas $d_2$ meets $g_3$.
Also,
although $g_3$
is more preferred than $g_4$ and $g_5$ individually, $g_4$ and $g_5$
together are more preferred than $g_3$. Hence, $d_1$ is a \pfrSet{}
decision and $d_2$ is not.

\end{example}

\section{
	\Flat\ Explanations}
\label{sec:explanation}

Thus far, we have introduced several decision making criteria for
selecting ``good'' decisions amongst a set of candidates. Since there
can be multiple decisions, goals and reference relations involved in a
decision making process, it would be useful to {\em explain} the
reason for selecting a decision by identifying decisions
and goals that ``influence'' the selection. In this section we define {\em
  \flat\ explanations} for a decision meeting (or not meeting) certain
decision criteria as follows.

\begin{definition}
  \label{dfn:explanation}
  
  Given $\nadf$ and $d \in \decisions$

  \begin{itemize}
    \item 
      $\dmg(d)=\goals$ is a {\em \flat\ explanation for $d$ being strongly
      dominant}; 

    \item
      $\goals \setminus \dmg(d)$ is a {\em \flat\ explanation for $d$ {\bf
        not} being strongly dominant};

    \item 
      $\expPair{\dmg(d)}{\goals \setminus \dmg(d)}$
      is a {\em \flat\ explanation for $d$ being dominant};
      
    \item
      $\{\expPair{d_i}{g_j} | d_i \in \decisions, g_j \in 
      \goals$, and $g_j \notin \dmg(d), g_j \in \dmg(d_i)\}$
      is a {\em \flat\ explanation for $d$ {\bf not} being dominant};

    \item 
      $\expPair{G}{\{(g_1, d_1),\ldots,(g_n, d_n)\}}$ for which $G
      \subseteq \dmg(d)$ and $g_i \in \dmg(d), g_i \notin \dmg(d_i)$
      \st{} for each $d' \in \decisions, d' \neq d$, either
      $\dmg(d') \subseteq G$ or $d' \in \{d_1, \ldots, d_n\}$ is a
         {\em \flat\ explanation for $d$ being weakly dominant}; 
      
    \item
      $\{d' \in \decisions|\dmg(d) \subset \dmg(d')\}$ is a {\em
      \flat\ explanation for $d$ {\bf not} being weakly dominant}.
  \end{itemize}
\end{definition}

Basically, to explain a decision being strongly dominant (or not), we
identify the goals it meets being the set of all goals 
(or not). To explain a decision being dominant, we identify the goals
the decision meets and does not meet. To explain a decision not being
dominant, we identify decision-goal pairs such that the goal in each
pair is met by the decision in the pair but not by the decision under
consideration. To explain a decision being weakly dominant, we
identify goals met by this decision and goal-decision pairs \st{} the
goal in a pair is met by the decision under consideration but not by the
decision in the pair. To explain a decision not being weakly dominant, 
we identify the set of decisions that meet more goals than the
decision under consideration.

\begin{example}
  \label{exp:explanation}

  Given the ADF in Example~\ref{exp:adf}:
  \begin{itemize}
  \item
    $\{cheap, near\}$ is a \flat\ explanation for $ic$ being strongly
    dominant as these are the two goals to be met in this example; 
  \item
    $\{cheap\}$ is a \flat\ explanation for $jh$ being not strongly
    dominant as $cheap$ is the goal not met by $jh$;
  \item
    $\{cheap, near\}$ is a \flat\ explanation for $ritz$ being not
    strongly dominant as $cheap$ and $near$ are the goals not met
      by $ritz$.
  \end{itemize}

  Given the ADF in Example~\ref{expD}:
  \begin{itemize}
  \item
    $\expPair{\{near\}}{\{cheap\}}$ is a \flat\ explanation for $jh$
    being dominant as $near$ is a goal met by $jh$ but $cheap$ is
      not; 
  \item
    $\{\expPair{jh}{near}\}$ is a \flat\ explanation for $ritz$ being not
    dominant as $near$ is a goal met by $jh$ but not by $ritz$.
  \end{itemize}

  Given the ADF in Example~\ref{expWD}:
  \begin{itemize}
  \item
    $\expPair{\{clean\}}{\expPair{near}{ic}}$ is a \flat\ explanation for
    $jh$ being weakly dominant as in addition to meeting the goal
    $clean$, $jh$ also meets the goal $near$, which is not met by the
    alternative decision $ic$; 
  \item
    $\expPair{\{clean\}}{\expPair{cheap}{jh}}$ is a \flat\ explanation
    for $ic$ being weakly dominant as in addition to meeting the
      goal $clean$, $ic$ also meets the goal $cheap$, which is not met
      by the alternative decision $jh$; 
  \item
    $\{jh, ic\}$ is a \flat\ explanation for $ritz$ not being weakly
    dominant as both $jh$ and $ic$ meet all goals met by $ritz$
      and more.
  \end{itemize}  
\end{example}

 \Flat\ explanations fulfil several properties, considered next.

\begin{proposition}
  \label{prop:expProperty}

  Given $\nadf$ and $d \in \decisions$

  \begin{enumerate}
    \item 
      $d$ has at least one \flat\ explanation as given in
      Definition~\ref{dfn:explanation};

    \item
      if $d$ is not weakly dominant, let $E$ be a \flat\ explanation for
      $d$; then there is no $E'$ which is a \flat\ explanation for $d$ \st{} $E' \neq
      E$.
  \end{enumerate}  
\end{proposition}

Proposition~\ref{prop:expProperty} sanctions, firstly, that
each decision in an ADF always admits at least one \flat\
explanation: this is desirable as it implies that each ``good'' decision can be somewhat justified. Secondly, unless a decision is weakly dominant, its
\flat\ explanation (per Definition~\ref{dfn:explanation}) is
unique: this is desirable as it implies consistency across multiple requests for explanation for the same ``good'' decision (these requests may be by the same explainee at different times or by multiple explainees).

Analogously to the case of ADFs,  we would like to explain why decisions in PDFs
are \pfrSet{} or not. We can do so as follows.
\begin{definition}
  \label{dfn:explanationPfrSet}
  
Given $\neadfps$ and $d \in \decisions$,

\begin{itemize}
\item 
  a {\em \flat\ explanation for $d$ being \pfrSet} is a pair
  $\expPair{G}{\{(S_1, d_1'), \ldots, (S_n, d_n')\}}$ where $G
  \subseteq \dmg(d)$, $S_i \subseteq \dmg(d), S_i \not\subseteq
  \dmg(d_i')$ and there is no $S' \subseteq \dmg(d_i')$ \st{} $S'
  \preferenceSet S_i$, for $i \in \{1,\ldots,n\}$. Moreover, for each
  $d' \in \decisions, d' \neq d$, $\dmg(d') \subseteq G$ or $d' \in
  \{d_1, \ldots, d_n\}$;  

\item 
  a {\em \flat\ explanation for $d$ {\bf not} being \pfrSet} is
  $\{d' \in \decisions|$ either $\dmg(d) \subset \dmg(d')$ or there
  exists $S \subseteq \dmg(d')$, \st{} for all $S'\subseteq \dmg(d), S
  \preferenceSet S'\}$.
\end{itemize}
\end{definition}

Definition~\ref{dfn:explanationPfrSet} is given in the same spirit as
Definition~\ref{dfn:explanation}, where a decision (not) being \pfrSet{}
is explained in terms of its relation with other decisions and goals as well as with the
preference relations.

\begin{example}
  \label{exp:explanationPfrSet}

  Given the PDF shown in Example~\ref{DGExpS}:

  \begin{itemize}
  \item
    $\expPair{\{g_2\}}{\{\expPair{\{g_4,g_5\}}{d_2}\}}$ is a \flat\
    explanation for $d_1$ being \pfrSet{} as in addition to $d_1$
    meeting $g_2$, $d_1$ also meets the more preferred goals $g_4, g_5$,
    which are not met by $d_2$; 
    
  \item
    $\{d_1\}$ is a \flat\ explanation for $d_2$ being not \pfrSet{}
    as $d_1$ meets more preferred goals than $d_2$.
  \end{itemize}  
\end{example}

\begin{proposition}
  \label{prop:expPropertyPfrSet}

Given $\neadfps$ and $d \in \decisions$,

\begin{enumerate}
\item
  $d$ has at least one \flat\ explanation as given in
  Definition~\ref{dfn:explanationPfrSet};

\item
	if $d$ is not \pfrSet{}, let $E$ be a \flat\ explanation for $d$ not being \pfrSet; then
		there is no $E'$ which is a \flat\ explanation for $d$ not being \pfrSet{}
		\st{} $E' \neq E$.
\end{enumerate}
\end{proposition}

The explanations defined in this section are \flat\ in that they provide the basic reasons behind a decision, but ignore any indication of the role the underlying decision criterion has in explaining the decisions. \Dialogical\ explanations, defined next, remedy this issue, while still embedding all information in \flat\ explanations, as we will show. 

\section{\Dialogical\ Explanations for ADFs and PDFs}
\label{sec:dial}

Thus far, we have introduced a set of decision criteria and \flat\
explanations for decisions meeting these criteria or not. 
In this section we define \dialogical\ explanations for  these decisions. 
To do so, given an ADF or PDF and a decision criterion, we construct an
ABA framework so that admissible arguments are selected decisions in
the ADF \wrt{} the decision criterion.
We can then use special forms of dispute trees as \dialogical\ explanations, explaining the mechanisms by means of which those decisions are inferred as ``good'', in such a way that corresponding \flat\ explanations can be extracted from them too. 
We  first give preliminary background on ABA and (our variants of) dispute trees (Section~\ref{sec:bkgpre}). Then, we define \dialogical\ explanations for ADFs (Section~\ref{sec:ABADec})  and PDFs (Section~\ref{sec:ABADecP}). 
The proofs of the technical results for this section 
are in Appendix~\ref{sec:proof}.

\subsection{Preliminaries}
\label{sec:bkgpre}
\subsubsection{ABA}
\label{sec:bkg}

Material in this section is mostly adapted from \cite{Toni14,Cyras17}. 
An \emph{ABA framework} is a tuple $\langle \mc{L,R,A,C}\rangle$
where  
\begin{itemize}
\item $\langle \mc{L,R}\rangle$ is a deductive system, with a
 {\em language}  $\mc{L}$ and a set of inference {\em rules}
 $\mc{R}$ of the form $\sent_0 \gets \sent_1,\ldots, \sent_m (m > 0)$
 or $\sent_0 \leftarrow$ with  $\sent_i \in \mc{L}$, for $i=0, \ldots,
 m$,

\item $\mc{A} \subseteq \mc{L}$ is a (non-empty) set, whose
 elements are referred to as {\em assumptions},

\item $\mc{C}$ is a total mapping from $\mc{A}$ into
 $2^{\mc{L}} - \{\{\}\}$, where each $c \in \mc{C}(\alpha)$
  is a {\em contrary} of $\alpha$.

\end{itemize}

Basically, ABA frameworks can be defined for any logic specified by
means of inference rules. Some of the sentences in the underlying
language are assumptions, and each such sentence can be
``contradicted'' by any of its contraries. The following example is a
simple illustration, with a set of propositions as language. 

\begin{example}
\label{ABAAFEXP}
Let $\lracF$ be as follows:

$\bullet$
$\mc{L} = \{a, b, c, p, q, r,s\}$
and 
$\mc{R} = \{p \la a, \quad q \la b,s,\quad r \la c,\quad  s\la \}$,

$\bullet$
$\mc{A} = \{a, b, c\}$,

$\bullet$
$\mc{C}(a) = \{q\}$, $\mc{C}(b) = \{p\}$,
  $\mc{C}(c) = \{r, q\}$. 
\end{example}
Given a rule $\sent_0 \gets \sent_1,\ldots,\sent_m$ or $\sent_0
\leftarrow$ in an ABA framework, $\sent_0$ is referred as the {\em
  head} and $\sent_1,\ldots,\sent_m$ or the empty sequence,
\respectively, as the {\em body} of the rule. ABA frameworks where no
assumption occurs as the head of a rule are 
{\em
  flat}. As an illustration, the ABA framework in
Example~\ref{ABAAFEXP} is flat, as assumptions there only occur in the 
body of rules. In this paper, all ABA frameworks will be
flat. Therefore, all remaining definitions in this section are
restricted to the case of flat ABA frameworks. 

In ABA, informally, arguments are deductions of claims supported by
sets of assumptions, and attacks against arguments are directed at the
assumptions in their supports, and are given by arguments with an
assumption's contrary as their claim. Formally, given an ABA framework
\lracF, a {\em deduction for $\sent \in \mc{L}$ supported by} $\sentset
\subseteq \mc{L}$ is a finite tree with nodes labelled by sentences in
$\mc{L}$ or by $\tau$\footnote{The symbol $\tau$ is such that $ \tau
  \notin \mc{L}$. $\tau$ stands for ``true'' and intuitively
  represents the empty body of rules~\cite{Dung09}.}, such that 
\begin{enumerate}
\item the root is labelled by $\sent$
\item for every node $N$
\begin{itemize}
\item if $N$ is a leaf then $N$ is labelled either by a sentence in
  $\sentset$ or  by $\tau$;
\item if $N$ is not a leaf and $\sent_0$ is the label of $N$, then there
 is an inference rule $\sent_0 \gets \sent_1,...,\sent_m (m \ge 0)$
 and either $m = 0$ and the child of $N$ is $\tau$ or $m > 0$ and $N$
 has $m$ children, labelled by $\sent_1,...,\sent_m$ (\respectively) 
\end{itemize}
\item $\sentset$ is the set of all sentences labelling the leaf nodes.
\end{enumerate}
Then, an {\em argument for $\sent \in \mathcal{L}$ supported by $\asmset \subseteq \mathcal{A}$} is a deduction for  $\sent$  supported by $\asmset$.
We will use the shorthand $\argu{\asmset}{\sent}$ to denote an argument for
$\sent$ supported by $\asmset$. Given argument $\argu{\asmset}{\sent}$,
$\asmset$ is referred to as the {\em support} and $\sent$ as the {\em
  claim} of the argument. Figure~\ref{argEXPS} illustrates this notion
of argument for the ABA framework in Example~\ref{ABAAFEXP}.

\begin{figure}[h]
\begin{scriptsize}
\[\xymatrix@1@=10pt{
p                  &&&&& q                           &&&&& r \\
	a \ar@{-}[u]           &&&& b\ar@{-}[ur]  &  &s\ar@{-}[ul]       &&&& c \ar@{-}[u]\\
		   &&&&           & &  \tau \ar@{-}[u]   &&&& \\
{\argu{\{a\}}{p}}  &&&&& {\argu{\{b\}}{q}} &&&&& {\argu{\{c\}}{r}} }\] 
\end{scriptsize}
\caption{Arguments in Example~\ref{ABAAFEXP}, as trees (at the top) and
  using the shorthand (at the bottom).}
\label{argEXPS}
\end{figure}
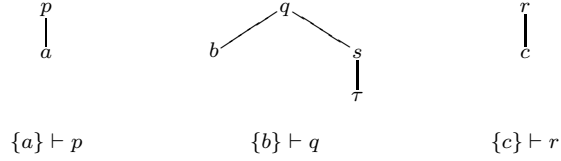

Argumentation frameworks in general are useful for resolving conflicts between 
information, as captured by arguments. Conflicts are typically
represented as \emph{attacks} between arguments~\cite{Dung95}. In ABA,
attacks are defined in terms of the notion of contrary of assumptions,
as follows, given an ABA framework
\lracF:

\begin{itemize}
\item an argument $\asmset_1 \vdash \sent_1$ attacks an argument
  $\asmset_2 \vdash \sent_2$ \ifaf{} the claim $\sent_1$ of the first argument
  is a contrary of one of the assumptions in the support $\asmset_2$
  of the second argument (i.e. $\exists \alpha \in \asmset_2$ such that
  $\sent_1 \in \mc{C}(\alpha)$);
\end{itemize}
This notion of attack between arguments can be lifted up to 
sets of arguments and sets of assumptions, as follows:
\begin{itemize}
\item a set of arguments \argAs{} attacks a set of arguments \argBs{} if
 some argument in \argAs{} attacks some argument in \argBs;
\item a set of assumptions $\asmset_1$ attacks a set of assumptions
  $\asmset_2$ \ifaf{} an argument supported by a subset of
  $\asmset_1$ attacks an argument supported by a subset of $\asmset_2$.
\end{itemize}
As an illustration, for the ABA framework in Example~\ref{ABAAFEXP},
\begin{itemize}
\item $\argu{\{a\}}{p}$ attacks $\argu{\{b\}}{q}$,
$\argu{\{b\}}{q}$ attacks $\argu{\{a\}}{p}$, $\argu{\{b\}}{q}$ attacks
$\argu{\{c\}}{r}$, and $\argu{\{c\}}{r}$ attacks $\argu{\{c\}}{r}$
\item 
$\{\argu{\{a\}}{p}\}$ attacks $\{\argu{\{b\}}{q},\argu{\{c\}}{r}\}$ 
\item 
\{$a$\} attacks \{$b$\}, \{$b$\} attacks \{$a,c$\}, \{$b$\}
attacks \{$c$\}, and \{$c$\} attacks any set of assumptions containing $c$.
\end{itemize}

With argument and attack defined, all argumentation semantics
given for abstract argumentation \cite{Dung95} can be applied in (flat)
ABA. These semantics can be defined for assumptions, as in 
\cite{Bondarenko93,Bondarenko97}, or, equivalently, for
arguments~\cite{Dung07,Toni14}. In this paper, given an ABA framework
$AF=\lracF$, we will use the following notions of admissibility for
sets of assumptions and for individual arguments: 
\begin{itemize}
\item 
\emph{a set of assumptions is admissible} (in $AF$) \ifaf{} it
does not attack itself and it attacks all $\asmset \subseteq \mc{A}$
that attack it;  

\item 
\emph{an argument $\argu{\asmset}{\sent}$ is admissible (in $AF$)
  supported by $\asmset'\subseteq \mc{A}$} \ifaf{} $\asmset \subseteq
\asmset'$ and $\asmset'$ is admissible (in $AF$).

\end{itemize}
As an illustration, for the ABA framework in Example~\ref{ABAAFEXP},
\begin{itemize}
\item the
sets of assumptions $\{a\}$, $\{b\}$, $\{\}$ are all admissible
(and no other set of assumptions is admissible); 

\item arguments $\argu{\{a\}}{p}$,  $\argu{\{b\}}{q}$, $\argu{\{a\}}{a}$, $\argu{\{b\}}{b}$
are all admissible (and no other argument is admissible).
\end{itemize}

\subsubsection{Dispute trees}
\label{sec:pre}

Here, unless specified otherwise, we will
assume as given a generic ABA framework 
$\lracF$.

We will use variants of the {\em abstract dispute trees} of \cite{Dung06b} to
provide explanations for recommended decisions. 
An \emph{abstract dispute tree}
for an argument \argA{} is a (possibly
infinite) tree \atree{} such that:\footnote{Here, $a$ stands for
  'abstract'. Also, `proponent' and `opponent' should be seen as roles/fictitious
  participants in a debate rather than actual agents.} 

\begin{enumerate}
\item every node of \atree{} holds an argument $\argB$ 
	and is labelled by either {\em proponent} (P) or {\em opponent} (O),
  but not both, denoted by $L: \argB$, for $L \in \{$P,O$\}$;  
\item the root of \atree{} is a \pro{} node holding $\argA$;
\item for every \pro{} node $N$ holding an argument $\argB$, and
 for every argument $\tt C$ that attacks $\argB$, there
 exists a child of $N$, which is an \opp{} node holding $\tt C$;
\item for every \opp{} node $N$ holding an argument $\tt B$, there
 exists 
		\emph{at most}\footnote{In the original definition of abstract dispute tree~\cite{Dung06b}, every \opp{} node is required to have \emph{exactly} one child. We incorporate this requirement into the definition of \emph{admissible} dispute tree given later, so that our notion of admissible abstract dispute tree and the admissible abstract dispute trees of ~\cite{Dung06b} coincide.} one child of $N$ which is a \pro{} node holding an argument
 which attacks some assumption $\alpha$ 
 in the support of $\tt B$;\footnote{An argument attacks an assumption
		if the argument supports a contrary of the assumption.}  if $N$ has a child attacking $\alpha$, then $\alpha$ 
		is said to be the {\em culprit} in $\tt B$; 
\item there are no other nodes in \atree{} except those given by 1-4
 above.
\end{enumerate}
The set of all assumptions in (the support of arguments held by) the
\pro{} nodes in \atree{} is called the {\em defence set} of
\atree. 

Abstract dispute trees can be used as the basis for 
computing various argumentation semantics, including the admissibility semantics, as follows\del{,
given an ABA framework $AF$}:

\begin{itemize}
\item  
	Let an abstract dispute tree \atree{} be {\em admissible} \ifaf{} 
		each \opp{} node has {\em exactly} one child and no
culprit in the argument of an \opp{} node in \atree{} belongs to
the defence set of \atree. 
		\item The defence set of an admissible abstract
			dispute tree \del{for an argument $\argA$ (in $AF$)} is admissible (Theorem
		5.1 in \cite{Dung06b}), and thus the root node of an admissible dispute tree 
		is admissible
		. 
	\item If an argument $\argA$ \del{(in $AF$)} is admissible then there exists an admissible abstract
		dispute tree for $\argA$ \del{(in $AF$)} (Theorem
		5.1 in \cite{Dung06b}).
\end{itemize}

%
Figure~\ref{argEXPS-ADT} gives an example of an (infinite) admissible
	dispute tree, with (admissible) defence set $\{a\}$, for the ABA framework in Example~\ref{ABAAFEXP}. 

%
%
        
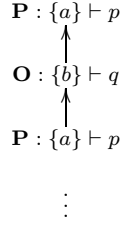
\begin{figure}
\begin{scriptsize}
\[\xymatrix@1@=14pt{
\pro: \argu{\{a\}}{p} & \\
\opp: \argu{\{b\}}{q}\ar[u] & \\
\pro: \argu{\{a\}}{p}\ar[u] & \\                   
\vdots & 
}
\] 
\end{scriptsize}
\caption{Admissible abstract dispute tree for
	the ABA framework in Example~\ref{ABAAFEXP}}
\label{argEXPS-ADT}
\end{figure}

We will use three variants of abstract dispute trees, given below.

%

A \emph{maximal dispute tree} \cite{AA-CBR-KR16} for some argument
$\argA$ is an abstract dispute tree \atree{} for $\argA$ \st{} for all
opponent nodes $\opp: \argB$ in \atree{} that are leaf nodes there is
no argument $\tt C$ \st\ $\tt C$ attacks $\argB$. 

Note that admissible abstract dispute trees are maximal, but not all
maximal dispute trees are
admissible~\cite{AA-CBR-KR16}. Given an ABA framework \lracF{} with
  $\mc{R} = \{c \gets \}$,
%
  $\mc{A} = \{a, b\}$,
%
  $\mc{C}(a) = \{b\}$, $\mc{C}(b) = \{c\}$,
Figure~\ref{fig:maximalTrees} shows two abstract dispute trees for the
argument $\argu{\{a\}}{a}$, the tree on the left-hand side is maximal
whereas the one on the right-hand side is not.

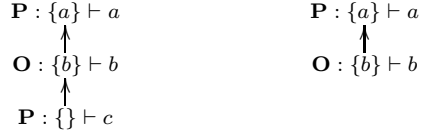
\begin{figure}
  \begin{scriptsize}
    \[\xymatrix@1@=10pt{
      \pro: \argu{\{a\}}{a}       &&&&&& \pro: \argu{\{a\}}{a} \\
      \opp: \argu{\{b\}}{b}\ar[u] &&&&&& \opp: \argu{\{b\}}{b} \ar[u] \\
      \pro: \argu{\{\}}{c}\ar[u]  &&&&&&  \\      
    }\]
  \end{scriptsize}
  \caption{Two abstract dispute trees for the argument
    $\argu{\{a\}}{a}$. The tree shown on the left-hand side is
    maximal; the tree on the right-hand side is not.}
  \label{fig:maximalTrees}  
\end{figure}



\begin{definition}
  \label{dfn:LAandBE}
Given any abstract dispute tree $\atree$, let
$LA(\atree)$ denote the set $\{\asm|\anode{\_}{\argu{\asmset}{\_}}$ is
a leaf node in $\atree, \asm \in \asmset\}$ and $LO(\atree)$ denote the
set $\{N|N=\anode{O}{\_}$ is a leaf node in $\atree$$\}$.

\begin{itemize}
\item
  Let $\mc{AT}$ be the set of all admissible abstract dispute trees
  for some argument $\argA$. Then $\atree \in \mc{AT}$ is {\em
    \leastAsm{}} \ifaf{} there is no $\atree_0$ in $\mc{AT}$ \st{}
  $LA(\atree_0) \subset LA(\atree)$. 


\item 
  Let $\mc{MT}$ be the set of all maximal dispute trees for some
  non-admissible argument $\argA$. Then $\mtree \in \mc{MT}$ is {\em
    best-effort} \ifaf{} there is no $\mtree_0$ in $\mc{MT}$ \st{}
  $LO(\mtree_0) \subset LO(\mtree)$. 
\end{itemize}
\end{definition}

\begin{example}
	\label{ex:leastAsm}
Given an ABA framework \lracF{} with $\mc{R} = \{a \gets b; a \gets c;
c \gets\}$, $\mc{A} = \{b, p, q\}$, $\mc{C}(b) = \{z\}$, $\mc{C}(p) =
\{q\}$, $\mc{C}(q) = \{a\}$, Figure~\ref{fig:LA} shows two admissible  
abstract dispute trees for $\argu{\{p\}}{p}$. The tree on the
	left-hand side is \leastAsm, 
	as it has a single leaf node $\anode{\pro}{\argu{\{\}}{a}}$ and there is
        no strict subset of the empty set, so there does not exist
        a tree with a leaf node containing fewer assumptions than the
        node $\anode{\pro}{\argu{\{\}}{a}}$. 
        On the other hand, the tree on the right-hand
        side is not \leastAsm{} as it has a leaf node
        $\anode{\pro}{\argu{\{b\}}{a}}$ and the tree on the left-hand
        side is an abstract dispute tree for the same argument
        $\argu{\{a\}}{a}$ but with a smaller set of assumptions in its
        leaf node, i.e. $\{\} \subset \{b\}$.        
\end{example}

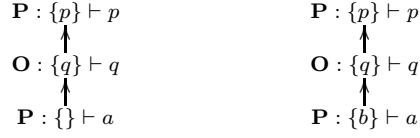
\begin{figure}
  \begin{scriptsize}
    \[\xymatrix@1@=10pt{
      \pro: \argu{\{p\}}{p}       &&&&&& \pro: \argu{\{p\}}{p} \\
      \opp: \argu{\{q\}}{q}\ar[u] &&&&&& \opp: \argu{\{q\}}{q} \ar[u] \\
      \pro: \argu{\{\}}{a}\ar[u]  &&&&&& \pro: \argu{\{b\}}{a}\ar[u] \\            
    }\]
  \end{scriptsize}
	\caption{Two admissible abstract dispute trees for the argument
	$\argu{\{p\}}{p}$ for Example~\ref{ex:leastAsm}: the tree on the 
	left
	is
    \leastAsm; the tree on the 
	right
	is not.}
  \label{fig:LA}  
\end{figure}

\begin{example}
	\label{ex:bestEffort}
Given an ABA framework \lracF{} with $\mc{R} = \{b \gets; d \gets; f
\gets\}$, $\mc{A} = \{a,c,e\}$, $\mc{C}(a)=\{b,c\}$,
$\mc{C}(c)=\{d,e\}$, $\mc{C}(e)=\{f\}$, Figure~\ref{fig:BE} shows two
	non-admissible, maximal abstract dispute trees for $\argu{\{a\}}{a}$.
not.
The tree on the left-hand side is best-effort
as there is no maximal dispute tree for
  $\argu{a}{a}$ which does not contain the node
  $\anode{\opp}{\argu{\{\}}{b}}$; thus there is no ``smaller'' maximal
dispute tree for $\argu{a}{a}$ than the one shown on the left-hand
side. On the other hand, the tree on the right-hand side is not best-effort, as
the tree on the left-hand side contains fewer opponent leaf nodes.

\end{example}

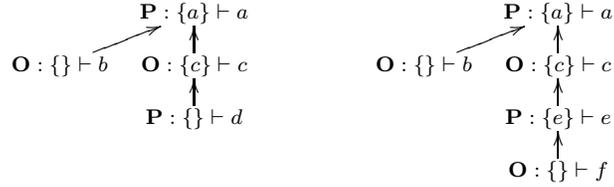
\begin{figure}
  \begin{scriptsize}
    \[\xymatrix@1@=10pt{
                                  & \pro: \argu{\{a\}}{a}       &&&&                             & \pro: \argu{\{a\}}{a} \\
      \opp: \argu{\{\}}{b}\ar[ur] & \opp: \argu{\{c\}}{c}\ar[u] &&&& \opp: \argu{\{\}}{b}\ar[ur] & \opp: \argu{\{c\}}{c}\ar[u] \\
                                  & \pro: \argu{\{\}}{d}\ar[u]  &&&&                             & \pro: \argu{\{e\}}{e}\ar[u]  \\            
                                  &                             &&&&                             & \opp: \argu{\{\}}{f}\ar[u] 
    }\]
  \end{scriptsize}
	\caption{Two maximal abstract dispute trees for the argument
	$\argu{\{a\}}{a}$ for Example~\ref{ex:bestEffort}: the tree on the 
	left
	is
    best-effort; the tree on the 
	right
	is not.}
  \label{fig:BE}  
\end{figure}

In the remainder we will refer to abstract dispute trees simply as dispute trees.

\subsection{\Dialogical\ Explanations for ADFs}
\label{sec:ABADec}

In this section we introduce 
one ABA framework
for each decision criterion for ADFs. These ABA frameworks share some rules,
assumptions and contraries, given next. Here and in the remainder of
this section we will assume as given a generic ADF
$\nadf$, unless
specified otherwise.

\begin{definition}
\label{dfn:coreADFs}
The {\em core ABA framework corresponding to}
$\nadf$
is $\lracFcore$,
where\footnote{When defining ABA frameworks, we omit to indicate the
  language component, as this can be easily inferred from the other
  components (specifically, the language is the set of all sentences occurring in rules,
  assumptions,  and contraries).}

\begin{itemize}
\item
$\Rcore=
  \{met(d,g) \gets | d \in \decisions,  g \in \goals,
  g \in \dmg(d)
  \}$, 
\item 
$\Acore=\{notMet(d,g) | d \in \decisions,  g \in \goals \}$,
\item 
for any $notMet(d,g) \in \Acore$: $\Ccore(notMet(d,g))=\{met(d,g)\}$. 
\end{itemize}
\end{definition}
Intuitively, a full representation, in terms of rules, of the ADF is
included in 
$\Rcore$, and each decision can be
assumed not to meet any goal ($\Acore$), unless it can be proven to
meet it ($\Ccore$). We illustrate the core ABA framework 
with the following example. 

\begin{example}
\label{exp2-core}
(Example~\ref{exp:adf} continued.) The core ABA framework
corresponding to
	\nadf{}
is $\lracFcore$, in which

\begin{itemize} 
\item 
\Rcore\ consists of:

\begin{tabular*}{0.8\textwidth}{@{\extracolsep{\fill} }lll}
$met(ic,cheap) \gets$ & $met(ic,near) \gets$ & $met(jh,near) \gets$\\
\end{tabular*}

\item \Acore\ consists of:

\begin{tabular*}{0.7\textwidth}{@{\extracolsep{\fill} } l l l}
$notMet(jh,cheap)$ & $notMet(ic,cheap)$ & $notMet(ritz,cheap)$ \\
$notMet(jh,near)$ & $notMet(ic,near)$ & $notMet(ritz,near)$ 
\end{tabular*}

\item
for any $d \in \decisions=\{jh,ic,ritz\}$ and $g \in \goals=\{cheap,near\}$: 

$\Ccore(notMet(d,g)) =\{met(d,g)\}$.
\end{itemize}

	In this ABA framework, 
	argument  $\{notMet(jh,near)\} \vdash
notMet(jh,near)$ is attacked by argument $\{\} \vdash met(jh,near)$,
which cannot be attacked (as its support is empty). As another
example, argument  $\{notMet(jh,cheap)\} \vdash notMet(jh,cheap)$
cannot be attacked. 
\end{example}

We then define ABA frameworks to represent the decision criteria,
all incorporating the core ABA framework.

\subsubsection{
Strongly dominant ABA frameworks and \dialogical\ explanations} 

\begin{definition}
  \label{dfn:sdABA}

	For decisions $\decisions$ and goals $\goals$, let 
	the \emph{strongly dominant component} be the ABA framework with

\begin{itemize}
\item
  $\mc{R}_s = \{notSDom(d) \gets notMet(d,g) \; | \; d \in
  \decisions, g \in \goals \}$,
  
\item
  $\mc{A}_s = \{ sDom(d) \; | \; d \in \decisions\}$, 
  
\item
  for any $sDom(d) \in \mc{A}$: $\mc{C}_s(sDom(d)) = \{notSDom(d)\}$.
\end{itemize}

Then,
%
%
  \label{dfn:sdABAADF}
  the {\em strongly dominant ABA framework corresponding to}
  $\nadf$
  is $\lracF$, where

\begin{itemize}
\item
  $\mc{R}= \Rcore \cup \mc{R}_s$,

\item
  $\mc{A}= \Acore \cup \mc{A}_s$, 

\item
for any \asm{} in $\mc{A}$:
		\(\mc{C}(\asm) =
  \begin{cases}
    \Ccore(\asm)   & \quad \text{if } \asm \in \Acore,\\
    \mc{C}_s(\asm) & \quad \text{if } \asm \in \mc{A}_s.\\
  \end{cases}
		\)
  
%
\end{itemize}

\label{SDABA}
\end{definition}

The intuition behind Definition~\ref{SDABA} is as follows: given a
decision $d$, we can assume that $d$ is strongly dominant ($sDom(d)\in
\mc{A}$), unless it can be proven not to be so by proving that it does
not meet some goal (using the rule for $d$ and the goal in question in
$\mc{R}_s$). By definition of the core ABA framework,
then, proving that a decision does not meet a goal can always be
achieved, by assuming so, unless there are reasons against this
assumption, by supporting its contrary. We illustrate the notion of
strongly dominant ABA framework corresponding to a decision framework
as follows.

\begin{example}
\label{exp2}
(Example~\ref{exp2-core} continued.) The strongly dominant ABA
framework corresponding to
$\nadf$
is $\lracF$, in which

\begin{itemize}
\item \mc{R} consists of \Rcore\ in Example~\ref{exp2-core} together with:

  \begin{tabular}{ll}
    $notSDom(ic) \gets notMet(ic,near)$ & 
    $notSDom(ic) \gets notMet(ic,cheap)$ \\
    $notSDom(jh) \gets notMet(jh,near)$ &
    $notSDom(jh) \gets notMet(jh,cheap)$ \\
    $notSDom(ritz) \gets notMet(ritz,near)$ &
    $notSDom(ritz) \gets notMet(ritz,cheap)$ \\
  \end{tabular}
    
\item\mc{A} consists of \Acore\ in Example~\ref{exp2-core} together with:

  \begin{tabular}{lll}
$sDom(jh)$ &  $sDom(ic)$ & $sDom(ritz)$
  \end{tabular}
  
\item 
for any $d \in \decisions=\{jh,ic,ritz\}$ and $g \in \goals=\{cheap,near\}$: 

\begin{tabular}{ll}
  $\mc{C}(notMet(d,g))=\Ccore(notMet(d,g))$ &
  $\mc{C}(sDom(d)) = \{notSDom(d)\}$
  \end{tabular}
\end{itemize}

In this ABA framework,
	the argument $\argu{\{sDom(ic)\}}{sDom(ic)}$ is attacked by 
\begin{center} 
 $\argu{\{notMet(ic,cheap)\}}{notSDom(ic)}$ and 
 $\argu{\{notMet(ic,near)\}}{notSDom(ic)}$ 
\end{center} 
{\noindent}which, in turn, are attacked, \respectively, by arguments
\begin{center}
$\argu{\{\}}{met(ic,cheap)}$ and $\argu{\{\}}{met(ic,near)}$.
\end{center}
{\noindent}Further, there is no argument attacking
either $\argu{\{\}}{met(ic,cheap)}$ or $\argu{\{\}}{met(ic,near)}$.
Hence $\argu{\{sDom(ic)\}}{sDom(ic)}$ is admissible.
Instead, $\argu{\{sDom(jh)\}}{sDom(jh)}$ is not admissible, as it is
attacked by $\argu{\{notMet(jh,cheap)\}}{notSDom(jh)}$, which
cannot be attacked. 
\end{example}

In this example, admissible arguments in the strongly dominant ABA
framework and strongly dominant decisions in the ADF correspond.
This correspondence holds in general: 

\begin{theorem}
\label{SDABAThm}
Let $AF$ be the strongly dominant ABA framework corresponding to
$\nadf$. Then, for all decisions $d \in \decisions$, $d$ is
strongly dominant in
$\nadf$
\ifaf{} $\argu{\{sDom(d)\}}{sDom(d)}$ is
admissible in $AF$.
\end{theorem}

Thus, using the terminology of \cite{Cyras21}, strongly dominant ABA frameworks can be seen as \emph{complete} (argumentation-based) representations of the underlying decision problems. 
Dispute trees (of various types), drawn from these ABA frameworks, can then be used to explain argumentatively why decisions
are (strongly/weakly) dominant or not, as follows.

\begin{definition}
Let $AF$ be the strongly dominant ABA framework corresponding to
$\nadf$.
Then, for $d \in \decisions$:
  \begin{itemize}
    \item an {\em \dialogical\ explanation for $d$ being strongly dominant}
	    is a \leastAsm{} 
		  dispute tree 
	for $\argu{\{sDom(d)\}}{sDom(d)}$ in $AF$;
    \item an {\em \dialogical\ explanation for $d$ {\bf not} being strongly dominant}  is a best-effort 
	    dispute tree for $\argu{\{sDom(d)\}}{sDom(d)}$ in $AF$.
  \end{itemize}
\end{definition}

\begin{example}
	\label{exp2-dialx} (Example~\ref{exp2} continued.) 
	As an illustration, Figure~\ref{SDFig} uses a \leastAsm{} 
dispute
  tree 
  as a \dialogical\ explanation for $ic$ being strongly dominant in Example~\ref{exp:adf}. 
 The root of this tree puts forward the
claim (supported by \pro) that $ic$ is a strongly dominant decision. 
This statement is challenged by two arguments (from \opp): $ic$ is not
strongly dominant if it is not $cheap$ and if it is not $near$. These two
arguments are counter-attacked by further arguments (by \pro): $ic$ is
cheap and near, and thus strongly dominant. 
  This tree/explanation basically gives the reasons behind the
selection of $ic$ as strongly dominant. 
The explanation can be given an intuitive reading, for example,
in terms
of a debate between the (fictional) \emph{proponent} (\pro) and \emph{opponent}
(\opp) players, as follows: 
\begin{quote}
player {\bf P} starts by arguing that $ic$ is strongly
          dominant;
	  \\
player {\bf O} states that this is not so unless $ic$
          meets the goal $cheap$;
	 \\ 
	  player {\bf P} replies that $ic$ meets the goal
          $cheap$;
	  \\
	  player {\bf O} then goes back to saying that $ic$ is not strongly dominant if it does
	  not meet the goal $near$;
	  \\
	  player {\bf P} concludes saying that $ic$ meets the goal $near$. 
\end{quote}
The dialogue ends successfully for {\bf P}, with {\bf O} having nothing more to contribute.

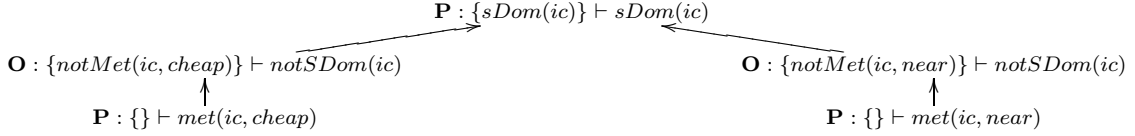
\begin{figure}[t]
\begin{scriptsize}
\[\xymatrix@1@=10pt{
& \pro: \argu{\{sDom(ic)\}}{sDom(ic)} \\
\opp: \argu{\{notMet(ic,cheap)\}}{notSDom(ic)} \ar[ru] & & \opp:
\argu{\{notMet(ic,near)\}}{notSDom(ic)} \ar[lu] \\ 
\pro: \argu{\{\}}{met(ic,cheap)} \ar[u] & & \pro: \argu{\{\}}{met(ic,near)}
\ar[u] }\] 
\end{scriptsize}
	\caption{\LeastAsm{} 
	dispute tree sanctioning
          $\argu{\{sDom(ic)\}}{sDom(ic)}$ as admissible (and $ic$ as
          strongly dominant), for Example~\ref{exp2-dialx}. 
} 
\label{SDFig}
\end{figure}

As a further illustration, Figure~\ref{fig:NSDFig} uses a best-effort 
dispute tree giving  an \dialogical\ explanation for $jh$ not being strongly
dominant in Example~\ref{exp:adf}, as an
argumentative reading of the selection of $jh$: unlike $ic$, since $jh$ does not meet
the goal $cheap$, the root argument $\argu{\{sDom(jh)\}}{sDom(jh)}$,
cannot defend the attack from $\argu{\{notMet(jh,cheap)\}}{notSDom(jh)}$.
 This can be 
	  read again as a dialogue between {\bf P} and {\bf O}, as follows: 
	  \begin{quote}
		  {\bf P} states that $jh$ is strongly dominant;
		  \\
		  {\bf O} replies that
          $jh$ is not unless it meets the goal $near$;
		  \\
		  {\bf P} states that $jh$
          meets $near$;\\ {\bf O} brings  a further objection that $jh$ is not strongly dominant if it
	  does not meet the goal $cheap$
	  \end{quote}
	  to which {\bf P} cannot find a reply, thus losing to {\bf O}.
\end{example}

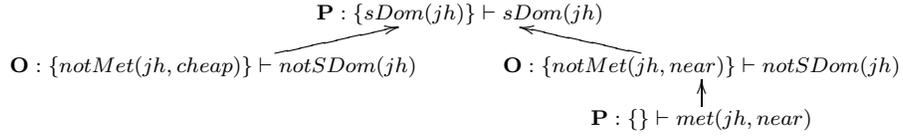
\begin{figure}[t]
\begin{scriptsize}
\[\xymatrix@1@=10pt@C=-40pt{
& \pro: \argu{\{sDom(jh)\}}{sDom(jh)} \\
\opp: \argu{\{notMet(jh,cheap)\}}{notSDom(jh)} \ar[ru] & & \opp:
\argu{\{notMet(jh,near)\}}{notSDom(jh)} \ar[lu] \\ 
& & \pro: \argu{\{\}}{met(jh,near)}
\ar[u] }\] 
\end{scriptsize}
	\caption{Best-effort 
	dispute tree sanctioning
          $\argu{\{sDom(jh)\}}{sDom(jh)}$ as not admissible (and $jh$ 
          as not strongly dominant) in Example~\ref{exp2-dialx}. 
} 
\label{fig:NSDFig}
\end{figure}

In addition to providing \dialogical\ explanations per-se, (various
types of) dispute trees can also be used to \emph{extract}
\flat\ explanations for decisions (not) meeting the various
decision criteria as defined in Definition~\ref{dfn:explanation}.  

\begin{proposition}
  \label{prop:SDExp}
  
  Let $AF$ be the strongly dominant ABA framework corresponding to
  \nadf.
  For any strongly dominant decisions $d \in \decisions$,
	let $\atree$ be  an \dialogical\ explanation for $d$ being strongly dominant (in the form of 
	\leastAsm{} dispute tree for 
	$\argu{\{sDom(d)\}}{sDom(d)}$ in $AF$). Then 
  \begin{center}
    $\{g|\pro:\argu{\{\}}{met(d,g)}$ is a leaf node in $\atree\}$
  \end{center}
	is a \flat{} explanation for $d$ being strongly dominant.
\end{proposition}


\begin{example}
  \label{exp:expSDTreeIN}

  (Example~\ref{exp2-dialx} continued.) Given the 
	\leastAsm{} dispute tree/ \dialogical\ explanation 
	in Figure~\ref{SDFig},
	it is immediate to see that arguments held by the leaf nodes,
  $\pro:\argu{\{\}}{met(ic,cheap)}$ and $\pro:\argu{\{\}}{met(ic,near)}$,
	give the {\flat} explanation $\{cheap, near\}$ for $ic$ being strongly dominant. 
		This
  can be read as:
    ``since $ic$ meets both goals $cheap$ and $near$, it is strongly
	dominant''. Whereas the \flat\ explanation refers to the components of the decision problem only, the \dialogical\ explanation unearths the reasoning behind $ic$ being strongly dominant. 
\end{example}

\begin{proposition}
  \label{prop:NSDExp}
  
  Let $AF$ be the strongly dominant ABA framework corresponding to
  $\nadf$.
  For any $d \in \decisions$, if $d$ is not strongly dominant,
	let $\atree$ be an \dialogical\ explanation for $d$ not being strongly dominant. Then, 
  \begin{center}
    $\{g|\opp:\argu{\{notMet(d,g)\}}{notSDom(d)}$ is a leaf node in
    $\atree\}$    
  \end{center}
	is a \flat{} explanation for $d$ not being strongly dominant.
\end{proposition}

\begin{example}
  \label{exp:expSDTree}
  
  (Example~\ref{exp2-dialx} continued.) Given the best-effort 
	dispute tree/\dialogical\ explanation
	in Figure~\ref{fig:NSDFig}, it is immediate to see that the  
  only leaf node labelled by O is
	$\argu{\{notMet(jh,cheap)\}}{notSDom(jh)}$. Thus, $\{cheap\}$ is a \flat{}
  explanation for $jh$ not being strongly dominant. This can be read as: 
    ``Since $jh$ does not meet the goal $cheap$, it is not strongly
	dominant''. Again, whereas the \flat\ explanation refers to the components of the decision problem only, the \dialogical\ explanation unearths the reasoning behind $ic$ not being strongly dominant.  
\end{example}

\subsubsection{
Dominant ABA frameworks and \dialogical\ explanations} 

Dominant decisions can also be given an
equivalent argumentative formulation, by defining a corresponding
dominant ABA framework 
also incorporating the core ABA framework, as follows.

\begin{definition}
  \label{dfn:dABA}

  For decisions $\decisions$ and goals $\goals$, 
	let
	the \emph{dominant component} be the ABA framework with

\begin{itemize}
\item
$\mc{R}_{d} =
\{notDom(d) \gets notMet(d,g) \ | \ d \in \decisions, g \in \goals\},$

\item
$\mc{A}_{d} = \{dom(d) | d \in \decisions\} \cup \{noOthers(d, g)| d
  \in \decisions, g \in \goals\} \cup \{notMet(d,g) | d \in
  \decisions,  g \in \goals \}$,

\item
for any $dom(d) \in \mc{A}_{d}$: $\mc{C}_{d}(dom(d)) = \{notDom(d)\}$,


for any $noOthers(d,g) \in \mc{A}_{d}$: $\mc{C}_{d}(noOthers(d,g)) =
\{met(d',g) | d' \in \decisions, d' \neq d\}$,

for any $notMet(d,g) \in \mc{A}_{d}$: $\mc{C}_{d}(notMet(d,g)) =
\{noOthers(d,g)\}$. 

\end{itemize}  

Then,
\label{dfn:dABADF}
the {\em dominant ABA framework corresponding to} $\fa$ is $\lracF$,
where:

\begin{itemize}
\item
  $\mc{R}=\Rcore \cup \mc{R}_{d}$, 

\item
  $\mc{A}=\Acore \cup \mc{A}_{d}$,

\item
  for all \asm{} in $\mc{A}$:
		\(\mc{C}(\asm) =
  \begin{cases}
    \Ccore(\asm)   & \quad \text{if } \asm \in \Acore \text{ and } \asm \notin \mc{A}_d,\\
    \mc{C}_d(\asm) & \quad \text{if } \asm \in \mc{A}_d \text{ and } \asm \notin \Acore,\\
    \Ccore(\asm) \cup \mc{C}_d(\asm) & \quad \text{if } \asm \in \mc{A}_d \cap \Acore.\\
  \end{cases}
		\)

%
%
\end{itemize}

\label{dominantABA}
\end{definition}

Intuitively, a decision $d$ is selected (as dominant) either if it
meets all goals, or for goals that $d$ does not meet, there is no
other $d'$ meeting them. Hence the contrary of $notMet(d,g)$
(representing that $d$ does not meet $g$) is either $met(d,g)$
(representing that $d$ meets $g$) or $noOthers(d,g)$ (representing
that
, although $d$ does not meet $g$, no other decision
does either). 

\begin{theorem}
\label{DABAThm}
Let $AF$ be the dominant ABA framework corresponding to
$\nadf$.
Then
for all decisions $d \in \decisions$, $d$ is dominant \ifaf{}
$\argu{\{dom(d)\}}{dom(d)}$ is admissible in $AF$.
\end{theorem}

\begin{example}
	\label{ex-aba-dom}
	(Example~\ref{expD} continued.)
As an illustration, given the dominant ABA framework 
corresponding to the ADF in Example~\ref{expD}, 
$jh$ is dominant since $\argu{\{dom(jh)\}}{dom(jh)}$ is
admissible. Indeed, the two counter-arguments 
$\argu{\{notMet(jh,g)\}}{notDom(jh)}$, for $g=near$ and $cheap$, 
are attacked, \respectively, by
argument $\argu{\{\}}{met(jh,near)}$ and argument 
$\argu{\{noOthers(ritz,cheap)\}}{noOthers(jh,cheap)}$, 
neither of which can be attacked.
\end{example}

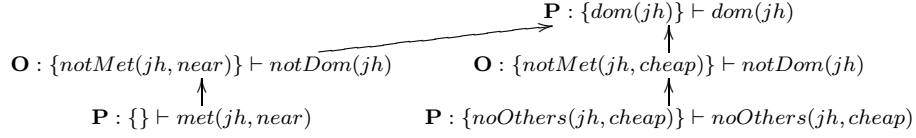
\begin{figure}[t]
\begin{scriptsize}
\[\xymatrix@1@=10pt{
& \pro: \argu{\{dom(jh)\}}{dom(jh)} \\
\opp: \argu{\{notMet(jh,near)\}}{notDom(jh)} \ar[ru] & \opp: \argu{\{notMet(jh,cheap)\}}{notDom(jh)} \ar[u] \\
\pro:  \argu{\{\}}{met(jh,near)} \ar[u]              & \pro: \argu{\{noOthers(jh,cheap)\}}{noOthers(jh,cheap)} \ar[u]
}\]
\end{scriptsize}
\caption{
	\LeastAsm{}
  dispute tree sanctioning 
$\argu{\{dom(jh)\}}{dom(jh)}$ as admissible (and $jh$ as dominant),
for Example~\ref{exp:DExp}. 
\label{DFig}
} 
\end{figure}

\begin{definition}
Let $AF$ be the dominant ABA framework corresponding to
$\nadf$.
Then, for $d \in \decisions$:
  \begin{itemize}
    \item an {\em \dialogical\ explanation for $d$ being dominant}
	    is a \leastAsm\ 
		  dispute tree 
	for $\argu{\{dom(d)\}}{dom(d)}$ in $AF$;
\item an {\em \dialogical\ explanation for $d$ {\bf not} being dominant}  is a best-effort 
	dispute tree for $\argu{\{dom(d)\}}{dom(d)}$ in $AF$.
  \end{itemize}
\end{definition}

\Flat{} explanations for decisions being dominant can also be extracted from
dispute trees, formally: 

\begin{proposition}
  \label{prop:DExp}
  
  Let $AF$ be the dominant ABA framework corresponding to
  $\nadf$.
  For
	any dominant $d \in \decisions$, let $\atree$ be an \dialogical\ explanation for $d$ being dominant (in the form of a \leastAsm{}
	dispute tree for $\argu{\{dom(d)\}}{dom(d)}$).
  Then the pair $\expPair{G}{F}$, where
  \begin{itemize}
    \item
      $G = \{g|\pro:\argu{\{\}}{met(d,g)}$ is a leaf node in $\atree\}$,
    \item
      $F = \{g|\pro:\argu{\{noOthers(d,g)\}}{noOthers(d,g)}$ is a
      leaf node in $\atree\}$, 
  \end{itemize}
	is a \flat{} explanation for $d$ being dominant.
\end{proposition}


\begin{example}
  \label{exp:DExp}
	(Example~\ref{ex-aba-dom} continued.) 
	Figure~\ref{DFig} shows an \dialogical\ explanation for $jh$ being dominant in the decision problem from Example~\ref{expD}.  We can see that since
  $\pro:\argu{\{\}}{met(jh,near)}$ and
  $\pro:\argu{\{noOthers(jh,cheap)\}}{noOthers(jh,cheap)}$ are the
	two proponent leaf nodes in the tree, we obtain, as $G$ and $F$ in Proposition~\ref{prop:DExp}, $G=\{near\}$ and
  $F=\{cheap\}$. Thus, a
	\flat{} explanation for $jh$ being dominant is
  \expPair{\{near\}}{\{cheap\}}, which can be read as:
    ``$jh$ is dominant because it meets the goal $near$
	and the for the goal $cheap$ which $jh$ does not meet, no other decision meets it either''. Instead, the \dialogical\ explanation in Figure~\ref{DFig} can be read, for example, as the following dialogue providing the reasons behind $jh$ being dominant: 
	\begin{quote}
		{\bf P}: $jh$ is dominant;
		\\
		{\bf O}: $jh$ is not unless it meets the goal $near$;
		\\
		{\bf P}: $jh$ meets $near$;
		\\
		{\bf O:} $jh$ is not dominant unless it meets $cheap$;
		\\
		{\bf P:} $jh$ does not meet $cheap$, but no other decision meets $cheap$ either.
	\end{quote}
\end{example}

\begin{proposition}
  \label{prop:NDExp}
  
  Let $AF$ be the dominant ABA framework corresponding to
  $\nadf$.
  For any $d \in \decisions$, if $d$ is not dominant,
	let $\atree$ be an \dialogical\ explanation for $d$ not being dominant. Then 
  \begin{center}
    $\{\expPair{d_i}{g_j} | \opp:\argu{\{\}}{met(d_i,g_j)}$
    is a leaf node in $\atree \}$
  \end{center}
	is a \flat{} explanation for $d$ not being dominant.
\end{proposition}

\begin{example}
  \label{exp:expDTree}
  
	(Example~\ref{ex-aba-dom} continued.) A best-effort 
	dispute
  tree for the argument $\argu{\{dom(ritz)\}}{dom(ritz)}$ is shown in
	Figure~\ref{fig:domRITZ}, providing an \dialogical\ explanation for $ritz$ not being dominant in the decision problem from Example~\ref{expD}. From this 
	\dialogical\ explanation, we can see that
  $\argu{\{\}}{met(jh,near)}$ is the only argument in an \opp{} leaf
	node. Thus, $\{\expPair{jh}{near}\}$ is a \flat{} explanation for $ritz$ not 
  being dominant. This can be read as:
    ``$ritz$ is not dominant because it does not meet the goal $near$ yet $jh$ meets it''.
	As in the previous illustrations, the \dialogical\ explanation can be given a dialogical reading indicating the reasoning leading to sanctioning $ritz$ as not being dominant.
\end{example}

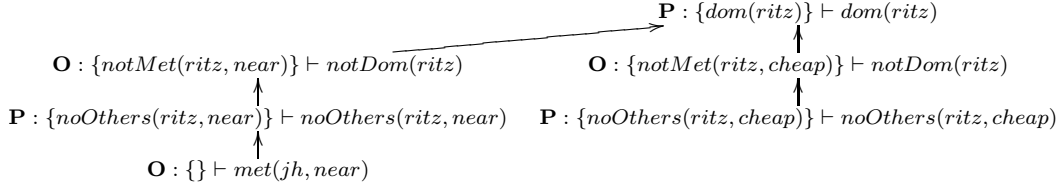
\begin{figure}[t]
\begin{scriptsize}
\[\xymatrix@1@=10pt{
& \pro: \argu{\{dom(ritz)\}}{dom(ritz)} \\
\opp: \argu{\{notMet(ritz,near)\}}{notDom(ritz)} \ar[ru] & \opp: \argu{\{notMet(ritz,cheap)\}}{notDom(ritz)} \ar[u] \\
\pro: \argu{\{noOthers(ritz,near)\}}{noOthers(ritz,near)} \ar[u] & \pro: \argu{\{noOthers(ritz,cheap)\}}{noOthers(ritz,cheap)} \ar[u] \\
\opp: \argu{\{\}}{met(jh,near)} \ar[u]                                                              
}\]
\end{scriptsize}
\caption{
	Best-effort dispute tree for
  $\argu{\{dom(ritz)\}}{dom(ritz)}$ in Example~\ref{exp:expDTree}.
\label{fig:domRITZ}
} 
\end{figure}

\subsubsection{
Weakly dominant ABA frameworks and \dialogical\ explanations} 
As for strongly dominant and dominant decisions, ABA can be used to 
explain weakly dominant decisions. 

\begin{definition}
\label{dfn:wdABA}

For decisions $\decisions$ and goals $\goals$, let
	the \emph{weakly dominant component} be the ABA framework with

\begin{itemize}
\item
$\mc{R}_{w} =  \{ notWDom(d) \gets met(d',g), notMet(d,g),
  notMore(d,d') | d,d' \in \decisions, d\neq d', g \in \goals\} \ \cup
  \ \{ more(d,d') \gets met(d,g), notMet(d',g) | $ 
$d,d' \in \decisions, d\neq d', g \in \goals\}$,

\item
$\mc{A}_{w} = \{wDom(d) | d \in \decisions\} \cup 
\{notMore(d,d') | d,d' \in \decisions, d \neq d'\}$,

\item
for any $wDom(d) \in \mc{A}_{w}$: $\mc{C}_{w}(wDom(d)) =
\{notWDom(d)\}$, 

for any $notMore(d,d')\in \mc{A}_{w}$: $\mc{C}_{w}(notMore(d,d')) =
\{more(d,d')\}$. 
\end{itemize}

Then,
\label{dfn:wdABADF}
the {\em weakly dominant ABA framework
  corresponding to} 
$\nadf$
is $\lracF$, where

\begin{itemize}
\item
$\mc{R} = \Rcore \cup \mc{R}_{w}$,

\item
  $\mc{A} = \Acore \cup \mc{A}_{w}$,

\item
for all \asm{} in $\mc{A}$,
		\(\mc{C}(\asm) =
  \begin{cases}
    \Ccore(\asm)   & \quad \text{if } \asm \in \Acore,\\
    \mc{C}_w(\asm) & \quad \text{if } \asm \in \mc{A}_w.\\
  \end{cases}
		\)
  
%
%
\end{itemize}
\end{definition}

Intuitively, a decision $d$ is selected (as weakly dominant) 
\ifaf{} there is no other decision $d'$ such that the set of goals
$d'$ meets is a superset of the set of goals $d$ meets. This can be
tested for all $d' \neq d$ by definition of contrary of $wDom(d)$, as
$notWDom(d)$, and by the rule with head such a contrary.  
Indeed, if we assume $d$ to be weakly dominant, to attack it,
one needs to find another decision $d'$ and a goal $g$ such that
$met(d',g)$, $notMet(d,g)$, and $notMore(d,d')$ all
hold. Also, to attack assumption $notMore(d,d')$, 
one needs to show that there exists some goal $g'$, such that
$met(d',g')$ and $notMet(d,g')$.

\begin{theorem}
\label{WDABAThm}
Let $AF$ be the weakly dominant ABA framework corresponds to
$\nadf$.
Then for all decisions $d \in \decisions$, $d$ is dominant
\ifaf{} $\argu{\{wdDom(d)\}}{wdDom(d)}$ is admissible in $AF$. 
\end{theorem}

\begin{example}
	\label{ex-aba-wdom}
	(Example~\ref{expWD} continued.)
As an illustration, given the weakly dominant ABA framework
corresponding to  the ADF in Example~\ref{expWD}, 
$ic$ is weakly dominant since 
$\argu{\{wDom(ic)\}}{wDom(ic)}$ is admissible. 
Indeed, the (two) arguments 
	$A$ and $B$ and the argument 
	$C$, all attacking
  $\argu{\{wDom(ic)\}}{wDom(ic)}$, are all counter-attacked (the first
  by argument $\argu{\{notMet(jh,cheap)\}}{more(ic,jh)}$, and the
  latter two by $\argu{\{\}}{met(ic,clean)}$, none of which can be
  attacked.)
\end{example}
 
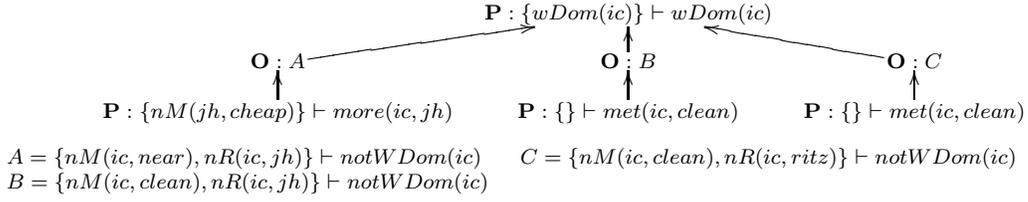
\begin{figure}[t]
\begin{scriptsize}
\[\xymatrix@1@=10pt{
& \pro: \argu{\{wDom(ic)\}}{wDom(ic)} \\
\opp: A \ar[ru] & \opp: B \ar[u] & \opp: C \ar[lu] \\
\pro: \argu{\{nM(jh,cheap)\}}{more(ic,jh)} \ar[u] & \pro:
\argu{\{\}}{met(ic,clean)}\ar[u] & \pro: \argu{\{\}}{met(ic,clean)}\ar[u] 
}\]
\begin{tabular}{ll}
  $A = \argu{\{nM(ic,near), nR(ic,jh)\}}{notWDom(ic)}$  & $C = \argu{\{nM(ic,clean), nR(ic,ritz)\}}{notWDom(ic)}$\\
  $B = \argu{\{nM(ic,clean), nR(ic,jh)\}}{notWDom(ic)}$ \\
\end{tabular}
\end{scriptsize}
\caption{
	\LeastAsm{} dispute tree sanctioning $\argu{\{wDom(ic)\}}{wDom(ic)}$ as admissible
(and $ic$ as weakly dominant) in Example~\ref{exp:sec5WD}. Here,
$nM$ and 
$nR$ are short-hands for $notMet$ and $notMore$, \respectively.
\label{WDFig}
} 
\end{figure}

\begin{definition}
Let $AF$ be the weakly dominant ABA framework corresponding to
$\nadf$.
Then, for $d \in \decisions$:
  \begin{itemize}
    \item an {\em \dialogical\ explanation for $d$ being weakly dominant}
	    is a \leastAsm\ 
		  dispute tree 
	for $\argu{\{wDom(d)\}}{wDom(d)}$ in $AF$;
\item an {\em \dialogical\ explanation for $d$ {\bf not} being weakly dominant}  is a best-effort 
	dispute tree for $\argu{\{wDom(d)\}}{wDom(d)}$ in $AF$.
  \end{itemize}
\end{definition}

Leaf nodes in 
dispute trees also give \flat{} explanations
for weakly dominant decisions, formally:

\begin{proposition}
  \label{prop:WDExp}

  Let $AF$ be the weakly dominant ABA framework corresponding to
$\nadf$.
  For any weakly dominant $d \in \decisions$, let
	$\atree$ be an \dialogical\ explanation for $d$ being weakly dominant (in the form of a \leastAsm\ 
	dispute tree for
	$\argu{\{wDom(d)\}}{wDom(d)}$). Then the pair $\expPair{G}{M}$, where
  \begin{itemize}
    \item
  $G=\{g|\pro:\argu{\{\}}{\{met(d,g)\}}$ or
  $\pro:\argu{\{notMet(\_,g)\}}{more(d,\_)}$ is a node in $\atree\}$ and 

      \item
  $M=\{(d',g)|\pro:\argu{\{notMet(d',g)\}}{more(d,d')}$ is a leaf node
        in $\atree\}$
  \end{itemize}
	is a \flat{} explanation for $d$ being weakly dominant.
\end{proposition}

\begin{example}
  \label{exp:sec5WD}
	(Example~\ref{ex-aba-wdom} continued.)
	Figure~\ref{WDFig} shows an \dialogical\ explanation
for $ic$ being weakly dominant in the decision problem from
	Example~\ref{expWD}. Here,
we can see that the leaf nodes are $\pro:
  \argu{\{notMet(jh,cheap)\}}{more(ic,jh)}$, $\pro:
  \argu{\{\}}{met(ic,clean)}$ and $\pro:
  \argu{\{\}}{met(ic,clean)}$. Thus,
    $\expPair{\{clean, cheap\}}{\{(jh,cheap)\}}$
	is a \flat{} explanation for 
	$ic$ being weakly dominant, which 
	can be read as ``A reason for $ic$ being weakly dominant is that $ic$ meets goals $clean$ and $cheap$, in which $cheap$ is not met by $jh$''.
\end{example}

\begin{proposition}
  \label{prop:NWDExp}
  
  Let $AF$ be the weakly dominant ABA framework corresponding to
$\nadf$.
  For any $d \in \decisions$, if $d$ is not weakly dominant,
	let $\atree$ be an \dialogical\ explanation for $d$ not being dominant (in the form of a best-effort 
	dispute tree for
	$\argu{\{wDom(d)\}}{wDom(d)}$). Then the set
  \begin{itemize}
  \item
    $\{d'|\opp:\argu{\{\}}{met(d',g)}$ a leaf node in $\atree$ or 
    $\opp:\argu{\{notMet(d,g),notMore(d,d')\}}{notWDom(d)}$ a leaf node
    in $\atree \}$
  \end{itemize}
	is a \flat{} explanation for $d$ not being weakly dominant.
\end{proposition}


\begin{example}
  \label{exp:expWDTree} 

	(Example~\ref{ex-aba-wdom} continued.)  An \dialogical\ 
explanation for $ritz$ not being weakly
	dominant in the decision problem of Example~\ref{expWD} is given in 	
  Figure~\ref{fig:wdomRitz}. Here, both $\argu{\{\}}{met(jh,clean)}$
  and $\argu{\{\}}{met(ic,clean)}$ are held by leaf nodes labelled by
	O. Thus $\{ic, jh\}$ is a \flat{} explanation for $ritz$ not being weakly
  dominant, which 
	can be read as: ``$ritz$ is not weakly dominant as $jh$ and $ic$ meet all goals met by $ritz$ and more''.
\end{example}

\begin{figure}
\begin{scriptsize}
\[\xymatrix@1@R=10pt@C=3pt{
& \pro: \argu{\{wDom(ritz)\}}{wDom(ritz)} \\
  \opp: A \ar[ru] & \opp: B \ar[u] & \opp: C \ar[lu] & \opp: D \ar[llu] \\
  \pro: E \ar[u]  & \pro: F \ar[u] & \pro: \argu{\{\}}{met(ritz,clean)} \ar[u] & \pro: \argu{\{\}}{met(ritz,clean)} \ar[u] \\
  \opp: \argu{\{\}}{met(jh,clean)} \ar[u] & \opp: \argu{\{\}}{met(ic,clean)} \ar[u] 
}\]
\begin{tabular}{ll}
$A = \argu{\{nM(ritz,near),nR(ritz,jh)\}}{nW(ritz)}$ &
$B = \argu{\{nM(ritz,cheap),nR(ritz,ic)\}}{nW(ritz)}$ \\
$C = \argu{\{nM(ritz,clean),nR(ritz,jh)\}}{nW(ritz)}$ &
$D = \argu{\{nM(ritz,clean),nR(ritz,ic)\}}{nW(ritz)}$ \\
$E = \argu{\{nM(jh,clean)\}}{more(ritz,jh)}$ &
  $F = \argu{\{nM(ic,clean)\}}{more(ritz,ic)}$
  \end{tabular}
  \end{scriptsize}
\caption{
A 
	best-effort dispute tree for $\argu{\{wDom(ritz)\}}{wDom(ritz)}$ for
Example~\ref{exp:expWDTree}. Here, $nM$, $nR$ and $nW$ are
short-hands for $notMet$, $notMore$ and $notWDom$, respectively.
  \label{fig:wdomRitz}
} 
\end{figure}
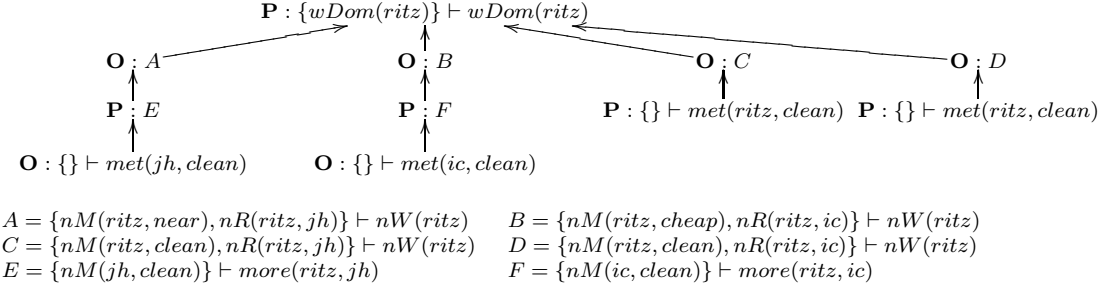

\subsection{\Dialogical\ Explanations for  PDFs}
\label{sec:ABADecP}

The problem of determining \pfrGoal{} decisions in a PDF can also be
equivalently understood as the problem of determining admissible
arguments in an ABA framework, defined as follows:

\begin{definition}
  \label{dfn:pfrGoalSet}

  Let $\cgset$ be the comparable goal set in $\fe = \eadfps$, the {\em
    \pfrSet{} component} is an ABA framework \lracFPG{}, where 

  \begin{itemize}
  \item
    $\mc{R}_p =
    \{pfr(s_t, s_r) \gets | s_t, s_r \in \cgset,
    s_r \preferenceSetstrict s_t \} \cup$ \\
    $\{\notMetSet(d,s) \gets notMet(d,g) | d \in
    \decisions, s \in \cgset, g \in s \}\cup$  \\
    $\{better(d,d',s) \gets \metSet(d,s'), \notMetSet(d',s')
    ,pfr(s',s) | d,d' \in \decisions, d\neq d', s,s' \in
    \cgset, s\neq s' \} \cup$; \\
    $\{\notPS(d) \gets \metSet(d',s), \notMetSet(d,s),
    notBetter(d,d',s) | d,d' \in \decisions, d \neq d', s \in \cgset
    \}$; 
   
  \item
    $\mc{A}_p = \{ \pS(d),  \metSet(d, s), notBetter(d, d', s)  | \;
    d,d' \in \decisions, d \neq d', s \in \cgset\}$;  

  \item
    for any  $\pS(d)\in \mc{A}$:  $\mc{C}_p(\pS(d)) = \{\notPS(d)\}$;

    for any $\metSet(d, s)\in \mc{A}$: $\mc{C}_p(\metSet(d,s)) =
    \{\notMetSet(d,s)\}$; 

    for any $notBetter(d, d', s) \in \mc{A}$:
    $\mc{C}_p(notBetter(d,d',s)) = \{better(d,d',s)\}$.
  \end{itemize}
\end{definition}

Rules in a \pfrSet{} component can be read as follows.
\begin{itemize}
\item
  $pfr(s_t, s_r) \gets$ represents that a comparable goal set $s_t$ is preferred
  to another comparable goal set $s_r$.
  
\item
  $\notMetSet(d,s) \gets notMet(d,g)$ for $g \in s$
  states that a decision $d$ does not meet a comparable goal set $s$
  if $d$ does not meet some goal $g$ in $s$.

\item
$better(d,d',s) \gets \metSet(d,s'), \notMetSet(d',s'), pfr(s',s)$
  states that decision $d$ is better than $d'$ \wrt{} a comparable
  goal set $s$ if $d$ meets some goal set $s'$ that is not met by $d'$
  and $s'$ is preferred to $s$.  

\item
$\notPS(d) \gets \metSet(d',s), \notMetSet(d,s), notBetter(d,d',s)$
  states that decision $d$ is not \pfrSet{} if (1) there exists
  another decision $d'$ meeting all goals in a comparable goal set
  $s$, (2) $d$ does not meet all goals in $s$ and (3) comparing with
  $d'$, $d$ does not meet any more preferred goal set than $s$.
\end{itemize}
All decisions are assumed to be \pfrSet{} as $\pS(d)$ is an assumption
in the ABA framework for all decision $d$. Then all such assumptions
are examined by arguments for $\notPS(d)$, constructed with relevant
rules and assumptions. 

With the \pfrSet{} component defined, we are ready to obtain the ABA
framework for \pfrSet{} decisions by taking the ``union'' of the
\pfrSet{} component and the core ABA framework. 

\begin{definition}
\label{dfn:pfrGoalSetABA}
Let $\fe = \eadfps$, $\lracFPG{}$ the \pfrSet{} component and let
$\lracFcore$ be the core ABA framework corresponding to
$\nadf$.
Then,
the {\em \pfrSet{} ABA framework corresponding to} $\fe$ is $\lracF$
where:

\begin{itemize}
\item
$\mc{R}= \Rcore \cup \mc{R}_p$;

\item
$ \mc{A}=\Acore \cup \mc{A}_p$; 

\item
for all \asm{} in $\mc{A}$:
		\(\mc{C}(\asm) =
  \begin{cases}
    \Ccore(\asm)   & \quad \text{if } \asm \in \Acore,\\
    \mc{C}_p(\asm) & \quad \text{if } \asm \in \mc{A}_p.\\
  \end{cases}
		\)
 
%
%
%
\end{itemize}
\end{definition}

The definition of \pfrSet{} ABA framework (corresponding to $\fe$) 
is given in the same spirit as for the definitions of corresponding
ABA frameworks given earlier for (strongly/wea\-kly) dominant
decisions, to guarantee an argumentative reading of  \pfrGoal{}
decisions, as follows:  

\begin{theorem}
\label{thmTran2}
Let $AF$ be the \pfrSet{} ABA framework corresponding to $
\eadfps$. Then, for all $d \in \decisions$, $d$ is \pfrGoal{} 
\ifaf{} $\argu{\{\pS(d)\}}{\pS(d)}$ is admissible in $AF$.
\end{theorem}

The following example illustrates the argumentative re-interpretation 
of \pfrSet{} decisions.

\begin{table}
\scriptsize
\begin{center}
\begin{tabular}{|r|cc|}
\hline
      & $jh$ & $ic$ \\
\hline
\rowcolor{Gray}
cheap &        0          &        1                \\ 
near  &        1          &        1                \\
\rowcolor{Gray}
quiet  &        1          &        0               \\
\hline
\end{tabular}
\end{center}
\caption{\Tdg{} 
  for Example~\ref{exp:pfrSet}.\label{FTtable}}
\end{table}

\begin{example}
\label{exp:pfrSet}
Consider $ \eadfps$ with $\Tdg$ in Table~\ref{FTtable} and 
$\preferenceSet$ such that $$\{quiet\} \preferenceSetstrict \{cheap\}
\preferenceSetstrict \{near\} \preferenceSetstrict \{quiet, near\} $$ 
The comparable goal set is $\cgset= \{\{quiet, near\},$ $\{near\},
\{cheap\}, \{quiet\}\}$. Let $qnS,nS,cS$ and $qS$ stand for the
elements of $\cgset$ (\respectively).

Also, we use $b$, $nB$, $nM$ and $nS$ as short-hands for $better$,
$notBetter$, $notMet$ and $\notMetSet$, \respectively. Then
\Rcore{} consists of

\begin{tabular*}{0.95\textwidth}{@{\extracolsep{\fill} } l l l l}
$met(ic,cheap) \gets$    & $met(ic,near) \gets$  & $met(jh,near) \gets$   & $met(jh,quiet) \gets$  \\
\end{tabular*}

\noindent
$\mc{R}_p$ consists of


$\{\notPS(ic) \gets \metSet(jh,s), nS(ic,s),
nB(ic,jh,s) | s \in \{qnS, nS, cS, qS\}\}$; 

$\{\notPS(jh) \gets \metSet(ic,s), nS(jh,s),
nB(jh,ic,s) | s \in \{qnS, nS, cS, qS\}\}$; 


$\{nS(ic,s) \gets nM(ic,g) | g \in
\{near,quiet, cheap\}, s \in \{qnS, nS, cS, qS\}, g \in s\} $; 


$\{nS(jh,s) \gets nM(jh,g) | g \in
\{near,quiet, cheap\}, s\in \{qnS, nS, cS, qS\}, g \in s\} $; 

$\{b(d,d',qS) \!\gets \! \metSet(d,s'), \!nS(d',s')| d
\!\in \!\{ic,jh\}, d'\! \in \!\{ic,jh\} \!\setminus \!\{d\}, s'
\!\in\!  \{cS, nS, qnS\} 
\} $; 

$\{b(d,d',cS) \!\gets \! \metSet(d,s'), \!nS(d',s')| d \!\in \!\{ic,jh\}, d'\! \in \!\{ic,jh\} \!\setminus \!\{d\}, s' \!\in\! 
\{nS, qnS\} 
\} $; 

$\{b(d,d',nS) \!\gets \! \metSet(d,qnS), \!nS(d',qnS)| d \!\in \!\{ic,jh\}, d'\! \in \!\{ic,jh\} \!\setminus \!\{d\}
\} $; 

\noindent\mc{A} consists of the following elements, with \mc{C} as in Definition~\ref{dfn:pfrGoalSetABA}:

\begin{tabular*}{0.95\textwidth}{@{\extracolsep{\fill} }llll}
$\pS(ic)$         & $nM(ic,near)$ & $nM(ic,cheap)$  & $nM(ic,quiet)$ \\
$\pS(jh)$         & $nM(jh,near)$ & $nM(jh,cheap)$  & $nM(jh,quiet)$ \\
$\metSet(ic,qnS)$ & $\metSet(ic,cS)$   & $\metSet(ic,qS)$     & $\metSet(ic,nS)$ \\
$\metSet(jh,qnS)$ & $\metSet(jh,cS)$   & $\metSet(jh,qS)$     & $\metSet(jh,nS)$ \\
$nB(ic,jh,qnS)$ & $nB(ic,jh,cS)$ & $nB(ic,jh,nS)$ & $nB(ic,jh,qS)$ \\
$nB(jh,ic,qnS)$ & $nB(jh,ic,cS)$ & $nB(jh,ic,nS)$ & $nB(jh,ic,qS)$ \\
\end{tabular*}

\medskip

In this example, it is easy to see that $\argu{\{\pS(jh)\}}{\pS(jh)}$
is admissible, as well as the root of the admissible (\leastAsm) dispute tree in
Figure~\ref{fig:pfrSetExp}. The root argument claims that $jh$ is a
\pfrSet{} decision, and the tree provides an \dialogical\ explanation, in the same spirit as for other decision criteria earlier in this section. This explanation can be provided, for example, the following intuitive reading. The root argument is attacked by five arguments, $A-E$
in the tree, stating various reasons against $jh$ being \pfrSet. For
instance, argument $A$ can be read as the following objection 
\begin{quote}
$jh$ is not a \pfrSet{} decision if the alternative decision $ic$
  meets both goals $near$ and $quiet$, $jh$ does not meet $near$, and
  $jh$ does not meet more preferred goals than $near$ and $quiet$. 
\end{quote}
Each of the attacking arguments $A-E$ is counterattacked, by some
argument which is a leaf of the tree. For instance, the leaf argument 
$\argu{\{\}}{met(jh,near)}$, counter-attacking both $A$ and $C$, can
be read as 
\begin{quote}
but it is a fact that $jh$ meets goal $near$.
\end{quote}
The argument 
$F = \argu{\{\metSet(jh,qnS), notMet(ic,quiet)\}}{better(jh,ic,cS)}$
can be read as 
\begin{quote}
$jh$ is better than $ic$ as it meets more preferred goals than
  $cheap$, since $jh$ meets both $near$ and $quiet$ (which are
  together more preferred than $cheap$ alone, as indicated in the
  specification of the decision problem) yet $ic$ does not meet
  $quiet$. 
\end{quote}
For this same example, Figure~\ref{fig:NpfrSetExp} gives a maximal (best-effort) dispute tree with root $\argu{\{\pS(ic)\}}{\pS(ic)}$,
showing that this argument is non-admissible. 
This tree in provides an \dialogical\ explanation, which can be intuitively read dialogically as follows, for example:
\begin{quote}
	the proponent {\bf P} claims that $ic$
          is \pfrSet;
	  \\
	  this is challenged by {\bf O} putting
          forward arguments that $ic$ does not meet the goal $near$,
          which are in the set $\{quiet, near\}$ met by $jh$, 
          and $ic$ does not meet more preferred goals than the set
          $\{quiet,near\}$;
	  \\
	  {\bf P} challenges that $jh$ does not meet
          $quiet$, so it does not meet the set $\{quiet, near\}$. \\
	  {\bf
          O} confirms that $jh$ indeed meets $quiet$. (Left-most
          branch of the tree.) Then the debate moves to whether $ic$
          meets $quiet$, etc.
\end{quote}
\end{example}

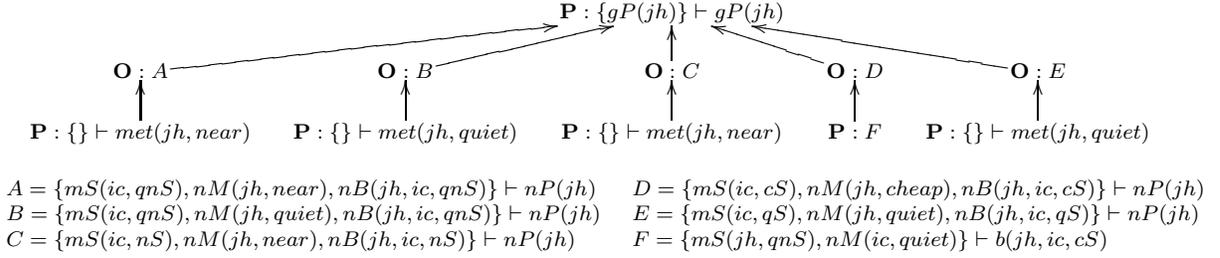
\begin{figure}
\begin{scriptsize}
\[\xymatrix@1@=14pt{
             &             & \pro:\argu{\{\pS(jh)\}}{\pS(jh)} \\
\opp:A \ar[urr] & \opp:B \ar[ur] & \opp:C \ar[u] & \opp:D \ar[ul] & \opp:E \ar[ull]   \\
\pro:\argu{\{\}}{met(jh,near)} \ar[u] & \pro:\argu{\{\}}{met(jh,quiet)} \ar[u] & \pro:\argu{\{\}}{met(jh,near)} \ar[u] & \pro:F \ar[u] & \pro:\argu{\{\}}{met(jh,quiet)} \ar[u] }\]
\hspace{-10pt}\begin{tabular}{ll}
$A=\argu{\{mS(ic,qnS),nM(jh,near),nB(jh,ic,qnS)\}}{nP(jh)}$&
$D=\argu{\{mS(ic,cS),nM(jh,cheap),nB(jh,ic,cS)\}}{nP(jh)}$\\
$B=\argu{\{mS(ic,qnS),nM(jh,quiet),nB(jh,ic,qnS)\}}{nP(jh)}$&
$E=\argu{\{mS(ic,qS),nM(jh,quiet),nB(jh,ic,qS)\}}{nP(jh)}$\\
$C=\argu{\{mS(ic,nS),nM(jh,near),nB(jh,ic,nS)\}}{nP(jh)}$&
$F=\argu{\{mS(jh,qnS), nM(ic,quiet)\}}{b(jh,ic,cS)}$\\
\end{tabular}
\end{scriptsize}
	\caption{\LeastAsm\ 
	dispute tree for Example~\ref{exp:pfrSet}, showing that 
  $\argu{\{\pS(jh)\}}{jh}$ is admissible. Here, $mS$, $nM$, $nB$ and
  $nP$ are short-hands for $\metSet$, $notMet$, $notBetter$,
  and $\notPS$ \respectively.} 
\label{fig:pfrSetExp}
\end{figure}

This example illustrates how \leastAsm\ 
dispute trees (as in Figure~\ref{fig:pfrSetExp}) and best-effort 
dispute tree (as in Figure~\ref{fig:NpfrSetExp}) can be used as \dialogical\ explanations for \pfrSet{} decisions. Formally:

\begin{definition}
Let $AF$ be the 
	\pfrSet{} ABA framework corresponding to $ \eadfps$. 
  Then, for $d \in \decisions$:
  \begin{itemize}
	  \item an {\em \dialogical\ explanation for $d$ being \pfrGoal} is a \leastAsm\ 
		  dispute tree 
	for $\argu{\{\pS(d)\}}{\pS(d)}$ in $AF$;
\item an {\em \dialogical\ explanation for $d$ {\bf not} being \pfrGoal}  is a best-effort 
	dispute tree for $\argu{\{\pS(d)\}}{\pS(d)}$ in $AF$.
  \end{itemize}
\end{definition}

As in ADFs, in addition to using 
dispute tree to explain
decisions, we can explicitly extract, from these trees, \flat{} explanations for
\pfrSet{} decisions. 

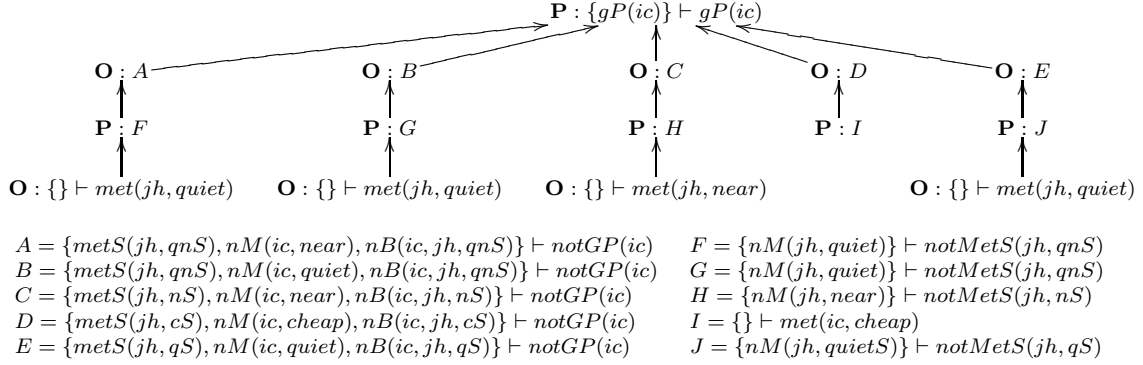
\begin{figure}
\begin{scriptsize}
\[\xymatrix@1@=14pt{
               &             & \pro:\argu{\{\pS(ic)\}}{\pS(ic)} \\
  \opp:A \ar[urr] & \opp:B \ar[ur] & \opp:C \ar[u] & \opp:D \ar[ul] & \opp:E \ar[ull] \\
  \pro:F \ar[u]   & \pro:G \ar[u]  & \pro:H \ar[u] & \pro:I \ar[u]  & \pro:J \ar[u] \\
  \opp:\argu{\{\}}{met(jh,quiet)} \ar[u] & \opp:\argu{\{\}}{met(jh,quiet)} \ar[u] & \opp:\argu{\{\}}{met(jh,near)} \ar[u] & & \opp:\argu{\{\}}{met(jh,quiet)} \ar[u] }\]

\begin{tabular}{ll}
  $A = \argu{\{\metSet(jh,qnS),nM(ic,near),nB(ic,jh,qnS)\}}{\notPS(ic)}$&
  $F = \argu{\{nM(jh,quiet)\}}{\notMetSet(jh,qnS)}$\\
  $B = \argu{\{\metSet(jh,qnS),nM(ic,quiet),nB(ic,jh,qnS)\}}{\notPS(ic)}$&
  $G = \argu{\{nM(jh,quiet)\}}{\notMetSet(jh,qnS)}$\\
  $C = \argu{\{\metSet(jh,nS),nM(ic,near),nB(ic,jh,nS)\}}{\notPS(ic)}$&
  $H = \argu{\{nM(jh,near)\}}{\notMetSet(jh,nS)}$\\
  $D = \argu{\{\metSet(jh,cS),nM(ic,cheap),nB(ic,jh,cS)\}}{\notPS(ic)}$&
  $I = \argu{\{\}}{met(ic,cheap)}$\\
  $E = \argu{\{\metSet(jh,qS),nM(ic,quiet),nB(ic,jh,qS)\}}{\notPS(ic)}$&
  $J = \argu{\{nM(jh,quietS)\}}{\notMetSet(jh,qS)}$
\end{tabular}
\end{scriptsize}
	\caption{Best-effort 
	 dispute tree for Example~\ref{exp:pfrSet}, showing that 
  $\argu{\{\pS(ic)\}}{ic}$ is not admissible. Here, $nM$ and $nB$ are
  short-hands for $notMet$ and $notBetter$, \respectively.
\label{fig:NpfrSetExp}
}
\end{figure}

\begin{proposition}
    \label{prop:expPfrSet}

  Let $AF$ be the \pfrSet{} ABA framework corresponding to
  $\eadfps$. For any  \pfrSet\ $d \in \decisions$, let
  $\atree$ be 
	an \dialogical\ explanation for $d$ being \pfrSet. Then $\expPair{G}{M}$, where
  \begin{itemize}
  \item
    $G=\{g|\pro:\argu{\{\}}{\{met(d,g)\}}$ is a leaf node in
    $\atree\}$ 
  \item
    $M =
    \{(S,d')|\pro:\argu{\{\metSet(d,S),notMet(d',\_)\}}{better(d,d',\_)}$
    is a leaf node in $\atree\}$
  \end{itemize}
	is a \flat{} explanation for $d$ being \pfrSet{}.
\end{proposition}

\begin{example}
  \label{exp:expPfrSet}
	(Example~\ref{exp:pfrSet} continued.)
	For the \dialogical\ explanation for $jh$ being \pfrSet\ in
	Figure~\ref{fig:pfrSetExp}, the leaf nodes are
  $\pro: \argu{\{\metSet(jh,qnS), notMet(ic,quiet)\}}{better(jh,ic,cS)}$,
  $\pro: \argu{\{\}}{met(jh,near)}$ and
  $\pro: \argu{\{\}}{met(jh,quiet)}$.
  Thus, 
  \begin{center}
    $\expPair{\{near, quiet\}}{\{(ic,qnS)\}}$
  \end{center}
	is a \flat{} explanation for $jh$ being \pfrSet,  which can be read as ``$jh$ is \pfrSet{} as it meets $near$ and $quiet$, and the goal set $\{near, quiet\}$ is more preferred than any set of goals met by $ic$''.
\end{example}

\begin{proposition}
    \label{prop:expNPfrSet}

  Let $AF$ be the \pfrSet{} ABA framework corresponding to
  $\eadfps$. For any $d \in \decisions$, if $d$ is not \pfrSet, let
  $\atree$ be 
	an \dialogical\ explanation for $d$ not being \pfrSet. Then
  \begin{center}
    $\{d'|\opp:\argu{\{\}}{met(d',g)}$ a leaf node in $\atree\}$
  \end{center}
	is a \flat{} explanation for $d$ not being \pfrSet{}.
\end{proposition}

\begin{example}
  \label{exp:expNPfrSet}
 	(Example~\ref{exp:pfrSet} continued.)
	For the \dialogical\ explanation for $ic$ not being \pfrSet\ in 
  Figure~\ref{fig:NpfrSetExp}, we can see that
  $\argu{\{\}}{met(jh,near)}$ and $\argu{\{\}}{met(jh,quiet)}$ label
	\opp{} leaf nodes, thus, $\{jh\}$ is a \flat{} explanation
        for $ic$ not being \pfrSet{}, which can be read as: ``A reason
        for $ic$ not being \pfrSet{} is that $jh$ meets goals 
        that are more preferred than the ones met by $ic$''.
\end{example}

\section{
	Explainable Decision Making with Decision Graphs (DGs)}
\label{sec:dg}

ADFs and PDFs are very high-level formulations of decision problems, assuming the existence of ``tabular'' information linking decisions and goals directly. 
Richer representations, such as  graphs
, have been long used in decision making
, e.g. in the form of Bayesian (or belief) networks \cite{BenGal07} and influence
diagrams \cite{Howard2005}. 
In this section we will explore intantiations of
ADFs and PDFs with {\em Decision Graphs (DGs)}, giving 
a logical structure to the decision domain and
revealing the decision maker's uncertainties and constraints
\cite{Matt08-COMMA}.
We first define (various forms of) decision making with DGs (Section~\ref{sec:DMwithDGs}), followed by ABA mappings to capture this (varied) decision making (Section~\ref{sec:ABAwithDGs}), to conclude with explanations (Section~\ref{sec:XwithDGs}).

\subsection{Instantiating ADFs, PDFs and \Flat\ Explanations with 
DGs}
\label{sec:DMwithDGs}

DGs are generalised {\em and-or trees} \cite{Luger08}. In a DG, nodes
are of three types: \emph{decisions}, \emph{goals}, and \emph{intermediates}, representing, \respectively,
candidate decisions, goals, and attributes in the modelled decision problem. Edges represent relations
amongst nodes, e.g. an edge from a decision to an intermediate
represents that the decision ``has'' the intermediate attribute; an
edge from an intermediate to a goal represents that the intermediate
``satisfies'' the goal; an edge from an intermediate to another
intermediate represents that the former ``leads to'' the
latter. Edges are of two types: \emph{standard} and \emph{defeasible}. A standard edge from a node to another represents that the former \emph{definitely} leads to the latter, whereas a defeasible edge represents a \emph{tentative} association, to be dropped if some \emph{defeasible condition} (expressed in a suitable formal language) for the edge are satisfied. Finally, edges (of whichever type) are \emph{tagged} with a natural number: 
edges (from node a to node b) provide (i) stand-alone support (from a to b) if they are tagged with 1, and (ii) conjoined support with other edges with the same tag, if greater than 1.  Formally: 

\begin{definition}
\label{dfn:decGraph}
A {\em Decision Graph (DG)} is a tuple \edgTuple, in which 
	\begin{itemize}
		\item \dgPair{} is a directed acyclic graph with  
			\begin{itemize}
				\item 
$\node = \dNode \cup \iNode \cup \gNode$ a set of {\em nodes}
  \st{} $\dNode$, $\iNode$ and $\gNode$ are pairwise disjoint 
  and
$\dNode \neq \emptyset$ is a set of {\em decisions};
$\iNode$ is a set of {\em intermediates}; 
$\gNode \neq \emptyset$ is a set of {\em goals};

\item
	$\edge=\sEdge  \cup \dEdge$ a set of {\em edges} \st{}  $\sEdge$ and $\dEdge$ are pairwise disjoint and $\sEdge$ is a set of \emph{standard} edges, $\dEdge$ is a set of \emph{defeasible} edges; 
					edges are of the form
					$\egPair{n_i}{n_j}$ 
					for $n_i, n_j
					\in \node$ \st{}
$\egPair{n_i}{n_j}$ is in $\edge$ \ifaf{}
either $n_i \in \dNode$ and $n_j \in \iNode \cup \gNode$,
or $n_i \in \iNode$ and $n_j \in \iNode \cup \gNode$;
each $e \in \edge$ is {\em tagged} with a number $i$ \st{} $i
					\in \mathbb{N}$, denoted by $\myTag{e} = i$ (if the tag of an edge is
					$1$, then it is often omitted);

\end{itemize}
		\item  \info{} (referred to as {\em \bSet})
is a (finite) set of implications of the form:
$$t_n \wedge \ldots \wedge t_1 \rightarrow t_0$$ 
			for $n \geq 0$ and $t_i \in \LB$ (for $i=0, \ldots, n$) with $\LB$ a formal language such that for each defeasible edge
$e = \egPair{n}{n'} \in \dEdge$ with $\myTag{e} = i$ there is
			a sentence $\neg dEdge(n,n',i) \in \LB$
.
	\end{itemize}	
\end{definition}

We modify Example~\ref{exp:adf} to illustrate DGs. 
\begin{example}
\label{exp:graph}
\label{exp:infoExp}
Figure~\ref{fig:edg} shows an example of a DG. 
The agent wants this accommodation to be convenient and cheap. The two
candidates are Imperial College London Student Accommodation ($ic$)
and Ritz Hotel ($ritz$). $ic$ is \pounds 50 a night and in South
Kensington ($inSK$). Ritz is \pounds 200 a night and in Piccadilly
($inPic$). Ritz is also having a promotion discount. \pounds 50 per
night is $cheap$ so $ic$ is cheap. \pounds 200 a night with the
current promotion discount makes $ritz$ $cheap$ as well. Hence both
places are $cheap$. However, South Kensington is $near$ so it is
$convenient$ whereas Piccadilly is not. Intuitively, $ic$ is the
better choice.
Here:
\begin{itemize}
\item
the decisions are: $\dNode = \{ic, ritz\}$;
\item
the goals are: $\gNode = \{convenient, cheap\}$;
\item
the intermediates are: $\iNode = \{inSK, 50, inPic, 200, discount, near\}$.
\end{itemize}
The 
edges from $ic$ to 50 and $ic$ to $inSK$ are defeasible
(shown with dashed arrows in the figure), meaning that it is defeasibly known that
$ic$ is $\pounds 50$ a night and in South Kensington. The remaining
edges are strict (shown with solid arrows in the figure), meaning that they are standard.
	Also, we let the \bSet{} \info{} be:
$$\info{} = \{\rightarrow termTime; termTime \rightarrow \neg
dEdge(ic,50,1)\}$$  
with language $\LB = \{termTime, \neg dEdge(ic,50,1), \neg
dEdge(ic,inSK,1)\}$. The two implications in \info{} state that: it is
currently term time and if it is term time, then $ic$ is not $\pounds 50$ per night.
	Finally, all edges are tagged with 1 (mostly omitted, as per convention), except for two edges (tagged with 2): they indicate that $cheap$ is supported by the conjunction of $200$ and $discount$; instead, for example, $50$ alone supports $cheap$. 
\end{example}

\begin{figure}
\begin{minipage}{1\textwidth}
\begin{minipage}{0.5\textwidth}
\begin{scriptsize}
\centerline{
\xymatrix@1@=12pt@R=12pt{
                 & ic \ar@{-->}@/_/[ld] \ar@{-->}[d] && ritz \ar@/_/[ld] \ar[d] \ar@/^/[rd] \\
inSK \ar[d]      & 50 \ar@/_1pc/[rrdd]|-{1} & inPic  & 200 \ar[dd]|-{2} & discount \ar@/^/[ldd]|-{2}\\
near \ar@/_/[rd] &                          &                    \\
                 & convenient              && cheap             
}
}
\end{scriptsize}
\end{minipage}
\hspace{10pt}
\begin{minipage}{0.5\textwidth}
$\info{} = \{\rightarrow termTime;$ 

\hspace{28pt}$termTime \rightarrow \neg dEdge(ic,50,1)\}$
\end{minipage}
\end{minipage}
	\caption{DG for Example~\ref{exp:infoExp}. Solid arrows
  ($\longrightarrow$) represent strict edges whereas dashed arrows
  ($\DashedArrow[->,densely dashed]$) represent defeasible edges.
\label{fig:edg}}  
\end{figure}
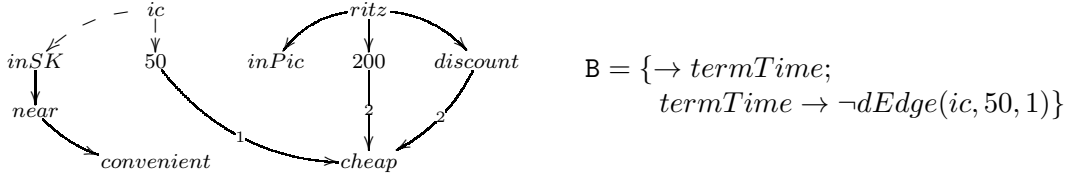

To instantiate the notion of {\em decisions-meeting-goals} with DGs, we
define a notion of {\em reachability} from a set of nodes to a
node, taking into account that defeasible edges may be \emph{blocked}, depending on whether this can be inferred from the \bSet. We will assume that 
$\LB$ is equipped with an inference mechanism $\vdash_{MP}$ amounting to the repeated application of modus ponens with implications: formally, for $s \in \LB$, $\info \vdash_{MP}s$ \ifaf{} there is a sequence $s_1, \ldots, s_m=s$ of sentences in $\LB$ and $m$ implications in $\info$ such that, for each $i \in \{1, \ldots, m\}$, either $\rightarrow s_i \in \info$ or $t_1, \ldots, t_n \rightarrow s_i \in \info$ and $\{t_1, \ldots, t_n\} \subseteq \{s_1, \ldots s_{i-1}\}$. 

\begin{definition}
\label{dfn:reach}
\label{dfn:unEdge}
Given a DG $G=\edgTuple$ 
 with $\edge = \sEdge \cup \dEdge$, let the
{\em blocked edges} of $G$ be $\unreach(G) \subseteq \dEdge$ \st: 
	$\unreach(G) = \{e \in \dEdge | e = \egPair{n_i}{n_j}, \myTag{e} = k,
	\info \vdash_{MP} \neg dEdge(n_i,n_j,k)\}$. 	
Then, for $n \in \node, N \subseteq \node$, 
	we say that $n$ is {\em reachable} from $N$ \ifaf{} one of the
following two conditions holds:

\begin{enumerate}[label={C\arabic*.}]
\item
	there exists a tag $k$ \st{} $N = \{n_i \in \node | e_i = \egPair{n_i}{n} \in
\edge \setminus \unreach(G)$ and $\myTag{e_i} = k\}$; or 

\item
there exists some $N' \subseteq \node$ \st{} $n$ is reachable from
$N'$ and for each $n' \in N'$, $n'$ is reachable from $N$.
\end{enumerate}
\end{definition}

Definition~\ref{dfn:reach} is given recursively with C1 being the base
case. C1 specifies that if there is a set of nodes $N$ all
having edges with the same tag pointing to some node $n$ and none being blocked, then $n$ is
reachable from $N$. C2 specifies that if there is some set of nodes
$N'$ satisfying $C1$ ($n$ being reachable from $N'$) and every single
node $n'$ in $N'$ is reachable from $N$, then $n$ is reachable from $N$.

\begin{example}
\label{exp:blockEdge}
(Example~\ref{exp:infoExp} continued.) There are two defeasible edges
in this example: $\egPair{ic}{inSK}$ and $\egPair{ic}{50}$. It is easy
to see that $\egPair{ic}{50}$ is blocked whereas $\egPair{ic}{inSK}$
is not. Thus, 
\label{exp:edg}
\begin{itemize}
\item
$inSK$ 
		is reachable from $\{ic\}$, whereas $50$ is not, since the edge
$\egPair{ic}{50}$ is blocked, 
		 
\item
$inPic$, $200$, $discount$ are reachable from $\{ritz\}$,
\item
$near$ is reachable from $\{inSK\}$ and $\{ic\}$,
\item
$convenient$ is reachable from $\{near\}$, $\{inSK\}$ and $\{ic\}$,
\item
$cheap$ is reachable from $\{50\}$, $\{200, discount\}$, 
		and $\{ritz\}$, but not $\{ic\}$ since the edge
$\egPair{ic}{50}$ is blocked.
\end{itemize}
\end{example}

With reachability defined, decisions-meeting-goals in DGs is defined as
follows.
\begin{definition}
\label{dfn:meet}
Given a DG $\edgTuple$, $\node = \dNode \cup \iNode \cup \gNode$, with
$\dNode$ the decisions and $\gNode$ the goals, a decision $d \in
\dNode$ {\em meets} a goal $g \in \gNode$, \ifaf{} $g$ is reachable
from $\{d\}$. We again use $\dmg(d) \subseteq \gNode$ to
denote the set of goals met by $d$.
\end{definition}

\begin{example}
\label{exp:meet}

(Example~\ref{exp:
	blockEdge} continued.) The decision $ic$ meets the goal
	$convenient$ but not the goal $cheap$, whereas the decision $ritz$ 
	meets the goal
$cheap$.
\end{example}

\begin{proposition}
\label{prop:DGinstanceADF}
For any DG $G$=\edgTuple, there is an 
ADF
$\df = \nadf$
\st{} a
decision $d \in \decisions$ meets a goal $g \in \goals$ in $G$
\ifaf{} $d$ meets $g$ in $\df$.
\end{proposition}

The proof of this and all other results in this section are also
  in Appendix~\ref{sec:proof}.

Note that Proposition~\ref{prop:DGinstanceADF} sanctions that DGs
are instances of ADFs. It is easy to see that ADFs are also, trivially, instances of DGs, namely: 

\begin{proposition}
\label{prop:ADFinstanceDG}
For any ADF
$\df = \nadf$,
there is a DG $G$ \st{} a
decision $d \in \decisions$ meets a goal $g \in \goals$ in $\df$
\ifaf{} $d$ meets $g$ in $G$.
\end{proposition}

By virtue of Proposition~\ref{prop:DGinstanceADF},
decision criteria, i.e. (strong / weak) dominance, directly apply in
DGs. 

\begin{example}
\label{exp:wdom}
(Example~\ref{exp:
	meet} continued.) 
	Given that $ic$ meets the
  goal $convenient$ and $ritz$ meets the goal $cheap$, both $ic$
  and $ritz$ are weakly dominant.
\end{example}

Also, 
preferences over goals can be added to DGs, using partial orders over sets of goals as in the case of ADFs:

\begin{definition}
  \label{dfn:pdg}

  A {\em \PDG{} (PDG)} is a tuple \pdgTuple, in which \edgTuple{} is
  a DG with $\node = \dNode \cup \iNode \cup \gNode$ and
  $\preferenceSet$ is a partial order over $2^{\gNode}$.
\end{definition}

	\begin{proposition}
		\label{prop:DGinstancePDF}
		For any PDG $G$=\pdgTuple, there is a 
PDF $\df=\neadfps$ \st{} a
decision $d \in \decisions$ meets a goal $g \in \goals$ in $G$
\ifaf{} $d$ meets $g$ in $\df$. 
	\end{proposition}

		Then, by virtue of Proposition~\ref{prop:DGinstancePDF}, Definition~\ref{decFuncPSet} can be directly applied
	to define \emph{\pfrSet{} decisions in PDGs} as illustrated next.

\iffalse 	NOT NEEDED? 
\begin{definition}
  \label{dfn:pfrSetPDG}

  A decision $d \in \decisions$ is {\em \pfrSet} (in $\pdgTuple$)
  \ifaf{} it is weakly dominant in \edgTuple{} and for all weakly
  dominant decisions $d' \in \dNode$ such that $d \neq d'$, 
  for all $s \in \cgset$ (the comparable goal set in $\pdgTuple$):
\begin{itemize}
\item if $s \not \subseteq \dmg(d)$ and $s \subseteq
\dmg(d')$, then there exists $s' \in \cgset$, such that

\hspace{20pt}
$s' \preferenceSet s$, 
\hspace{40pt}$s' \subseteq \dmg(d)$, and 
\hspace{40pt}$s' \not\subseteq \dmg(d')$.
\end{itemize}
\end{definition}
\fi

\begin{example}
  \label{exp:pdg}
	(Example~\ref{exp:wdom} continued.) 
  We add preference to the DG shown in
  Figure~\ref{fig:edg}: between the two goal nodes, {\em
    convenient} and {\em cheap}, we let $\{convenient\} \preferenceSet
  \{cheap\}$. 
	Since both $ic$
  and $ritz$ are weakly dominant but $\{convenient\} \preferenceSet
  \{cheap\}$, $ritz$ is \pfrSet.
\end{example}

Finally, note that all notions of \flat\ explanations are directly
applicable to DGs and PDGs, given that they give rise to ADFs and
PDFs, \respectively.
For illustration, in Example~\ref{exp:wdom},
$\expPair{\{convenient\}}{\{(cheap,ritz)\}}$
is a \flat\ explanation for $ic$ being weakly dominant, which can be
read as ``decision $ic$ is weakly dominant because it meets the goal
$convenient$, although the decision $ritz$ meets another goal $cheap$
not met by $ic$''; 
and, in Example~\ref{exp:pdg},
\expPair{\{cheap\}}{\expPair{\{cheap\}}{ic}}
is a \flat\ explanation for $ritz$ being \pfrSet, which can be read as
``$ritz$ is \pfrSet as it meets the goal $cheap$, and the alternative decision $ic$
does not meet anything more preferred than $cheap$''.

Note that other definitions of \flat\ explanations are possible
  when using DGs, e.g. of the kind illustrated in the Introduction
  with the illustrative toy example, whereby the reasons for goals
  being met are accommodated within the explanations. We leave the
  exploration of other forms of \flat\ explanation with DGs as future
  work. Instead, in the remainder of this section we will focus on
  \dialogical\ explanations, based on ABA mappings that also
  incorporate the notion of reachability, as defined next.

\subsection{(Strongly, Weakly) Dominant and \PfrSet{} ABA Frameworks with 
DGs}
\label{sec:ABAwithDGs}

	To build an ABA counterpart for 
explanation of decisions with 
DGs and PDGs, we need a change in the
decisions-meeting-goals relation to reflect the ``reasoning'' with DGs. We first define the {\em core ABA
  framework corresponding to} DGs without defeasible edges and thus with empty \bSet\ \info, as follows.

\begin{definition}
  \label{dfn:coreDGs}

	The {\em core ABA framework corresponding to} $\edgTupleEmpty$, $\node =
  \dNode \cup \iNode \cup \gNode$
	and $\edge=\sEdge
	$ with $\sEdge$ the strict edges, is $\lracFcoreG$, where

  \begin{itemize}
  \item
    $\RcoreG = \{edge(n_1,n_2,\myTag{e}) \gets | n_1, n_2 \in \node,
    e=\egPair{n_1}{n_2} \in \edge\} \ \cup$

    $\{reach(n_1,n_2) \gets edge(n_1,n_2,\myTag{e}) | n_1, n_2 \in
    \node, e=\egPair{n_1}{n_2} \in \edge\} \ \cup$  

    $\{reach(n_1,n_2) \gets reach(n_1,n_3), edge(n_3,n_2,\myTag{e}),
    \neg unreachableSib(n_3,n_2,\myTag{e},n_1) | n_1, $ $n_2, n_3 \in
    \node, n_3 \not\in \decisions, n_1 \neq n_2, n_1 \neq n_3,$ $e =
    \egPair{n_3}{n_2} \in \edge, \egPair{n_1}{n_2} \notin \edge\} \ \cup$

    $\{unreachableSib(n_3,n_2,\myTag{e},n_1) \gets
    edge(n_4,n_2,\myTag{e}), \neg reach(n_1,n_4) | n_1, n_2, n_3, n_4
    \in \node, n_1 \neq n_2, n_1 \neq n_3, n_4 \neq n_3, e =
    \egPair{n_3}{n_2} \in \edge, \egPair{n_4}{n_2} \in \edge \}
    \ \cup$

    $\{met(n_1, n_2) \gets reach(n_1, n_2) | n_1 \in \decisions, n_2
    \in \goals\}$;  

  \item
    $\AcoreG = \{\neg unreachableSib(n_2, n_3,\myTag{e},n_1) | \neg
    unreachableSib(n_2, n_3,\myTag{e},n_1)$ is in the body of a rule
    in $\RcoreG \} \cup$


    $\{\neg reach(n_1, n_2) | \neg reach(n_1, n_2)$ is in the body of
    a rule in $\RcoreG \} \cup$


    $\{notMet(d,g) | d \in \decisions, g \in \goals \}$;

  \item 
    for any $\neg unreachableSib(n_3,n_2,t,n_1)$ in $\AcoreG$,
    $\mc{C}_g(\neg unreachableSib(n_3,n_2,t,n_1)) =$ \\$
    \{unreachableSib(n_3,n_2,t,n_1)\}$;

    for any $\neg reach(n_1,n_2)$ in $\AcoreG$, $\mc{C}_g(\neg
    reach(n_1,n_2)) = \{reach(n_1,n_2)\}$;

    for any $notMet(d,g) \in \AcoreG$:
    $\mc{C}_g(notMet(d,g))=\{met(d,g)\}$.  
    
  \end{itemize}
\end{definition}

Although the construction of the core ABA framework 
underpinning this definition 
is somewhat tedious, its underlying intuition is
straightforward, 
describing decisions meeting goals in
DGs. Intuitively, Definition~\ref{dfn:coreDGs} can be understood as follows.
A decision $d$ meets a goal $g$ if the node $g$ is reachable from the
singleton set $\{d\}$. We know that a node $n_2$ is reachable from a
singleton set $\{n_1\}$ under one of the two conditions: 
\begin{enumerate}
\item
if there is an edge from $n$ to $n_2$ (regardless of the tag of the
edge), represented with the rule:
	$reach(n_1,n_2) \gets edge(n_1,n_2,\myTag{e})$, or

\item
if there is a node $n_3$ such that both of the following two conditions
hold: 
\begin{enumerate}
\item
$n_3$ is reachable from $\{n_1\}$, there is an edge from $n_3$ 
  to $n_2$ with some tag $t$, and

\item
it is not the case that there exists some node $n_4 \neq n_3$, such
that there is an edge from $n_4$ to $n_2$ with tag $t$, and $n_4$ is
not reachable from $\{n_1\}$, represented by rules: 

\hspace{-20pt}
	$reach(n_1,n_2) \gets reach(n_1,n_3), edge(n_3,n_2,t),
\neg unreachableSib(n_3,n_2,t,n_1)$; 
%
	and

\hspace{-20pt}
	$unreachableSib(n_4,n_2,t,n_1) \gets edge(n_4,n_2,t),
n_4 \neq n_3, \neg reach(n_1,n_3)$.

\end{enumerate}
\end{enumerate}


We give an illustration of core ABA framework for a DG without defeasible information in Example~\ref{exp:dgCore-app} in Appendix~\ref{app:ABA}.

If now we consider also defeasible information (i.e. defeasible edges and defeasible condition) we obtain the following augmented core ABA framework:

\begin{definition}
  \label{dfn:coreABADGD}

  Given a DG $\edgTuple$, $\edge = \dEdge \cup \sEdge$ with $\dEdge$
  the defeasible edges, $\sEdge$ the strict edges, let $\lracFcoreG$
	be the core ABA framework corresponding to $\edgTupleEmptyStrict$. Then, the {\em
    core ABA framework corresponding to} $\edgTuple$ is
  $\lracFcoreGD$, where

  \begin{itemize}
  \item
    $\RcoreGD = \RcoreG \ \cup$
    
    $\{edge(n_1,n_2,\myTag{e}) \gets dEdge(n_1,n_2,\myTag{e}) | e =
    \egPair{n_1}{n_2} \in \dEdge\} \ \cup$

    $\{t_0 \gets t_1, \ldots, t_n | t_n \wedge \ldots \wedge t_1
    \rightarrow t_0 \in \info\} $
   %
		  ;

  \item
    $\AcoreGD = \AcoreG \cup \{dEdge(n_1,n_2,\myTag{e}) |
    e = \egPair{n_1}{n_2} \in \dEdge\}$; 

  \item
    for all $dEdge(n,n',\myTag{e}) \in \AcoreGD$,
    $\CcoreGD(dEdge(n,n',\myTag{e}) = \{\neg
    dEdge(n,n',\myTag{e})\}$;

    for all $\neg unreachableSib(n_2, n_3,\myTag{e},n_1), \neg
    reach(n_1, n_2), notMet(d,g) \in \AcoreGD$,

    $\CcoreGD(\neg unreachableSib(n_2, n_3,\myTag{e},n_1)) =
    \CcoreG(\neg unreachableSib(n_2, n_3,\myTag{e},n_1))$; 

    $\CcoreGD(\neg reach(n_1, n_2)) = \CcoreG(\neg reach(n_1, n_2))$;

    $\CcoreGD(notMet(d,g)) = \CcoreG(notMet(d,g))$.
  \end{itemize}
\end{definition}

Definition~\ref{dfn:coreABADGD} is based on
Definition~\ref{dfn:coreDGs} with added structure to handle defeasible
edges and defeasible condition. Namely, all defeasible edges are represented as assumptions 
$dEdge(n_1,n_2,t)$ with contraries $\neg dEdge(n_1,n_2,t)$ for some
$n_1,n_2$ and $t$. Moreover, all implications in \bSet{} \info{} are
represented as rules. Hence, for a defeasible edge $e =
\egPair{n_1}{n_2}$ with $\myTag{e} = i$, if $e$ is blocked, then $\neg
dEdge(n_1,n_2,i)$ holds  in the core ABA framework corresponding to the
graph. 
It is immediate to see that  the 
    core ABA framework corresponding to $\edgTuple$, as given by Definition~\ref{dfn:coreABADGD}, is the same as the one given by Definition~\ref{dfn:coreDGs} when no  defeasible information is present in $\edgTuple$.

We give an illustration of core ABA framework for a DG without defeasible information in Example~\ref{exp:dgCoreD-app} in Appendix~\ref{app:ABA}.

As before, we can take the ``union'' of core ABA frameworks and
strongly dominant, dominant and weakly dominant components to obtain
ABA frameworks drawn from DGs. As such components, we can use exactly those we defined for ADFs.  Formally,

\begin{definition}
  \label{dfn:sdABADG}

  Given a DG $\dg = \edgTuple{}$, let \lracFcoreG{} be the core ABA
  framework corresponding to $\dg$ and let \lracFSD, \lracFD, \lracFWD{}
	be
  the strongly dominant, dominant, and weakly dominant components,
	as given in Definitions~\ref{dfn:sdABA}, \ref{dfn:dABA} and \ref{dfn:wdABA}
  \respectively. 
	Then the {\em Strongly Dominant, Dominant, Weakly
    Dominant ABA Frameworks drawn from} $\dg$ are \lracFiLiG{s},
  \lracFiLiG{d} and \lracFiLiG{w}, \respectively, in which: 

  \begin{itemize}
  \item
    $\mc{R}^g_s = \RcoreG \cup \mc{R}_s$; \
    $\mc{R}^g_d = \RcoreG \cup \mc{R}_d$; \
    $\mc{R}^g_w = \RcoreG \cup \mc{R}_w$;    

  \item
    $\mc{A}^g_s = \AcoreG \cup \mc{A}_s$; \
    $\mc{A}^g_d = \AcoreG \cup \mc{A}_d$; \
    $\mc{A}^g_w = \AcoreG \cup \mc{A}_w$;
    
  \item
for all \asm{} in $\mc{A}^g_s$:
		  \(\mc{C}^g_s(\asm) =
  \begin{cases}
    \CcoreG(\asm)  & \quad \text{if } \asm \in \AcoreG,\\
    \mc{C}_s(\asm) & \quad \text{if } \asm \in \mc{A}_s.\\
  \end{cases}
		  \)

for all \asm{} in $\mc{A}^g_d$:
		  \(\mc{C}^g_d(\asm) =
  \begin{cases}
    \CcoreG(\asm)  & \quad \text{if } \asm \in \AcoreG \text{ and } \asm \notin \mc{A}_d,\\
    \mc{C}_d(\asm) & \quad \text{if } \asm \in \mc{A}_d \text{ and }
    \asm \notin \AcoreG,\\
    \Ccore(\asm) \cup \mc{C}_d(\asm) & \quad \text{if } \asm \in \mc{A}_d \cap \AcoreG.\\    
  \end{cases}
		  \)

for all \asm{} in $\mc{A}^g_w$:
		  \(\mc{C}^g_w(\asm) =
  \begin{cases}
    \CcoreG(\asm)  & \quad \text{if } \asm \in \AcoreG,\\
    \mc{C}_w(\asm) & \quad \text{if } \asm \in \mc{A}_w.\\
  \end{cases}
		  \)

%
%
%
%
  \end{itemize}
\end{definition}

As before, strongly dominant, dominant and weakly dominant decisions
correspond to admissible arguments in their corresponding ABA
frameworks.

\begin{proposition}
\label{prop:SDABAGD}

Given a DG $\dg = \edgTuple{}$, $\node = \dNode \cup \iNode \cup
\gNode$ with $\dNode$ the decisions, $\gNode$ the goals, let $AF_s$,
$AF_d$ and $AF_w$ be the strongly dominant, dominant and weakly
dominant ABA frameworks, \respectively, drawn from $\dg$. Then for all
decisions $d \in \dNode$, $d$ is strongly dominant, dominant, weakly
dominant in $\dg$ \ifaf{} $\argu{\{sDom(d)\}}{sDom(d)}$,
$\argu{\{dom(d)\}}{dom(d)}$ and $\argu{\{wDom(d)\}}{wDom(d)}$ is
admissible in $AF_s$, $AF_d$ and $AF_w$, \respectively.
\end{proposition}

\label{sec:dgPref}

Also as before, ABA can be used to obtain \pfrSet{} decisions
in PDGs. We define the \emph{\pfrSet{} ABA framework corresponding to a
PDG} as the ``union'' of the core ABA framework and the \pfrSet{}
component, as follows: 

\begin{definition}
  \label{dfn:pdgABA}
  
  Given a PDG $\gp = \pdgTuple$, let $\lracFcoreGD$ be the core ABA
  framework corresponding to $\edgTuple$, $\lracFPG$ the \pfrSet{}
	component from Definition~\ref{dfn:pfrGoalSet}. Then, the {\em \pfrSet{} ABA framework corresponding to}
  $\gp$ is \lracF{} where:

  \begin{itemize}
  \item
    $\mc{R} = \RcoreGD \cup \mc{R}_p$;

  \item
    $\mc{A} = \AcoreGD \cup \mc{A}_p$;    
    
  \item
for all \asm{} in $\mc{A}$:
		  \(\mc{C}(\asm) =
  \begin{cases}
    \CcoreGD(\asm) & \quad \text{if } \asm \in \AcoreGD,\\
    \mc{C}_p(\asm) & \quad \text{if } \asm \in \mc{A}_p.\\
  \end{cases}
		  \)

%
%
%
%
%
%
  \end{itemize}
\end{definition}

\begin{proposition}
  \label{prop:pfrSetDGABA}

  Given $\gp = \pdgTuple$, with $\node = \dNode \cup \iNode \cup
  \gNode$, let $AF$ be the \pfrSet{} ABA framework corresponding to
  $\gp$. Then for all decisions $d \in \decisions$, $d$ is \pfrSet{}
  \ifaf{} $\argu{\{\pS(d)\}}{\pS(d)}$ is admissible in $AF$. 
\end{proposition}

In the next section we will show how the ABA reformulations for our various forms of decision making with DGs can serve as the basis for obtaining \dialogical\ explanations for the ``goodness'' (or ``badness) of decisions. Before we do so, though, we note that these ABA formulations can be seen as decision making abstractions in their own right, and thus provide forms of argumentation-based decision making. Indeed, they support the direct definition of reasons for goals being met, and one can easily think of further variants of DGs, e.g. where the \bSet can be defined in more general formal languages (of the kinds afforded by ABA) whereby, for example, one could allow arguments against the applicability of defeasible conditions. We leave these more general forms of argumentation-based decision making as future work.

\subsection{\Dialogical\ Explanations with 
DGs}
\label{sec:XwithDGs}


As in the abstract setting of ADFs and PDFs, \dialogical\ explanations for DGs can be defined in terms of (\leastAsm\ 
and best-effort
) dispute trees \wrt\ ABA frameworks drawn from DGs.  

\begin{example}
\label{exp:sdABAEDG}

	(Example~\ref{exp:edg} continued.) For the weakly dominant ABA framework drawn from the DG in
Figure~\ref{fig:edg}, 
	$\argu{\{sDom(ic)\}}{sDom(ic)}$
	is not admissible as shown by the best-effort dispute tree in Figure~\ref{fig:disputeTree2}. Indeed, as
the argument 
$B$,
stating $ic$ meets the goal $cheap$ because $ic$ is $\pounds 50$ per
night and $\pounds 50$ per night leads to $cheap$, is attacked by
$\argu{\{\}}{\neg dEdge(ic,50,1)}$, stating that $ic$ is not $\pounds
50$ per night, as it is known that during term time, $ic$ is not
$\pounds 50$ per night and it is currently term time. 
	Note that, in this illustration, the
	information presented in \info{} affects the decision, by providing an opponent argument defeating the proponent's argument $B$.

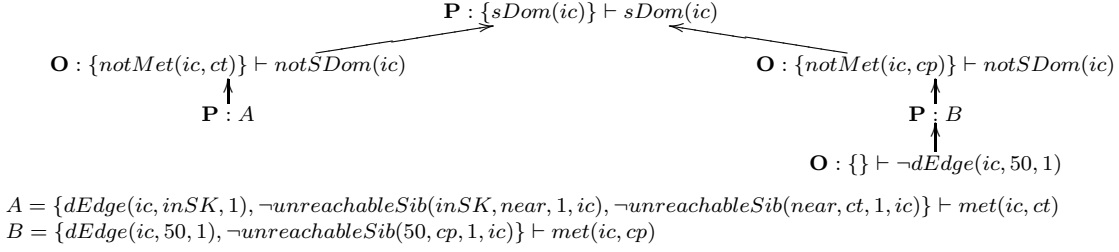
\begin{figure}
\begin{scriptsize}
\centerline{
\xymatrix@1@R=10pt@C=12pt{
& \pro:\argu{\{sDom(ic)\}}{sDom(ic)} \\
\opp:\argu{\{notMet(ic,ct)\}}{notSDom(ic)} \ar[ur] & & \opp:\argu{\{notMet(ic,cp)\}}{notSDom(ic)} \ar[lu] \\  
\pro:A \ar[u] & & \pro:B \ar[u] \\
         & & \opp:\argu{\{\}}{\neg dEdge(ic,50,1)} \ar[u]
}}

\medskip
$A = \argu{\{dEdge(ic,inSK,1), \neg unreachableSib(inSK,near,1,ic), \neg unreachableSib(near,ct,1,ic)\}}{met(ic,ct)}$ 

$B = \argu{\{dEdge(ic,50,1), \neg unreachableSib(50,cp,1,ic)\}}{met(ic,cp)}$ 

\end{scriptsize}
	\caption{
		Best-effort 
		 dispute tree for $\argu{\{sDom(ic)\}}{sDom(ic)}$ in Example~\ref{exp:sdABAEDG}.
  Here, $cp$ and $ct$ are short-hands for $cheap$ and $convenient$,
  \respectively.
\label{fig:disputeTree2}
} 
\end{figure}
\end{example}

As in Section~\ref{sec:explanation} \dialogical\
explanations for decisions (not) meeting the various decision
criteria integrate information about cooresponding \flat\ explanations, as follows. 

\begin{proposition}
  \label{prop:expDGS}

  Let $AF$ be the strongly dominant ABA framework corresponding to
  a DG $\dg$.

  For any strongly dominant decision $d$ in $\dg$, let $\atree$ be a
	\leastAsm{} 
	dispute tree for $\argu{\{sDom(d)\}}{sDom(d)}$.
  Then, 
  \begin{center}
    $\{g|N=\pro:\argu{\_}{met(d,g)}$ is a leaf node in $\atree$ or
	  is not the ancestor of any \opp{} leaf nodes$\}$
  \end{center}
	a \flat{} explanation for $d$ being strongly dominant.

	For any non-strongly dominant decision $d$, let $\atree$ be a best-effort 
  dispute tree for $\argu{\{sDom(d)\}}{sDom(d)}$. Then,
  \begin{center}
  $\{g|N=\opp:\argu{\{notMet(d,g)\}}{notSDom(d)}$ is a leaf node in
	  $\atree$ or 
	  is not the ancestor of any \opp{} leaf nodes$\}$
  \end{center}
	is a \flat{} explanation for $d'$ not being strongly dominant.
\end{proposition}

\begin{figure}
    \centerline{
      \xymatrix@1@=20pt@R=20pt{
        d_1 \ar[dd]|-{1} &                        & d_2 \ar[d]\ar[dl]\\
                         & a_1 \ar[dl]|-{2}\ar[d] & a_2 \ar[dl] \\
        g_1              & g_2                 
      }
    }
  \caption{DG for Example~\ref{exp:expDGS}.
	\label{fig:expDGS}}
\end{figure}
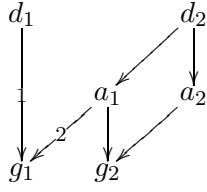

\begin{example}
\label{exp:expDGS}

We illustrate Proposition~\ref{prop:expDGS} with the DG 
	shown in Figure~\ref{fig:expDGS} (note that all edges are strict in this DG, and the accompanying \info\ is empty). Here, we have two decisions $d_1$
and $d_2$ along with two goals $g_1$ and $g_2$. $d_1$ meets $g_1$,
$d_2$ meets both $g_1$ (via $a_1$) and $g_2$ (via both $a_1$ and
$a_2$). Thus, $d_1$ is not strongly dominant but $d_2$ is.

	The ABA framework corresponding to this setting is given in Appendix~\ref{app:ABA}. 
	With this ABA framwork, 
dispute trees for both $\argu{\{sDom(d_1)\}}{sDom(d_1)}$ and
$\argu{\{sDom(d_2)\}}{sDom(d_2)}$ are shown in
Figures~\ref{fig:expDGSTree1} and \ref{fig:expDGSTree2} \respectively. From these
trees, we can easily see that a \flat{} explanation for $d_1$ not being
strongly dominant is $\{g_2\}$ (
as $g_2$ is not met by $d_1$) and a \flat{}
explanation for $d_2$ being strongly dominant is $\{g_1, g_2\}$ (
as both $g_1$ and $g_2$ are met by $d_1$).
\end{example}

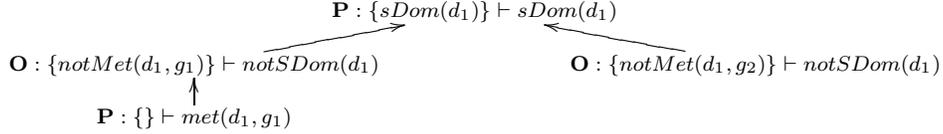
\begin{figure}
\begin{scriptsize}
\centerline{
\xymatrix@1@R=10pt@C=-20pt{
& \pro:\argu{\{sDom(d_1)\}}{sDom(d_1)} \\
\opp:\argu{\{notMet(d_1,g_1)\}}{notSDom(d_1)} \ar[ur] & & \opp:\argu{\{notMet(d_1,g_2)\}}{notSDom(d_1)} \ar[lu] \\  
\pro:\argu{\{\}}{met(d_1,g_1)} \ar[u] }}
\end{scriptsize}
\caption{
		Best-effort dispute tree for $\argu{\{sDom(d_1)\}}{sDom(d_1)}$ for Example~\ref{exp:expDGS}.
\label{fig:expDGSTree1}
} 
\end{figure}

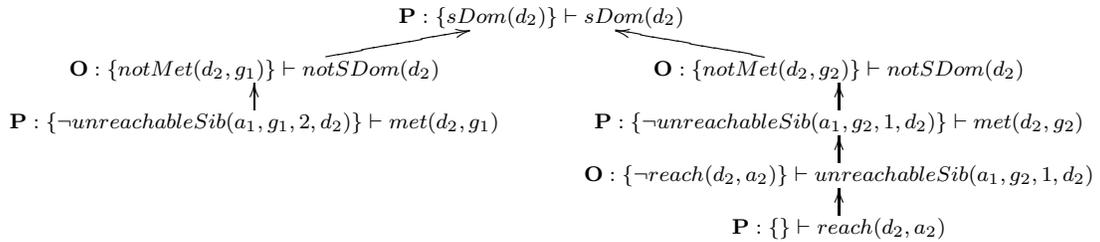
\begin{figure}
\begin{scriptsize}
\centerline{
\xymatrix@1@R=10pt@C=-40pt{
& \pro:\argu{\{sDom(d_2)\}}{sDom(d_2)} \\
\opp:\argu{\{notMet(d_2,g_1)\}}{notSDom(d_2)} \ar[ur] & & \opp:\argu{\{notMet(d_2,g_2)\}}{notSDom(d_2)} \ar[lu] \\  
\pro:\argu{\{\neg unreachableSib(a_1,g_1,2,d_2)\}}{met(d_2,g_1)} \ar[u] & & \pro:\argu{\{\neg unreachableSib(a_1,g_2,1,d_2)\}}{met(d_2,g_2)} \ar[u] \\
                                                                       & & \opp:\argu{\{\neg reach(d_2,a_2)\}}{unreachableSib(a_1,g_2,1,d_2)}\ar[u] \\
                                                                       & & \pro:\argu{\{\}}{reach(d_2,a_2)}\ar[u] }}
\end{scriptsize}
	\caption{\LeastAsm{} 
	dispute tree for $\argu{\{sDom(d_2)\}}{sDom(d_2)}$ for Example~\ref{exp:expDGS}.
\label{fig:expDGSTree2}
} 
\end{figure}

\begin{proposition}
  \label{prop:expDGD}

  Let $AF$ be the dominant ABA framework corresponding to $\dg$.

  For any dominant decision $d$ in $\dg$, let $\atree$ be a \leastAsm{} 
	dispute tree for $\argu{\{dom(d)\}}{dom(d)}$. Then the
  pair $\expPair{G}{F}$, where 
  \begin{itemize}
    \item
	    $G = \{g|\pro:\argu{\{\}}{met(d,g)}$ is not 
		  ancestor of  any
      \opp{} leaf nodes in $\atree\}$, 
    \item
      $F = \{g|\pro:\argu{\{noOthers(d,g)\}}{noOthers(d,g)}$ is not
		  ancestor of 
		  any \opp{} leaf nodes in $\atree\}$,  
  \end{itemize}
	is a \flat{} explanation for $d$ being dominant.

	For any non-dominant decision $d$, let $\atree$ be the best-effort 
	dispute
  tree for $\argu{\{dom(d)\}}{dom(d)}$. Then
  \begin{itemize}
  \item
	  $\{g|\opp:\argu{\{notMet(d,g)\}}{notDom(d)}$ is 
		  ancestor of an
    \opp{} leaf node$\}$
  \end{itemize}
	is a \flat{} explanation for $d$ not being dominant.
\end{proposition}

\begin{figure}
    \centerline{
      \xymatrix@1@=20pt@R=20pt{
        d_1 \ar[dd]|-{1} &                        & d_2 \ar[d]\ar[dl]\\
                         & a_1 \ar[dl]|-{2}\ar[d] & a_2 \ar[dl] \\
        g_1              & g_2                    & & g_3 
      }
    }
  \caption{DG for Example~\ref{exp:expDGD}.
	(This is the same as the one in Figure~\ref{fig:expDGS} except for the presence of goal $g_3$.)  \label{fig:expDGD}
	}
\end{figure}
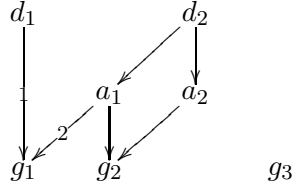

\begin{example}
\label{exp:expDGD}

We illustrate Proposition~\ref{prop:expDGD} with the DG
shown in Figure~\ref{fig:expDGD}. This example is adapted from
Example~\ref{exp:expDGS} with a newly added goal $g_3$ that is not met
by either $d_1$ or $d_2$. Thus, neither of the two is strongly
dominant any more. However, since $d_2$ still meets all goals met by
$d_1$ and more, $d_2$ is dominant and $d_1$ is not.
	The ABA framework corresponding to this setting is given in Appendix~\ref{app:ABA}. For this ABA framework,
Figures~\ref{fig:expDGDTree1} and \ref{fig:expDGDTree2} show \dialogical\ explanations in the form of 
dispute trees for $\argu{\{dom(d_1)\}}{dom(d_1)}$ and
$\argu{\{dom(d_2)\}}{dom(d_2)}$, \respectively. $\{g_2\}$ is a \flat{}
explanation for $d_1$ not being dominant as
$\opp:\argu{nM(d_1,g_2)}{nD(d_1)}$ is 
ancestor of the leaf node
$\opp:\argu{\{\}}{reach(d_2,a_2)}$. We read this as

\begin{quote}
  $d_1$ is not dominant because it does not meet the goal $g_2$, which
  is met by some other decision.
\end{quote}

\expPair{\{g_1, g_2\}}{\{g_3\}} is a \flat{} explanation for $d_2$
being dominant as $\pro:\argu{\{\neg
  uS(a_1,g_1,2,d_2\}}{met(d_2,g_1)}$ is a leaf node and 
$\pro:\argu{\{\neg uS(a_1,g_2,1,d_2)\}}{met(d_2,g_2)}$ is not
ancestor of any \opp{} leaf node. Moreover,
$\pro:\argu{\{nO(d_2,g_3)\}}{nO(d_2,g_3)}$ is also a leaf node in the
tree.
\begin{quote}
$d_2$ is dominant as it meets both goals $g_1$ and $g_2$ and for the
  goal $g_3$ it does not meet, $g_3$ is not met by the other decision
  $d_1$.
\end{quote}
\end{example}

\begin{figure}
\begin{scriptsize}
\centerline{
\xymatrix@1@R=10pt@C=0pt{
                                              & \pro:\argu{\{dom(d_1)\}}{dom(d_1)} \\
  \opp:\argu{\{nM(d_1,g_1)\}}{nD(d_1)} \ar[ur] & \opp:\argu{\{nM(d_1,g_2)\}}{nD(d_1)} \ar[u]     & \opp:\argu{\{nM(d_1,g_3)\}}{nD(d_1)} \ar[lu] \\  
  \pro:\argu{\{\}}{met(d_1,g_1)} \ar[u]        & \pro:\argu{\{nO(d_1,g_2)\}}{nO(d_1,g_2)} \ar[u] & \pro:\argu{\{nO(d_1,g_3)\}}{nO(d_1,g_3)} \ar[u] \\
                                              & \opp:\argu{\{\neg uS(a_1,g_2,1,d_2)\}}{met(d_2,g_2)} \ar[u] \\
                                              & \pro:\argu{\{\neg reach(d_2,a_2)\}}{uS(a_1,g_2,1,d_2)}\ar[u] \\
                                              & \opp:\argu{\{\}}{reach(d_2,a_2)}\ar[u] }}
\end{scriptsize}
\caption{
	Best-effort 
	 dispute tree for
	$\argu{\{dom(d_1)\}}{dom(d_1)}$ for Example~\ref{exp:expDGD}. Here, $nM$, $nD$, $nO$ and $uS$ are
  short-hands for $notMet$, $notDom$, $noOthers$ and $unreachableSib$,
  \respectively.
\label{fig:expDGDTree1}
} 
\end{figure}
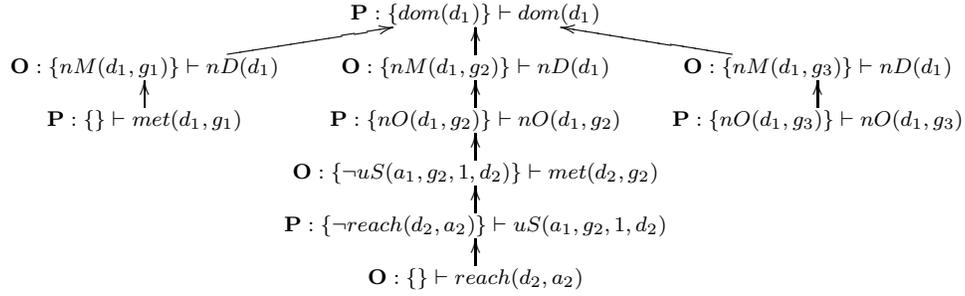

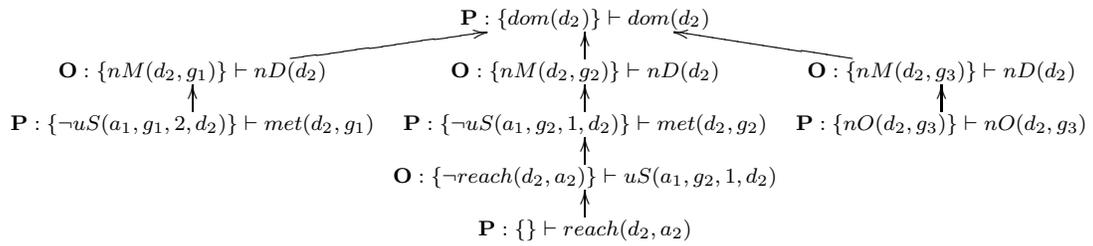
\begin{figure}
\begin{scriptsize}
\centerline{
\xymatrix@1@R=10pt@C=5pt{
                                                  & \pro:\argu{\{dom(d_2)\}}{dom(d_2)} \\
  \opp:\argu{\{nM(d_2,g_1)\}}{nD(d_2)} \ar[ur] & \opp:\argu{\{nM(d_2,g_2)\}}{nD(d_2)} \ar[u] & \opp:\argu{\{nM(d_2,g_3)\}}{nD(d_2)} \ar[lu] \\  
  \pro:\argu{\{\neg uS(a_1,g_1,2,d_2)\}}{met(d_2,g_1)} \ar[u] & \pro:\argu{\{\neg uS(a_1,g_2,1,d_2)\}}{met(d_2,g_2)} \ar[u] & \pro:\argu{\{nO(d_2,g_3)\}}{nO(d_2,g_3)} \ar[u] \\
                                                  & \opp:\argu{\{\neg reach(d_2,a_2)\}}{uS(a_1,g_2,1,d_2)}\ar[u] \\
                                                  & \pro:\argu{\{\}}{reach(d_2,a_2)}\ar[u] }}
\end{scriptsize}
	\caption{\LeastAsm{} 
	dispute tree for $\argu{\{dom(d_2)\}}{dom(d_2)}$ for Example~\ref{exp:expDGD}.
  Here, $nM$, $nD$, $nO$ and $uS$ are short-hands for
  $notMet$, $notDom$, $noOthers$ and $unreachableSib$, \respectively.
\label{fig:expDGDTree2}
} 
\end{figure}

\begin{proposition}
  \label{prop:expDGW}

  Let $AF$ be the weakly dominant ABA framework corresponding to
  $\dg$.

  For any $d$ in $\dg$, if $d$ is weakly dominant, let $\atree$ be a
	\leastAsm{} 
	dispute tree for the argument
  $\argu{\{wDom(d)\}}{wDom(d)}$. Then the pair $\expPair{G}{M}$, where  
  \begin{itemize}
  \item
  $G=\{g|\pro:\argu{\_}{\{met(d,g)\}}$ or
		  $\pro:\argu{\{notMet(\_,g)\}}{more(d,\_)}$ is not 
		  ancestor of
		  an
    \opp{} leaf node in $\atree\}$,

  \item
  $M=\{(d',g)|\pro:\argu{\{notMet(d',g)\}}{more(d,d')}$ is not
		  ancestor of an \opp{} leaf node in $\atree\}$
  \end{itemize}
	is  a \flat{} explanation for $d$ being weakly dominant.  
  
	For any non-weakly dominant decision $d$, let $\atree$ be a 
	best-effort dispute
  tree for the argument $\argu{\{wDom(d)\}}{wDom(d)}$. Then
  \begin{itemize}
  \item
    $\{d'|\opp:\argu{\{\_,notMore(d',d)\}}{notWDom(d)}$ is
		  ancestor of
		  an \opp{} leaf node in $\atree\}$.
  \end{itemize}
	is a \flat{} explanation for $d$ not being weakly dominant.
\end{proposition}

\begin{figure}[ht]
    \centerline{
      \xymatrix@1@=20pt@R=20pt{
        d_1 \ar[dd]|-{1} &                        & d_2 \ar[d]\ar[dl] & d_3 \ar[d] \\
                         & a_1 \ar[dl]|-{2}\ar[d] & a_2 \ar[dl]       & a_3 \ar[d] \\
        g_1              & g_2                    &                   & g_3 
      }
    }
  \caption{DG for Example~\ref{exp:expDGW}. (This is the same as
      the one in Figure~\ref{fig:expDGD} except for the presence of
      decision $d_3$ reaching $g_3$ via $a_3$.) 
  \label{fig:expDGW}}
\end{figure}
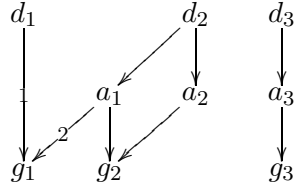

\begin{example}
  \label{exp:expDGW}

  We illustrate Proposition~\ref{prop:expDGW} with the DG
  shown in Figure~\ref{fig:expDGW}. This example is adapted from
  Example~\ref{exp:expDGS} and \ref{exp:expDGD} with a newly added
  decision $d_3$ meeting the goal $g_3$ (via $a_3$). Thus, both $d_2$
  and $d_3$ are weakly dominant whereas $d_1$ is not. 
	The ABA
	framework corresponding to this setting is given in Appendix~\ref{app:ABA}. For this ABA framework, 
  Figures~\ref{fig:expDGWTree1}, \ref{fig:expDGWTree2} and
  \ref{fig:expDGWTree3} show three \dialogical\ explanations in the form of 
  dispute trees for
  $\argu{\{wDom(d_1)\}}{wDom(d_1)}$, $\argu{\{wDom(d_2)\}}{wDom(d_2)}$
  and $\argu{\{wDom(d_3)\}}{wDom(d_3)}$, \respectively. From these
  trees, we can extract \flat{} explanations for decisions as follows.
  
  From Figure~\ref{fig:expDGWTree1}, $\{d_2\}$ is a \flat{} explanation
  for $d_1$ not being weakly dominant (because $\opp:\argu{\{\neg
    uS(a_1,g_2,1,d_2),nM(d_1,g_2),nE(d_1,d_2)\}}{nW(d_1)}$ is
  ancestor of
  the leaf node $\opp:\argu{\{\}}{reach(d_2,a_2)}$). We can 
  read this as:  
  \begin{quote}
    Since $d_2$ meets more goals than $d_1$, $d_1$ is not weakly dominant.
  \end{quote}

  From Figure~\ref{fig:expDGWTree2}, since both argument $\argu{\{\neg
    uS(a_1,g_1,2,d_2)\}}{met(d_2,g_1)}$ and argument $\argu{\{\neg
    uS(a_1,g_1,2,d_2),notMet(d_3,g_1)\}}{more(d_1,d_3)}$ are not
  ancestors of any $\opp$ leaf node, $G=\{g_1\}$ and
  $M=\{\expPair{d_3}{g}\}$. Thus,
	$\expPair{\{g_1\}}{\{\expPair{d_3}{g}\}}$ is a \flat{} explanation for
  $d_2$ being weakly dominant. We can read this as
  \begin{quote}
    $d_2$ is weakly dominant because it meets the goal $g_1$, which is
    not met by $g_3$.
  \end{quote}

  From Figure~\ref{fig:expDGWTree3}, since both argument
  $\argu{\{\neg uS(a_3,g_3,1,d_3),notMet(d_2,g_3)\}}{more(d_3,d_2)}$
  and argument $\argu{\{\neg
    uS(a_3,g_3,1,d_3),notMet(d_1,g_3)\}}{more(d_3,d_1)}$ are not
  ancestors of any \opp{} leaf node,
  $\exp{\{g_3\}}{\{\expPair{d_1}{g_3}, \expPair{d_1}{g_3}\}}$ is a \flat{}
  explanation for $d_3$ being weakly dominant. We can read this as
  \begin{quote}
    The decision $d_3$ is weakly dominant because it meets the goal
    $g_3$, which is not met by $d_1$ or $d_2$.
  \end{quote}  
\end{example}

\begin{figure}[bht]
\begin{scriptsize}
\centerline{
\xymatrix@1@R=10pt@C=8pt{
                                        & \pro:\argu{\{wDom(d_1)\}}{wDom(d_1)} \\
  \opp:A \ar[ur] & \opp:B \ar[u] & \opp:C \ar[ul] \\  
  \pro:\argu{\{\}}{met(d_1,g_1)}\ar[u] & \pro:\argu{\{\neg reach(d_2,a_2)\}}{\neg uS(a_1,g_2,1,d_2)}\ar[u] & \pro:\argu{\{notMet(d_3,g_1)\}}{more(d_1,d_3)}\ar[u] \\
                                       & \opp:\argu{\{\}}{reach(d_2,a_2)} \ar[u] }}

\medskip
\hspace{-10pt}\begin{tabular}{ll}
$A=\argu{\{\neg uS(a_1,g_1,2,d_2),nM(d_1,g_1),nE(d_1,d_2)}{nW(d_1)}$ & $C=\argu{\{\neg uS(a_3,g_3,1,d_3),nM(d_1,g_3),nE(d_1,d_3)}{nW(d_1)}$\\
$B=\argu{\{\neg uS(a_1,g_2,1,d_2),nM(d_1,g_2),nE(d_1,d_2)}{nW(d_1)}$\\
\end{tabular}

\end{scriptsize}
\caption{
	Best-effort
	 dispute tree for
	$\argu{\{wDom(d_1)\}}{wDom(d_1)}$ for Example~\ref{exp:expDGW}. Here, $nM$, $nE$,  $nW$ and $uS$
  are short-hands for $notMet$, $notMore$, $notWDom$ and
  $unreachableSib$, \respectively.
\label{fig:expDGWTree1}
}
\end{figure}

\begin{figure}[ht]
\begin{scriptsize}
\centerline{
\xymatrix@1@R=10pt@C=-50pt{
                 & \pro:\argu{\{wDom(d_2)\}}{wDom(d_2)} \\
  \opp:\argu{\{nM(d_2,g_1),nE(d_2,d_1)\}}{nW(d_2)} \ar[ur] & & \opp:\argu{\{\neg uS(a_3,g_3,1,d_3),nM(d_2,g_3),nE(d_2,d_3)\}}{nW(d_2)} \ar[ul] \\
  \pro:\argu{\{\neg uS(a_1,g_1,2,d_2)\}}{met(d_2,g_1)} \ar[u] && \pro:\argu{\{\neg uS(a_1,g_1,2,d_2),notMet(d_3,g_1)\}}{more(d_1,d_3)}\ar[u] }}
\end{scriptsize}
	\caption{\LeastAsm{} 
	dispute tree for
	$\argu{\{wDom(d_2)\}}{wDom(d_2)}$ for Example~\ref{exp:expDGW}. Here, $nM$, $nE$,  $nW$ and $uS$
  are short-hands for $notMet$, $notMore$, $notWDom$ and
  $unreachableSib$, \respectively.
\label{fig:expDGWTree2}
} 
\end{figure}
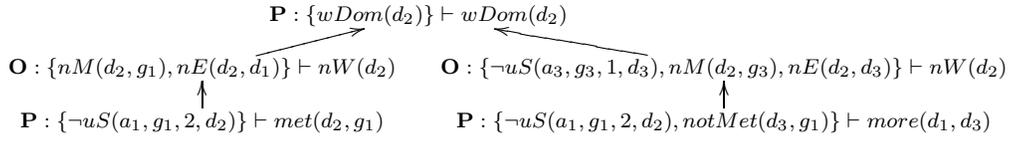

\begin{figure}[b]
\begin{scriptsize}
\centerline{
\xymatrix@1@R=10pt@C=40pt{
  & \pro:\argu{\{wDom(d_3)\}}{wDom(d_3)} \\
  \opp:A \ar[ur] & \opp:B \ar[u] & \opp:C \ar[ul] \\  
  \pro:D \ar[u]  & \pro:D \ar[u] & \pro:E \ar[u] }}

\medskip

\hspace{-5pt}\begin{tabular}{ll}
$A=\argu{\{\neg uS(a_1,g_1,2,d_2),nM(d_3,g_1),nE(d_3,d_2)}{nW(d_3)}$ & $D=\argu{\{\neg uS(a_3,g_3,1,d_3),notMet(d_2,g_3)\}}{more(d_3,d_2)}$\\
$B=\argu{\{\neg uS(a_1,g_2,1,d_2),nM(d_3,g_2),nE(d_3,d_2)}{nW(d_3)}$ & $E=\argu{\{\neg uS(a_3,g_3,1,d_3),notMet(d_1,g_3)\}}{more(d_3,d_1)}$\\
$C=\argu{\{nM(d_3,g_1),nE(d_3,d_1)\}}{nW(d_3)}$\\
\end{tabular}

\end{scriptsize}
	\caption{
		\LeastAsm{} 
		dispute tree for
	$\argu{\{wDom(d_3)\}}{wDom(d_3)}$ for Example~\ref{exp:expDGW}. Here, $nM$, $nE$,  $nW$ and $uS$
  are short-hands for $notMet$, $notMore$, $notWDom$ and
	$unreachableSib$, \respectively.
\label{fig:expDGWTree3}
} 
\end{figure}

\begin{proposition}
  \label{prop:expDGP}

  Let $AF$ be the \pfrSet{} ABA framework corresponding to a decision
  graph $\dg$.

  For any \pfrSet{} decision $d$ in $\dg$, let $\atree$ be a
	\leastAsm{} 
	dispute tree for $\argu{\pS(d)}{\pS(d)}$. Then, 
  the pair $\expPair{G}{M}$, where

  \begin{itemize}
    \item
	    $G=\{g|\pro:\argu{\_}{\{met(d,g)\}}$ is not 
		  ancestor of any
      $\opp$ leaf node in $\atree\}$ and

    \item
      $M =
      \{(S,d')|\pro:\argu{\{\metSet(d,S),notMet(d',\_)\}}{better(d,d',\_)}$
      is not 
		  ancestor of any $\opp$ leaf node in $\atree\}$
  \end{itemize}
	is a \flat{} explanation for $d$ being \pfrSet.

  For any non-\pfrSet{} decision $d$ in $\dg$, let $\atree$ be a best-effort
	dispute tree for $\argu{\pS(d)}{\pS(d)}$. Then,

  \begin{itemize}
    \item
	    $\{d'|\opp:\argu{\_,notBetter(d,d',\_)}{\notPS(d)}$ is 
		  ancestor of an $\opp$ leaf node in $\atree\}$ 
  \end{itemize}    
	is a \flat{} explanation for $d$ being not \pfrSet.  
\end{proposition}

\begin{example}
  \label{exp:expDGP}

  We illustrate Proposition~\ref{prop:expDGP} again with the DG
  shown in Figure~\ref{fig:expDGD}. However, we impose the
  preference relation $\{g_3\} \preferenceSet \{g_1, g_2\}$. As shown
  in Example~\ref{exp:expDGW}, both $d_2$ and $d_3$ are weakly
  dominant and $d_2$ meets $\{g_1,g_2\}$ and $d_3$ meets
  $\{g_3\}$. Thus, $d_3$ is preferred to $d_2$ so $d_3$ is a \pfrSet{}
  decision whereas $d_2, d_1$ are not. ($d_1$ is not even weakly
  dominant thus it cannot be \pfrSet.) 

	The ABA framework corresponding to this setting is given in Appendix~\ref{app:ABA}. 
With this ABA framework,	
  Figures~\ref{fig:expDGPTree1}, \ref{fig:expDGPTree2} and
  \ref{fig:expDGPTree3} show three 
  dispute trees for the
  arguments $\argu{\{\pS(d_1)\}}{\pS(d_1)}$,
  $\argu{\{\pS(d_2)\}}{\pS(d_2)}$ and $\argu{\{\pS(d_3)\}}{\pS(d_3)}$,
  \respectively, providing \dialogical\ explanations for $d_1$ and $d_2$ not being \pfrSet\ and for $d_3$ being \pfrSet, \respectively, and from which \flat\ explanations can be drawn as follows:
    \begin{itemize}
    \item
      $d_1$: $\{d_2, d_3\}$ is the \flat\ explanation for $d_1$ not being
      \pfrSet; this is obtained by extracting $d_2$ from the argument
      $C=\argu{\{mS(d_1,s12),nM(d_3,g_2),nB(d_3,d_1,s12)\}}{\notPS(d_3)}$
        in the node $\anode{\opp}{C}$, an ancestor of the node
        $\anode{\opp}{L}$, and
        $d_3$ from the argument
        $D=\argu{\{mS(d_3,s3),nM(d_2,g_3),nB(d_2,d_3,s3)\}}{\notPS(d_2)}$
        in the node $\anode{\opp}{D}$, an ancestor of the node
        $\anode{\opp}{M}$;                
      \item
        $d_2$: $\{d_3\}$ is the \flat\ explanation for $d_3$ not being
        \pfrSet; this is obtained by extracting 
        $d_3$ from the argument
        $D=\argu{\{mS(d_3,s3),nM(d_2,g_3),nB(d_2,d_3,s3)\}}{\notPS(d_2)}$
        in the node $\anode{\opp}{D}$, an ancestor of the node
        $\anode{\opp}{M}$;        
    \item
      $d_3$: $(\{g_3\}, \{(g_3,d_1),(g_3,d_2)\})$ is the \flat\
      explanation for $d_3$ being \pfrSet, extracted from arguments
      $G=\argu{\{\neg uS(a_3,g_3,1,d_3)\}}{met(d_3,g_3)}$ in the node
      $\anode{\pro}{G}$,
      $H=\argu{\{mS(d_3,s3),nM(d1,g3)\}}{b(d_3,g_1,s12)}$ in the node 
      $\anode{\pro}{H}$
      and $I=\argu{\{mS(d_3,s3),nM(d2,g3)\}}{b(d_3,g_2,s12)}$ in the
      node $\anode{\pro}{I}$.       
  \end{itemize}
\end{example}

\begin{figure}[t]
\begin{scriptsize}
\centerline{
\xymatrix@1@R=10pt@C=20pt{
  &&& \pro:\argu{\{\pS(d_1)\}}{\pS(d_1)} \\
  \opp:A \ar[urrr] & \opp:B \ar[urr] & \opp:C \ar[ur] && \opp:D \ar[ul] & \opp:E \ar[ull] & \opp:F \ar[ulll] \\
  \pro:G \ar[u]    & \pro:H \ar[u]   & \pro:I \ar[u]  && \pro:J \ar[u]  & \pro:H \ar[u]   & \pro:K \ar[u] \\
                   &                 & \opp:L \ar[u]  && \opp:M \ar[u]  &                 &  }}
\medskip
\begin{tabular}{ll}
$A=\argu{\{mS(d_2,s3),nM(d_1,g_3),nB(d_1,d_2,s3)\}}{\notPS(d_1)}$ & $H=\argu{\{\}}{met(d_1,g_1)}$\\
$B=\argu{\{mS(d_2,s12),nM(d_1,g_1),nB(d_1,d_2,s12)\}}{\notPS(d_1)}$ & $I=\argu{\{nM(d_2,g_2)\}}{nMS(d_2,s12)}$\\
$C=\argu{\{mS(d_2,s12),nM(d_1,g_2),nB(d_1,d_2,s12)\}}{\notPS(d_1)}$ & $J=\argu{\{nM(d_3,g_3)\}}{nMS(d_3,s3)}$\\
$D=\argu{\{mS(d_3,s3),nM(d_1,g_3),nB(d_1,d_3,s3)\}}{\notPS(d_1)}$ & $K=\argu{\{nM(d_3,g_2)\}}{nMS(d_3,s12)}$\\
$E=\argu{\{mS(d_3,s12),nM(d_1,g_1),nB(d_1,d_3,s12)\}}{\notPS(d_1)}$ & $L=\argu{\{\neg uS(a_1,g_2,1,d_2)\}}{met(d_2,g_2)}$\\
$F=\argu{\{mS(d_3,s12),nM(d_1,g_2),nB(d_1,d_3,s12)\}}{\notPS(d_1)}$ & $M=\argu{\{\neg uS(a_3,g_3,1,d_3)\}}{met(d_3,g_3)}$\\
$G=\argu{\{nM(d_2,g_3)\}}{nMS(d_2,s3)}$\\
\end{tabular}
\end{scriptsize}
	\caption{Best-effort 
	 dispute tree for
	$\argu{\{\pS(d_1)\}}{\pS(d_1)}$ for Example~\ref{exp:expDGP}. Here, $mS$, $nM$, $nB$, $nMS$ and
  $uS$ are short-hands for $\metSet$, $notMet$, $notBetter$,
  $\notMetSet$ and $unreachableSib$, \respectively.
\label{fig:expDGPTree1}
}
\end{figure}

\begin{figure}[t]
\begin{scriptsize}
\centerline{
\xymatrix@1@R=10pt@C=20pt{
  &&& \pro:\argu{\{\pS(d_2)\}}{\pS(d_2)} \\
  \opp:A \ar[urrr] & \opp:B \ar[urr] & \opp:C \ar[ur] && \opp:D \ar[ul] & \opp:E \ar[ull] & \opp:F \ar[ulll] \\
  \pro:G \ar[u]    & \pro:H \ar[u]   & \pro:I \ar[u]  && \pro:J \ar[u]  & \pro:H \ar[u]   & \pro:I \ar[u] \\
                   & \opp:K \ar[u]   &                && \opp:M \ar[u]  & \opp:K \ar[u]   &  \\
                   & \pro:L \ar[u]   &                &&                & \pro:L \ar[u] \\
}}
\medskip
\begin{tabular}{ll}
$A=\argu{\{mS(d_1,s3),nM(d_2,g_3),nB(d_2,d_1,s3)\}}{\notPS(d_2)}$ & $H=\argu{\{\neg uS(a_1,g_1,2,d_2)\}}{met(d_2,g_1)}$\\ 
$B=\argu{\{mS(d_1,s12),nM(d_2,g_1),nB(d_2,d_1,s12)\}}{\notPS(d_2)}$ & $I=\argu{\{\neg uS(a_1,g_2,1,d_2)\}}{met(d_2,g_2)}$\\
$C=\argu{\{mS(d_1,s12),nM(d_2,g_2),nB(d_2,d_1,s12)\}}{\notPS(d_2)}$ & $J=\argu{\{nM(d_3,g_3)\}}{\notMetSet(d_3,s3)}$\\
$D=\argu{\{mS(d_3,s3),nM(d_2,g_3),nB(d_2,d_3,s3)\}}{\notPS(d_2)}$ & $K=\argu{\{\neg reach(d_2,a_2)\}}{uS(a_1,g_2,1,d_2)}$\\
$E=\argu{\{mS(d_3,s12),nM(d_2,g_1),nB(d_2,d_3,s12)\}}{\notPS(d_2)}$ & $L=\argu{\{\}}{reach(d_2,a_2)}$\\
$F=\argu{\{mS(d_3,s12),nM(d_2,g_2),nB(d_2,d_3,s12)\}}{\notPS(d_2)}$ & $M=\argu{\{\neg uS(a_3,g_3,1,d_3)\}}{met(d_3,g_3)}$\\
$G=\argu{\{nM(d_1,g_3)\}}{\notMetSet(d_1,s3)}$\\
\end{tabular}
\end{scriptsize}
	\caption{Best-effort 
	 dispute tree for
  $\argu{\{\pS(d_2)\}}{\pS(d_2)}$ for Example~\ref{exp:expDGP}. Here, $mS$, $nM$, $nB$, $nMS$ and
  $uS$ are short-hands for $\metSet$, $notMet$, $notBetter$,
  $\notMetSet$ and $unreachableSib$, \respectively.
\label{fig:expDGPTree2}
}
\end{figure}
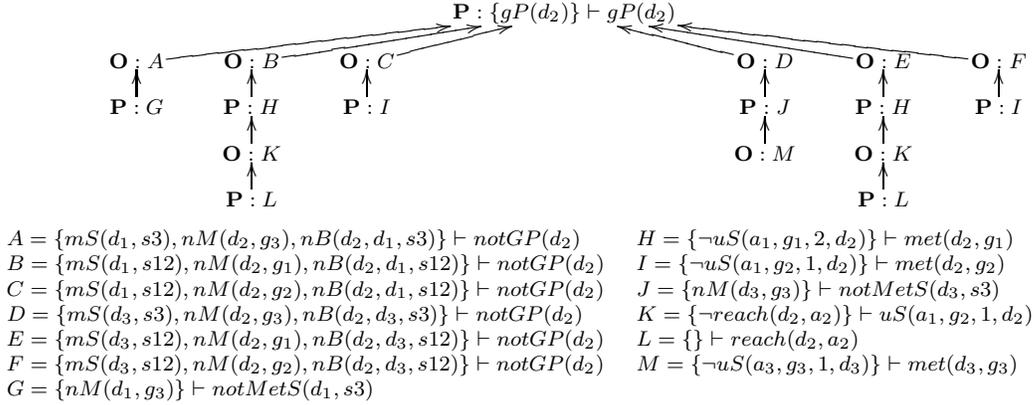

\begin{figure}[bht]
\begin{scriptsize}
\centerline{
\xymatrix@1@R=10pt@C=20pt{
  &&& \pro:\argu{\{\pS(d_3)\}}{\pS(d_3)} \\
  \opp:A \ar[urrr] & \opp:B \ar[urr] & \opp:C \ar[ur] && \opp:D \ar[ul] & \opp:E \ar[ull] & \opp:F \ar[ulll] \\
  \pro:G \ar[u]    & \pro:H \ar[u]   & \pro:H \ar[u]  && \pro:G \ar[u]  & \pro:I \ar[u]   & \pro:I \ar[u] \\
                   & \opp:J \ar[u]   & \opp:J \ar[u]  &&                & \opp:J \ar[u]   & \opp:J \ar[u] \\
                   & \pro:K \ar[u]   & \pro:K \ar[u]  &&                & \pro:K \ar[u]   & \pro:K \ar[u]
}}
\medskip
\begin{tabular}{ll}
$A=\argu{\{mS(d_1,s3),nM(d_3,g_3),nB(d_3,d_1,s3)\}}{\notPS(d_3)}$ & $G=\argu{\{\neg uS(a_3,g_3,1,d_3)\}}{met(d_3,g_3)}$\\
$B=\argu{\{mS(d_1,s12),nM(d_3,g_1),nB(d_3,d_1,s12)\}}{\notPS(d_3)}$ & $H=\argu{\{mS(d_3,s3),nM(d1,g3)\}}{b(d_3,g_1,s12)}$\\ 
$C=\argu{\{mS(d_1,s12),nM(d_3,g_2),nB(d_3,d_1,s12)\}}{\notPS(d_3)}$ & $I=\argu{\{mS(d_3,s3),nM(d2,g3)\}}{b(d_3,g_2,s12)}$\\ 
$D=\argu{\{mS(d_2,s3),nM(d_3,g_3),nB(d_3,d_2,s3)\}}{\notPS(d_3)}$ & $J=\argu{\{nM(d_3,g_3)\}}{\notMetSet(d_3,s3)}$ \\
$E=\argu{\{mS(d_2,s12),nM(d_3,g_1),nB(d_3,d_2,s12)\}}{\notPS(d_3)}$ & $K=\argu{\{\neg uS(a_3,g_3,1,d_3)\}}{met(d_3,g_3)}$ \\
$F=\argu{\{mS(d_2,s12),nM(d_3,g_2),nB(d_3,d_2,s12)\}}{\notPS(d_3)}$ 
\end{tabular}
\end{scriptsize}
	\caption{\LeastAsm{} 
	dispute tree for
  $\argu{\{\pS(d_2)\}}{\pS(d_2)}$ for Example~\ref{exp:expDGP}. Here, $mS$, $nM$, $nB$, $nMS$, $b$
  and $uS$ are short-hands for $\metSet$, $notMet$, $notBetter$,
  $\notMetSet$, $better$ and $unreachableSib$, \respectively.
\label{fig:expDGPTree3}
}
\end{figure}
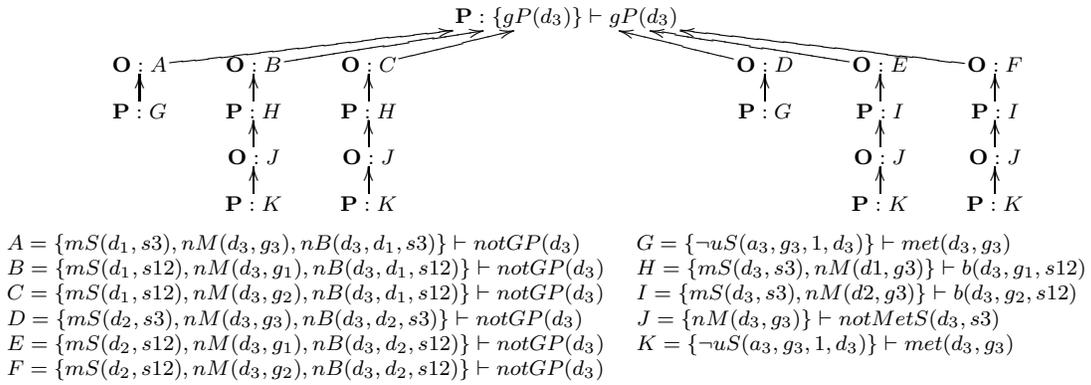

\section{Conclusion}
\label{sec:conclusion}

In this paper, we focused on two representations for
decision making, of different abstraction levels, as summarised in
Table~\ref{tableKRSum}
, and on a number of normative decision criteria for identifying ``good'' decisions therein, as summarised in Table~\ref{tableSum}.
For each decision making setting of interest (i.e. representation and criterion) we provided two forms of explanation: \emph{\flat} explanations, based solely on justifying ``good'' decisions in terms of the ingredients of the underlying representation, and \emph{\dialogical} explanations, based on spelling out the reasoning underpinning the considered decision criterion, while also incorporating the same information present in \flat\ explanations.  
To define \dialogical\  explanations, we used  
Assumption-based argumentation (ABA) as an equivalent representation (to the
representation for decision making and decision criterion).
Specifically, in the ABA reformulations, ``good'' decisions are claims of \emph{admissible arguments}
and vice versa. 
The \dialogical\ explanations are special (and novel) forms of dispute trees in ABA: \emph{\leastAsm\ dispute trees} (a restricted form of admissible dispute trees in ABA) for explaining why decisions are deemed ``good'', and \emph{best-effort dispute trees} (a restricted form of maximal, non-admissible dispute trees in ABA)  for explaining why decisions are not deemed ``good''.

\begin{table}
\begin{center}
\begin{tabular}{|l|}
\hline
Abstract Decision Frameworks (ADFs): \nadf{}  \\
$\bullet$  amounting to decisions $\decisions$, goals $\goals$ and a decision-meets-goal relation \dmg. \\
\hline
Preferential Decision Frameworks (PDFs): \neadfps{} \\
$\bullet$  amounting to ADFs with preferences over goals. \\
\hline
Decision Graphs (DGs): \edgTuple{} \\
	$\bullet$  amounting to relations ($\edge$) between decisions, goals, and intermediates (all in $\node$), \\
	\phantom{$\bullet$} with \bSet\ \info. \\
\hline
	\PDG s (PDGs): $\pdgTuple$ \\
$\bullet$ amounting to DGs  with preferences over goals.\\
\hline
\end{tabular}
\end{center}
\caption{Overview of decision making representations studied in this
  paper.}
\label{tableKRSum}
\end{table}

\begin{table}
\begin{center}
\begin{tabular}{|l|l|}
\hline
{\bf Decision Criterion} & {\bf Brief Description} \\
\hline
Strongly Dominant & Select decisions meeting all goals. \\
Dominant          & Select decisions meeting all goals ever met. \\
Weakly Dominant   & Select decisions meeting goals not met by others. \\
\PfrSet{}         & Select decisions meeting the most preferred goal sets. \\
\hline
\end{tabular}
\end{center}
\caption{Overview of decision criteria studied in this paper.}
\label{tableSum}
\end{table}

Our paper opens several avenues for future work, including those briefly outlined next.

Our decision settings (i.e. representations and criteria) correspond to some existing decision making frameworks (as shown in Appendix~\ref{sec:relation}); thus, our novel forms of explanation can be used to explain (``good'' or ``bad'') decisions in those frameworks as well. It would be interesting to see whether our approach could be adapted to provide explanations in other settings, e.g. when preferences over decisions or, in the case of decision graphs, intermediates are present or  in decision settings not restricted to a single fixed world state, such as 
decision making problems studied with Markov decision
processes \cite{Bellman57} or Sequential Decision Making
\cite{Barto89}, with  constantly evolving world
states. 

Our \dialogical\ explanations make use of novel, restricted types of dispute trees. It would be interesting to study their stand-alone properties in ABA and their usefulness in other settings where ABA or its instances are used for explanation via forms of dispute trees (e.g. as in \cite{TR}).

In the case of decision graphs, we have focused on \flat\ explanations matching exactly (by instantiating) those for ADFs and PDFs. However, as already mentioned at the end of Section~\ref{sec:DMwithDGs},
``deeper'' \flat\ explanations could be defined 
whereby the reasons for goals being met, in terms of reachability of goals from decisions in the graphs, are accommodated within the explanations themselves. 

We have defined 
\dialogical\ explanations as dispute trees, providing a range of intuitive readings thereof as disputes between fictional proponent and opponent players, in the spirit of dispute derivations in ABA (e.g. as
defined in \cite{aij12-XDDs}). Alternative readings may also be possible, for example as summaries of the salient points in the reasoning leading to sanctioning selected decisions (as ``good'' or ``bad''). Also, we leave as future work the automated extraction of these readings from templates, e.g. in the spirit of \cite{scheduling,IEs}.    

We have defined \dialogical\ explanations as static dispute trees. However, fragments of these trees could be extracted providing partial information on demand within interactive exchanges with the users, in the spirit of \cite{IEs}. This could be also useful to unearth, when decision graphs are used, the reasoning leading to establishing that a decision meets a goal (whereas in our current \dialogical\ explanations this reasoning is ``hidden'' within arguments which are nodes of dispute trees).

Finally, as mentioned at the end of Section~\ref{sec:ABAwithDGs}, our ABA formulations can be seens as decision making abstractions in their own right, and thus provide forms of argumentation-based decision making.  We leave these more general forms of argumentation-based decision making as future work.

\todo{add acks at some stage - for CR?}

\bibliography{bibliography}
\bibliographystyle{theapa}

\appendix

\section{Relations with Decision Making Methods}
\label{sec:relation}
In this section, we compare decision making with ADFs, DGs and several
other decision making methods. We show correspondences between several
of the proposed decision making criteria and existing decision making
methods in the literature.

In this appendix, in line with existing works, we use a tabular representation $\Tdg$ of $\dmg$ in ADFs and PDFs, whereby $\Tdg$ is a table of size $n \times l$ and for all $d \in
\decisions$, $g \in \goals$,
$\Tdgi{d}{g}=1$ 
\ifaf{} $g \in \dmg(d)$, and $\Tdgi{d}{g}=0$
otherwise.

\subsection{
  Strongly Dominant Decisions and 
  Multi-Attribute Decision Making}
\label{sec:MADM}
Multi-Attribute Decision Making (MADM) refers to decision making 
over available alternatives characterised by multiple, usually
conflicting, attributes \cite{Hwang81}. Here we show that our approach
models certain MADM problems well. These are MADM problems where
information is presented directly on attributes, equipped with {\em
  non-compensatory methods}, where, quoting from \cite{Yoon95}: 
\begin{quote}
A compensatory or non-compensatory distinction is made on the basis of
whether advantages of one attribute can be traded for disadvantages of
another or not. A choice strategy is compensatory if trade-offs among
attribute values are permitted, otherwise it is non-compensatory.
\end{quote}

In particular, MADM under the (non-compensatory) \emph{conjunctive
  method} \cite{Yoon95} can be formulated as follows. There is a set
of {\em alternatives} (\CNJalternatives) \st{} each $l \in
\CNJalternatives$ can be represented as a set of {\em attributes}
$\CNJattributes_l$. There is a set of requirements 
$\CNJrequirements$, \st{} each attribute $t$ can be tested against one
requirement $r_t \in \CNJrequirements$ ($r_t$ is the {\em matching
  requirement} for $t$). There is a boolean {\em
  requirement function} $\boolFunc$ which maps alternatives and
requirements to 1s and 0s. For an alternative $l$ with attributes
$\CNJattributes_l$, $l$ is selected by the conjunctive method \ifaf{}
for all attributes $t \in \CNJattributes_l$ and $r_t \in
\CNJrequirements$ ($r_t$ is the matching requirement for $t$),
$\boolFunc(t,r_t) = 1$, i.e. the conjunctive method
selects the alternatives whose attributes meet all requirements. 
For lack of a better word, we use the term {\em conjunctive framework}
to denote the tuple \cnjFmk{}. 

The following example illustrates conjunctive frameworks and the
conjunctive method by adapting an example from \cite{Yoon95}. 

\begin{example}
\label{adminExp}
A graduate school screens international applicants for admission. To
pass the screening, an applicant whose native language is not English
must meet the minimum scores on three tests: TOEFL, GRE and
GPA. Table~\ref{studentAdmissionTable} indicates the tests' results  
for five students. This decision problem can be represented as a
conjunctive framework with $\CNJalternatives=\{A_1, \ldots, A_5\}$,  
for any $i=1, \ldots, 5$, $\CNJattributes_{A_i}$ is the set of 
assignments to TOEFL, GRE, GPA given in  Table~\ref{studentAdmissionTable}
(e.g.  $\CNJattributes_{A_1}=\{ TOEFL=582, GRE=1420, GPA=2.8\}$),
and $\CNJrequirements=\{TOEFL\geq 552, GRE\geq 1200, GPA\geq 3.0\}$. Finally, 
the matching requirement for $T=v$ is $T \geq min_T$, for any $T$
amongst $TOEFL, GRE, GPA$, with $min_T$ as in
Table~\ref{studentAdmissionTable}, and $\boolFunc(T=v,T \geq min_T)=
1$ iff $v \geq min_T$. Given this conjunctive framework \cnjFmk, the
conjunctive method selects students $A_2$, $A_4$, $A_5$, whereas, 
quoting from \cite{Yoon95}:
\begin{quote}
Students $A_1$ and $A_3$ are rejected due to their low GPA and GRE
scores, respectively. Note that $A_1$ was rejected even though the
candidate has a very high GRE score.
\end{quote}

\begin{table}
\scriptsize
\begin{center}
\begin{tabular}{lccc}
\hline
Students & TOEFL & GRE  & GPA \\
\hline
\rowcolor{Gray}
$A_1$    & 582   & 1420 & 2.8 \\
$A_2$    & 563   & 1250 & 3.5 \\
\rowcolor{Gray}
$A_3$    & 620   & 1080 & 3.2 \\
$A_4$    & 558   & 1280 & 3.0 \\
\rowcolor{Gray}
$A_5$    & 600   & 1210 & 3.6 \\
\hline
Minimum & 550    & 1200 & 3.0 \\
\hline
\end{tabular}
\end{center}
\caption{Graduate School Admission for International Students
  \cite{Yoon95}.} 
\label{studentAdmissionTable}
\end{table}
\end{example}

We can model this problem also using a decision framework $\adf$, with 
$\decisions = \{A_1, \ldots,$ $A_5\}$, $\goals = \{TOEFL\geq 550,
GRE\geq 1200, GPA\geq 3.0\}$ and $\Tdg$ given in Table~\ref{adminTga}. 
Then, $A_2$, $A_4$ and $A_5$ are strongly dominant decisions as they
meet all three goals, and no other decision is strongly dominant.
\begin{table}
\scriptsize
\begin{center}
\begin{tabular}{lccc}
\hline
 & TOEFL $\geq$ 550  & GRE $\geq$ 1200  & GPA $\geq$ 3.0 \\
\hline
\rowcolor{Gray}
$A_1$    & 1   & 1  & 0  \\
$A_2$    & 1  & 1  & 1  \\
\rowcolor{Gray}
$A_3$    & 1  & 0  & 1  \\
$A_4$    & 1  & 1  & 1  \\
\rowcolor{Gray}
$A_5$    & 1  & 1  & 1  \\
\hline
\end{tabular}
\end{center}
\caption{\Tdg{} in the ADF for Example~\ref{adminExp}. Here, $\Tdgi{d_i}{g_j} = 1$
  \ifaf{} $d_i$ meets the minimum requirement for $g_j$,
  e.g. $\Tdgi{A_1}{TOEFL} = 1$ as the TOEFL score of student $A_1$ is
  582 and the minimum requirement is 550.}
\label{adminTga}
\end{table}

In general, there is a mapping from any conjunctive framework to an
``equivalent'' ADF such that the alternatives selected by the
conjunctive method in the former are strongly dominant decisions in
the latter. Formally:

\begin{proposition}
\label{prop:cnj}
Given a conjunctive framework $\cnjFmk$, let $\adf$ be \st{}
$\decisions=\CNJalternatives$, $\goals = \CNJrequirements$, 
$\Tdgi{l}{r} = 1$ if $\boolFunc(t,r) = 1$, for $r$ the matching
requirement of $t$ and $t$ attribute of $l$, $\Tdgi{l}{r} = 0$,
otherwise. Then $\adf$ is an ADF and the set of all strongly dominant
decisions in $\adf$ is the set of alternatives in $\cnjFmk$ selected
by the conjunctive method. 
\end{proposition}

The other direction of Proposition~\ref{prop:cnj} holds as well, as
follows.

\begin{proposition}
  \label{prop:cnj2}
Given an ADF $\adf$, let $\cnjFmk$ be \st{} $\CNJalternatives =
\decisions$, $\CNJrequirements = \goals$, $\boolFunc(t,r) = 1$, for
$r$ the matching requirement of $t$ and $t$ an attribute of $l$, if
$\Tdgi{l}{r} = 1$; $\boolFunc(t,r) = 0$, otherwise. Then $\cnjFmk$ is
a conjunctive framework and the set of alternatives in $\cnjFmk$
selected by the conjunctive method is the set of strongly dominant
decisions in $\adf$.
\end{proposition}

\subsection{Weakly Dominant Decisions and Pareto Optimisation}
\label{po}
 {\em Efficiency} in Pareto optimisation is one of the criteria used
 in classical decision making \cite{Emmerich06}. In this section, we
 show that {\em weakly dominant} decisions correspond to {\em
   efficient} solutions in Pareto optimisation.

We first summarise the notion of efficient solutions in Pareto
optimisation \cite{Emmerich06}. Two spaces are considered: the {\em
  decision space} \dSpace\ and the {\em objective space} \oSpace, with 
$\prec$ an order over $\oSpace$. A vector-valued objective function
$\ofunc : \dSpace \mapsto \oSpace$ provides a mapping from the decision
space to the objective space. The set of {\em feasible solutions} $\fSln$
is a subset of the decision space, and $\yImg$ is the image of $\fSln$
under $\ofunc$. The {\em Pareto front} $\yImg_N$ is defined as the set of
\emph{non-dominated solutions} in $\yImg = \ofunc(\fSln)$, i.e. $\yImg_N =
\{\mb{y} \in \yImg | \nexists \mb{y}' \in \yImg : \mb{y}' \prec
\mb{y}\}$. Finally, the {\em efficient set} is defined as the pre-image of the
Pareto front, $\fSln_E = \ofunc^{-1}(\yImg_N)$, and a decision is an
efficient solution if it belongs to the efficient set.  

We use the term {\em Pareto framework} to
denote the tuple \paretoFramework{} consisting of a decision
space \dSpace, an objective space \oSpace, an objective function
\ofunc, and an order $\prec$ over the objective space. The following
example illustrates Pareto frameworks and efficient solutions by
recasting the decision problem of Example~\ref{expWD}. 
\begin{example}
Let the decision space \dSpace{} be $\{jh, ic, ritz\}$ and the
objective space \oSpace{} be $2^{\{cheap, near\}}$, with $\prec$
defined as $\supset$. Also, let the objective function \ofunc\ be such
that:

\hspace{20pt}$\ofunc(jh) = \{near\}$,
\hspace{40pt}$\ofunc(ic) = \{cheap\}$,
\hspace{40pt}$\ofunc(ritz) = \{\}$.

Assume that the set feasible solutions $\fSln$ is the same as \dSpace. 
Since $\yImg = \ofunc(\fSln)$, $\yImg = \{\{near\}, \{cheap\}, \{\}\}$. 
By letting $\prec$ be $\supset$, we have $\ofunc(jh) \prec \ofunc(ritz)$ and
$\ofunc(ic) \prec \ofunc(ritz)$. Thus, the Pareto front $\yImg_N$ is
$\{\{near\}, \{cheap\}\}$, and the efficient set is $\{jh, ic\}$. 
As a result, both $jh$ and $ic$ are efficient solutions.
\end{example}

In the example, weakly dominant decisions correspond to efficient
solutions. This is the case in general, in the sense that there is a
mapping from any ADF to an ``equivalent'' Pareto framework  
\st{} weakly dominant decisions in the former are efficient
solutions in the latter (see Table~\ref{table:pareto} for a full
overview of the mapping). 
\begin{table}
\begin{center}
\begin{tabular}{|ll|}
\hline
Our notions                    & Pareto Optimisation   \\
\hline
\rowcolor{Gray}
ADFs                           & Pareto Frameworks \\
decisions (\decisions)         & decision space (\dSpace)   \\ 
\rowcolor{Gray}
sets of goals ($2^\goals$)      & objective space (\oSpace)   \\
goals met by decisions (\dmg)  & objective function (\ofunc) \\
\rowcolor{Gray}
	set of all weakly dominant decisions ($S_w$) & efficient set ($\fSln_{E}$) \\
goals met by weakly dominant decisions  & Pareto front ($\yImg_N$) \\
\hline
\end{tabular}
\caption{From weak dominance in ADFs to Pareto Optimisation.\label{table:pareto}} 
\end{center}
\end{table}
Formally:

\begin{proposition}
\label{prop:pareto}
Given an ADF $\adf$ with $\dmg$ as in Definition~\ref{dfn:adf}, let
$\paretoFramework$ be \st{} $\dSpace = \decisions, \oSpace = \goals,
\ofunc = \dmg, \prec = \supset$. Then,  $\paretoFramework$ is a Pareto
framework and $\{d \in \decisions| d$ is weakly dominant in $\adf\}$
is the efficient set of $\paretoFramework$. 
\end{proposition}

The other direction of Proposition~\ref{prop:pareto} holds as well, as
follows.
\begin{proposition}
\label{prop:pareto2}

Given a Pareto Framework $\paretoFramework$ with $\prec = \supset$,
let $\adf$ be an ADF with $\decisions = \dSpace, \goals = \oSpace,
\Tdg$ be \st{} $\Tdgi{d}{g} = 1$ \ifaf{} $g \in \ofunc(d)$. Then, the
efficient set of $\paretoFramework$ is 
	the set of all weakly dominant
decisions in $\adf$.
\end{proposition}

\subsection{\PfrGoal{} Decisions and the Lexicographic Method}
\label{sec:pfrGoalLexico}

One MADM method for selecting decisions with preference is the
{\em Lexicographic method}, where, quoting from \cite{Yoon95}:

\begin{quote}
  [The lexicographic method] uses one lexicographic attribute at a
  time to examine alternatives for elimination. After all the
  alternatives are examined, if more than one still remains, the
  process is repeated using another lexicographic attribute. $\ldots$
  The Lexicographic method examines alternatives in the order of
  attribute importance. 
\end{quote}

MADM under the Lexicographic method \cite{Yoon95} can be formulated as
follows. There is a set of {\em alternatives} (\LexAlternatives) and
each alternative has some {\em lexicographic attributes}
(\LexAttributes) from the set $\LexAttributes = \{x_i, \ldots,
x_m\}$. Let $x_1 \in 
\LexAttributes$ be the most important lexicographic attribute, $x_2
\in \LexAttributes$ be the second most important one, and so on. Then 
a set of alternatives $\LexAlternatives^1$ is selected \st{} 
\begin{equation*}
  \LexAlternatives^1 = \{a_i | \underset{i}{{\rm max}} \
  a_i \mbox{ has } x_1\}, \mbox{ for } a_i \in \LexAlternatives. 
\end{equation*}
If $\LexAlternatives^1$ is a singleton, then the alternative in
$\LexAlternatives^1$ is the most preferred alternative. If there
remain multiple alternatives, consider  
\begin{equation*}
  \LexAlternatives^2 = \{a_i | \underset{i}{{\rm max}} \
  a_i \mbox{ has } x_2\}, \mbox{ for } a_i \in \LexAlternatives^1. 
\end{equation*}
If $\LexAlternatives^2$ is a singleton, then stop and select the
alternative in $\LexAlternatives^2$. Otherwise continue this process
until either some singleton $\LexAlternatives^k$ is found or all
lexicographic attributes have been considered. If the final set
contains more than one alternative, then they are considered to be
equivalent. In this process, if for some lexicographic attribute $x$
there is no alternative has $x$, then no alternative is eliminated in
this iteration and the next important lexicographic attribute is
examined. 

\begin{example}
  \label{exp:lex}

  We again consider the decision making problem shown in
\FAN{Example~\ref{DGExpS}}. 
  We let the set of alternatives be $\{d_1,
  d_2\}$ and the set of lexicographic attributes be $\{g_1, g_2, g_3,
  g_4, g_5\}$. We let the importance amongst lexicographic attributes
  be $g_1 > g_2 > g_3 > g_4 > g_5$. The lexicographic method works as
  follows.

  Since neither $d_1$ nor $d_2$ has $g_1$, no alternative is
  eliminated in the first iteration.

\begin{equation*}
  \LexAlternatives^1 = \{d_1, d_2\}. 
\end{equation*}

  Then, we move to $g_2$, the second most important lexicographic
  attribute, since both of the two alternatives $d_1$ and $d_2$ have
  $g_2$, we have:

\begin{equation*}
  \LexAlternatives^2 = \{d_i | \underset{i}{{\rm max}} \
  d_i \mbox{ has } g_2\}, \mbox{ for } d_i \in \{d_1, d_2\} = \{d_i |
  d_i \mbox{ has } g_2\} = \{d_1, d_2\}. 
\end{equation*}
Then we move to the third important lexicographic attribute,
$g_3$. Since only $d_2$ has $g_3$, we have

\begin{equation*}
  \LexAlternatives^3 = \{d_i | \underset{i}{{\rm max}} \
  d_i \mbox{ has } g_3\}, \mbox{ for } d_i \in \{d_1, d_2\} = \{d_i |
  d_i \mbox{ has } g_3\} = \{d_2\}. 
\end{equation*}
Since $\LexAlternatives^3$ is a singleton set containing only $d_2$,
the lexicographic method terminates with $d_2$ being the selected
alternative. 
\end{example}

We can see that our modelling of preference ranking over goals
generalises the Lexicographic method, formally:

\begin{proposition}
  \label{prop:lexToPrefG}

  Given a set of alternatives \LexAlternatives{} and a set of
  lexicographic attributes \LexAttributes{}, let \eadfps{} be an PDF
  with $\decisions = \LexAlternatives$, $\goals = \LexAttributes$,
  $\Tdgi{d}{g} = 1$ \ifaf{} the alternative corresponds to $d$ has the
  attribute corresponds to $g$, and $\preferenceSet$ be \st{} $\{g_1\}
  \preferenceSet \{g_2\}$ for $g_1, g_2 \in \goals$ \ifaf{} the
  attribute corresponds to $g_1$ is more important than the attribute
  corresponds to $g_2$. Then alternatives selected by the
  Lexicographic method is the set of \pfrSet{} decisions.
\end{proposition}

The other direction of Proposition~\ref{prop:lexToPrefG} holds as
well. 

\begin{proposition}
  \label{prop:prefGToLex}

  Given an PDF \eadfps{} \st{} $\preferenceSet$ is a total order
  over $\{\{g\}|g \in \goals\}$, if we let
  $\LexAlternatives=\decisions$, $\LexAttributes=\goals$ and the
  importance of each $a \in \LexAttributes$ as described by
  $\preferenceSet$, then \pfrSet{} decisions in the PDF are 
  alternatives selected by the Lexicographic method with alternatives
  $\LexAlternatives$ and attributes $\LexAttributes$.
\end{proposition}

\pfrSet{} decisions selected in DGs also correspond to
decisions selected with the Lexicographic method.

\begin{proposition}
  \label{prop:Lexico2PDG}

  Given a set of alternatives \LexAlternatives{} and a set of
  lexicographic attributes \LexAttributes{}, let \pdgTuple{} be an
  PDG with $\node = \dNode \cup \iNode \cup \gNode$ \st{}
  \begin{itemize}
  \item
    $\dNode = \LexAlternatives$,

  \item
    $\iNode = \{\}$,

  \item
    $\gNode = \LexAttributes$,

  \item
    $\edge$ be \st{} for $a \in \LexAlternatives$, $x \in
    \LexAttributes$, $\egPair{a}{x} \in \edge$ \ifaf{} $a$ has $x$,

  \item
    $\info = \{\}$,

  \item
    $\preferenceSet$ be \st{} $\{g_1\} \preferenceSet \{g_2\}$ for
    $g_1, g_2 \in \gNode$ \ifaf{} the lexicographic attribute
    corresponds to $g_1$ is more important than the lexicographic
    attribute corresponds to $g_2$. 
  \end{itemize}

  Then alternatives selected by the Lexicographic method is the set of
  \pfrSet{} decisions in PDG.
\end{proposition}

\begin{proposition}
  \label{prop:PDG2Lexico}

  Given an PDG \pdgTuple with $\node = \dNode \cup \iNode \cup
  \gNode$, \st{} $\preferenceSet$ is a total order 
  over $\{\{g\}|g \in \gNode\}$, if we let
  \begin{itemize}
    \item 
      $\LexAlternatives=\dNode$;

    \item
      $\LexAttributes=\gNode$;

    \item
      an alternative $a \in \LexAlternatives$ has an lexicographic
      attribute $x \in \LexAttributes$ \ifaf{} $x$ can be reached from
      $\{a\}$ in \pdgTuple; and 

    \item    
      the importance of each $x \in \LexAttributes$ is \st{} for $x_1,
      x_2 \in \LexAttributes$, $x_1$ is less important than $x_2$
      \ifaf{} $\{x_1\} \preferenceSet \{x_2\}$;
  \end{itemize}
  then \pfrSet{} decisions in the PDF are alternatives selected by
  the Lexicographic method with alternatives $\LexAlternatives$ and
  attributes $\LexAttributes$.
\end{proposition}

\subsection{Proofs for all results in Appendix~\ref{sec:relation}}
\paragraph{Proof of Proposition~\ref{prop:cnj}}
\adf\ is an ADF by construction and Definition~\ref{dfn:adf}.
It is easy to see 
for each alternative $l$,
$l$ is selected in $C$ \ifaf{} $l$ satisfies all requirements
$\CNJrequirements$. This implies that each $d \in \decisions$ meets
all goals in $\goals$. Thus, by definition of strong dominance,
each selected alternative is a strongly dominant decision. \qed

\paragraph{Proof of Proposition~\ref{prop:cnj2}}
To show strongly dominant decisions are the set of alternatives
selected by the conjunctive method, we only need to show that all
dominant decisions are alternatives meeting all requirements. This is
easy to see as a strongly dominant decision meets all goals in its
corresponding ADF (Definition~\ref{dfn:SD}). Thus strongly 
dominant decisions are alternatives selected by the conjunctive
method. \qed

\paragraph{Proof of Proposition~\ref{prop:pareto}}
Follows trivially from the definitions. \qed

\paragraph{Proof of Proposition~\ref{prop:pareto2}}
Let $D$ be the efficient set. We need
to show that there is no decision $d\in D$ meeting strictly fewer goals
than some other decision $d' \in \decisions$. This is the
case as by definition of efficient set: since efficient decisions
meet goals in in the Pareto front, there is no decision meeting strictly
more goals than an efficient decision. Therefore efficient decisions
are weakly dominant.

Then the other direction. Let $D$ be the set of all weakly dominant
decisions, by Definition~\ref{dfn:wd}, there is no $d \in D$ meeting
strictly fewer goals than some $d' \in \decisions$. Thus, the set
$S=\{\dmg(d)|d \in D\}$ is the Pareto front and for each element $E$ in
$S$, $\dmg^{-1}(E)=d$ is an efficient solution.

\paragraph{Proof of Proposition~\ref{prop:lexToPrefG}}
To show that the set of alternatives selected by the lexicographic
method are \pfrSet{} decisions is to show that each selected
decision $d$ is (1) weakly dominant, and (2) goals met by other
decisions are no more preferred. 

(1) holds as if $d$ is not weakly dominant, then there
exists a decision $d'$ such that $d'$ meets more goals (\wrt{}
$\subseteq$) than $d$. This cannot be the case as if such $d'$ exists,
then the Lexicographic method would select $d'$ instead upon examining
some goal $g$ met by $d'$ but not $d$.

(2) holds as if there exists $d'$ meeting goal $g$ not met by $d$ and
$d$ meets no other goal $g'$ preferred to $g$, then $d$ will not be
selected by the Lexicographic method as upon examining decisions
meeting $g$, $d$ would be eliminated. \qed

\paragraph{Proof of Proposition~\ref{prop:prefGToLex}}
It is easy to show that the Lexicographic method selects \pfrSet{}
decisions if $\preferenceSet$ is a total order over $\{\{g\}|g \in
\goals\}$. Since the Lexicographic method examines lexicographic
attributes in the order of their importance, for any \pfrSet{}
decision $d$, $d$ exists in all $\LexAlternatives^i$. \qed

\paragraph{Proof of Proposition~\ref{prop:Lexico2PDG}}
As shown in the proof of Proposition~\ref{prop:lexToPrefG}, \pfrSet{}
decisions in ADFs can be used to model alternatives selected by the
lexicographic method. The construction of PDG in
Proposition~\ref{prop:Lexico2PDG} is given in the same spirit as the
ADF construction given in Proposition~\ref{prop:lexToPrefG}. Since DG
generalises ADF (Proposition~\ref{prop:ADFinstanceDG}), it is easy to
see that the correspondance between lexicographic method and \pfrSet{}
decisions in PDG holds. \qed

\paragraph{Proof of Proposition~\ref{prop:PDG2Lexico}}
As shown in the proof of Proposition~\ref{prop:prefGToLex}, \pfrSet{}
decisions in ADF can be mapped to selected alternatives by the
lexicographic method, as DG generalises ADF
(Proposition~\ref{prop:ADFinstanceDG}), it is easy to see that
\pfrSet{} decisions in PDG can also be mapped to selected alternatives
by the lexicographic method. \qed

\section{Proofs for the Results in Sections~\ref{sec:new}--\ref{sec:dg}}
\label{sec:proof}
\paragraph{Proof of Proposition~\ref{oneDominantSet}} Follows 
from Definition~\ref{dfn:d}. 

\paragraph{Proof of Proposition~\ref{sdISd}}
Follows from Definition~\ref{dfn:d}.
%

\paragraph{Proof of Proposition~\ref{prop:allEqual}} 
First we prove that if $S_{s} \neq \{\}$, then $S_{s} = S_{d}$. By
Proposition~\ref{sdISd}, $S_{s} \subseteq S_{d}$. We show that there
is no $d$ such that $d \in S_{d}$, $d \notin S_{s}$. Assuming
otherwise, (1) since $d \notin S_{s}$, $\dmg(d) \neq \goals$, hence
there is some $g \in \goals$ and $g \notin \dmg(d)$; (2) since $S_{s}
\neq \{\}$, there is $d' \in S_{s}$ such that $\dmg(d') = \goals$,
therefore $g \in \dmg(d')$. By (1) and (2), $d \notin
S_{d}$. Contradiction. 

Then we prove that $S_{s} = S_{w}$. Assume by contradiction that 
$S_{s} \subset S_{w}$. 
Then,
there exists $d \in S_{w}$, $d \notin S_{s}$. Since $S_{s} \neq \{\}$,
there exists $d' \in S_{s}$ and $\dmg(d') = \goals$. Since $d
\notin S_{s}$, $\dmg(d) \subset \goals$. Hence $\dmg(d) \subset
\dmg(d')$. Then, by Definition~\ref{dfn:wd}, $d \notin
S_{w}$. Contradiction. \qed

\paragraph{Proof of Proposition~\ref{prop:dEQw}}
By Proposition~\ref{dISwd}, we know that $S_{d} \subseteq S_{w}$. We
need to 
show $S_{w} \subseteq S_{d}$. Assume otherwise, i.e. there exists $d
\in S_{w}$ and $d \notin S_{d}$. Since $S_{d} \neq \{\}$, there
exists $d' \in S_{d}$, such that $\dmg(d') \supseteq \dmg(d'')$, for
all $d'' \in \decisions$. Hence $\dmg(d) \subseteq \dmg(d')$. 
Since $d \notin S_{d}$, $\dmg(d) \neq \dmg(d')$. Therefore, $\dmg(d)
\subset \dmg(d')$. Then, by Definition~\ref{dfn:wd}, $d \notin
S_{w}$. Contradiction. \qed

\paragraph{Proof of Theorem~\ref{twoSetWD}}

We will use the following lemma:
\begin{lemma}
\label{allSameDominant}
If, for all $d,d' \in \decisions$, $\dmg(d) =
\dmg(d')$, then 
the set of all dominant decisions is $\decisions$.
\end{lemma}
\noindent {\bf Proof:}
Since for all $d,d'
\in \decisions$, $\dmg(d) = \dmg(d')$, then there is no $g \in \goals$
such that $g \in \dmg(d)$ and $g \notin \dmg(d')$.
Hence, by Definition~\ref{dfn:d}, all $d \in \decisions$ are
dominant
. \qed
\\
\\
We now prove the theorem.
Since $
S_{d} = \{\}$, by Lemma~\ref{allSameDominant},
$|\decisions| > 1$ and 
$|S_w|>1$. 
Assume that for all $d, d' \in S_{w}$, with $d \neq d',
\dmg(d) = \dmg(d')$. Then there are two cases, both 
leading to contradiction:
\begin{enumerate}
\item
there is no $d'' \in \decisions \setminus S_{w}$. Then
$S_{w} = \decisions$, and, by the absurd assumption, 
$\dmg(d) = \dmg(d')$ for all $d, d' \in \decisions$. Then, by
Lemma~\ref{allSameDominant}, for all $d \in S_{w}, d \in 
S_d$.
But we also have $S_{d} = \{\}$, hence
contradiction.

\item
there exists some $d'' \in \decisions \setminus S_{w}$. 
Then there are five possibilities
, and they all give contradictions, as follows:

\begin{enumerate}
\item
$\dmg(d) \supset \dmg(d'')$. Not possible, as if so there
 would exists $d^* \in \decisions$ such that $d^*$ is dominant (with $d$
 a possible such $d^*$). 

\item
$\dmg(d) \subset \dmg(d'')$. Not possible, as if so there
 would exists $d^* \in \decisions \setminus S_{w}$ such that $d^*$ is
 dominant ($d''$ would be a possible such $d^*$).

\item
$\dmg(d) = \dmg(d'')$. Not possible, as if so $d''$ would be in
  $S_{w}$.

\item
None of (a)(b)(c) but $\dmg(d) \cap \dmg(d'') \neq \{\}$. 
Not possible, as if so there would exist $g \in \dmg(d''), g
\notin \dmg(d)$, hence there would exist $d^* \in \decisions 
\setminus S_{w}$ with $d^*$ weakly dominant ($d''$ would be a
possible such $d^*$), but
\FAN{$S_{w}$ is the set of weakly dominant decisions}
and $d^* \notin 
S_{w}$.

\item
None of (a)(b)(c), but $\dmg(d) \cap \dmg(d'') = \{\}$. Same as
case 2(d).

\end{enumerate}
\end{enumerate}

Both cases 1 and 2 give contradictions, and thus the theorem holds. \qed

\paragraph{Proof of Proposition~\ref{prop:expProperty}}
Given an ADF, for each decision $d$, since $d$ either meets a decision
criterion or not, it is always possible to have a \flat{} explanation for $d$
being (strongly/weakly) dominant or not.

If $d$ is not weakly dominant, since for any decision $d$, the set of
goals met by $d$, $\dmg(d)$, is unique, then the \flat{} explanation as given
in Definition~\ref{dfn:explanation} is unique.

\paragraph{Proof of Proposition~\ref{prop:expPropertyPfrSet}}
Similar to the proof of Proposition~\ref{prop:expProperty}, since $d$
is either \pfrSet{} or not, $d$ always has a \flat{} explanation. Since
$\dmg(d)$ is unique, if $d$ is not \pfrSet{}, then it has a unique \flat{}
explanation. 

\paragraph{Proof of Theorem~\ref{SDABAThm}}
Let $AF = \lracF$.

({\bf Part I.}) We first show that if $d$ is strongly dominant, then
$\argu{\{sDom(d)\}}{sDom(d)}$ is admissible. This is to show: 
\begin{enumerate}
\item
$\argu{\{sDom(d)\}}{sDom(d)}$ is an argument;

\item
there exists $\Delta \subseteq \mc{A}$ such that $\{sDom(d)\} \subseteq
\Delta$ and $\Delta$ withstands all attacks;

\item
$\Delta$ 
does not attack itself.
\end{enumerate}

Since $sDom(d) \in \mc{A}$, $\argu{\{sDom(d)\}}{sDom(d)}$ is an
argument, and 1. holds trivially.

The contrary of $sDom(d)$ is $notSDom(d)$, and the only rules with
head $notSDom(d)$ are of the form  
$notSDom(d) \gets notMet(d,g), isD(d), isG(g)$ for some $g \in \goals$.
Hence attackers of $\argu{\{sDom(d)\}}{sDom(d)}$ are of the form 
$\argu{\{notMet(d,g)\}}{notSDom(d)}.$ Since $d$ is
strongly dominant, $\dmg(d) = \goals$. Hence, for every $g \in
\goals$, $d$ meets $g$ 
and $met(d,g)\gets\in \mc{R}$.
So, for any $g \in \goals$, we can construct
an argument $\argu{\{\}}{met(d,g)}$ and such argument can not
be attacked (as it is supported by the empty set). Hence, for all
$g$, $\argu{\{notMet(d,g)\}}{notSDom(d)}$ is counter-attacked and
2. also holds, with $\Delta=\{sDom(d)\}$.

Since 
no argument for $notSDom(d)$ can be obtained with support a subset of 
$\Delta$
, $\Delta$
does not attack itself, and 3. also holds. 


({\bf Part II.}) We then show that if $\argu{\{sDom(d)\}}{sDom(d)}$ is
admissible then $d$ is strongly dominant
,
namely it meets all goals. 

As shown in Part I, attackers 
against 
$\argu{\{sDom(d)\}}{sDom(d)}$ are 
of the form
$\argu{\{notMet(d,g)\}}{notSDom(d)}$, for all $g$. Since 
$\argu{\{sDom(d)\}}{sDom(d)}$ is admissible, it
withstands all attacks. Hence, 
since 
$\mc{C}(notMet(d,g)) = \{met(d,g)\}$, by definition of $\mc{R}$,
for all $g$, 
$\argu{\{\}}{met(d,g)}$ 
is an argument. Furthermore, because
$met(d,g) \gets$ is a fact, $d$ meets $g$, for any $g \in \goals$, and thus $d$ is
strongly dominant. \qed

\paragraph{Proof of Proposition~\ref{prop:SDExp}}
This is easy to see as the \flat{} explanation for $d$ needs to be
$\{\dmg(d)\} = \goals$ (Definition~\ref{dfn:explanation}). For each
$\pro$ node $\pro:\argu{\{\}}{met(d,g)}$ in $\atree$, since it does not
  have any \opp{} child, by Definition~\ref{dfn:sdABA}, we know that
  $d$ meets $g$. Since $\atree$ is admissible, meaning that for all $g
  \in \goals$, there exists a $\pro$ node $\pro:\argu{\{\}}{met(d,g)}$
  in $\atree$, we conclude that the set
  $\{g|\pro:\argu{\{\}}{met(d,g)}$ is a leaf node in $\atree\}$ must
  be $\goals$. 
  Note that in this case, the fact that $\atree$ is \leastAsm\ does not play a role; indeed, every admissible dispute tree with the same root as $\atree$ is necessarily \leastAsm. \qed

\paragraph{Proof of Proposition~\ref{prop:NSDExp}}
To show that $S = \{g|\opp:\argu{\{notMet(d,g)\}}{notSDom(d)}$ is a leaf
node in $\atree\}$ is the \flat{} explanation for $d$ not being strongly
dominant, we need to show that $S = \goals \setminus \dmg(d)$. Since
$\atree$ is best-effort, meaning that although $\atree$ is not
admissible, it contains as few \opp{} leaf nodes as possible
(Definition~\ref{dfn:LAandBE}). Thus, by Definition~\ref{dfn:sdABA},
we know that for each $g \in \goals$, if $d$ does not meet $g$,
$\opp:\argu{\{notMet(d,g)\}}{notSDom(d)}$ must be a leaf node in
$\atree$. Therefore, $S = \goals \setminus \dmg(d)$. \qed

\paragraph{Proof of Theorem~\ref{DABAThm}}
Let $AF = \lracF$.

({\bf Part I.}) To prove that dominance implies admissibility is to prove:
\begin{enumerate}
\item
$\argu{\{dom(d)\}}{dom(d)}$ is an argument;

\item
there exists $\Delta \subseteq \mc{A}$ such that $\{dom(d)\} \subseteq
\Delta$ and $\Delta$ withstands all attacks;

\item
$\Delta$ 
does not attack itself.
\end{enumerate}

Since $dom(d) \in \mc{A}$, $\argu{\{dom(d)\}}{dom(d)}$ is an
argument, and 1. holds trivially.

Let $\Delta = \{dom(d), noOthers(d,g)| g\in \goals, g \not\in \dmg(d) \}$.
Attacks against $\{dom(d)\}$ are
of the form $\argu{\{notMet(d,g)\}}{notDom(d)}$, for all $g \in \goals$. Since 
$d$ is dominant, $d$ meets all goals  met by any decision in
$\decisions$. Hence, for each goal $g \in \goals$, either
\begin{enumerate}
\item
 $d$ meets $g$, hence $\argu{\{\}}{met(d,g)}$ exists and is not
 attacked; or
\item
$d$ does not meet $g$, but
there is no argument $\argu{\{\}}{met(d',g)}$ for any $d' \in
\decisions$ (indicating that $g$ is not met by any $d'$);
therefore $\argu{\{noOthers(d,g)\}}{noOthers(d,g)}$ can not
be attacked.
\end{enumerate}
Whichever the case, 
$\argu{\{notMet(d,g)\}}{notDom(d)}$ is counter-attacked by $\Delta$ so 
$\argu{\{dom(d)\}}{dom(d)}$ withstands all attacks.
Also, since  arguments of the form $\argu{\{noOthers(d,g)\}}{noOthers(d,g)}$,
such that  $g$ is not met by $d$, cannot be attacked, 2. also holds. 

It is
easy to see $\Delta$ does not attack itself, and 3. also holds. 

({\bf Part II.}) We then show that admissibility implies
dominance. Since 
$dom(d)$ belongs to an admissible set,
all arguments 
$\argu{\{notMet(d,g)\}}{notDom(d)}$ are attacked for all $g \in
\goals$. Since the contrary of $notMet(d,g)$ is 
$met(d,g)$ or $noOthers(d,g)$, $\argu{\{dom(d)\}}{dom(d)}$
withstanding all of its attacks means, for each $g$, either
$met(d,g)$ or $noOthers(d,g)$ can be ``proved''. Hence, 
either $g$ is met by $d$ or there is no $d' \in \decisions$
meeting $g$ and, by Definition~\ref{dfn:d}, $d$ is dominant. \qed

\paragraph{Proof of Proposition~\ref{prop:DExp}}
Since $\atree$ is admissible and \leastAsm{}, for each node
$\opp:\argu{\{notMet(d,g\}}{notDom(d)}$ in $\atree$, it must have a
$\pro$ node child $\pro:\argu{\{\}}{met(d,g)}$. Jointly, they give the
collection of goals met by $d$. Therefore, the set
$\{g|\pro:\argu{\{\}}{met(d,g)}$ is a leaf node in $\atree\} =
\dmg(d)$.

Moreover, for those nodes $\opp:\argu{\{notMet(d,g\}}{notDom(d)}$
without a $\pro:\argu{\{\}}{met(d,g)}$ as its child, (since $\atree$
is admissible), each of them must have a child of the form
$\pro:\argu{\{noOthers(d,g)\}}{noOthers(d,g)}$, stating that although
$d$ does not meet $g$, there is no other decision $d'$ meets
$g$. Thus, $g$s in those $\pro$ nodes give nodes not met by $d$.

\paragraph{Proof of Proposition~\ref{prop:NDExp}}
Since $\atree$ is best effort, for each goal $g$ not met by $d$,
there must be a node
$\pro:\argu{\{noOthers(d,g)\}}{noOthers(d,g)}$ claiming
that there is no other decision $d'$ meets $g$. Since $\atree$ is
not admissible, there must be some $g$ in those nodes met by some
other decision $d'$. Nodes containing those $g$s thus have children
of the form $\opp:\argu{\{\}}{met(d',g)}$. Those $g$s are met by $d'$s
but not $d$. \qed

\paragraph{Proof of Theorem~\ref{WDABAThm}}
Let $AF = \lracF$.

({\bf Part I.}) To prove that weak dominance implies
admissibility 
is to prove: 
\begin{enumerate}
\item
$\argu{\{wDom(d)\}}{wDom(d)}$ is an argument;

\item
there exists $\Delta \subseteq \mc{A}$ such that $\{wDom(d)\} \subseteq
\Delta$ and $\Delta$ withstands all attacks;

\item
$\Delta$ 
does not attack itself.
\end{enumerate}

Since $wDom(d) \in \mc{A}$, $\argu{\{wDom(d)\}}{wDom(d)}$ is an 
argument, and 1. trivially holds. 

Let $\Delta=\{wDom(d), notMet(d',g)|d' \in \decisions, d'\neq d, g \in \goals,
g \in \dmg(d), g \not\in\dmg(d')\}$.
Attacks against $\{wDom(d)\}$ are of the form 
$\argu{\{notMet(d,g), notMore(d,d')\}}{notWDom(d)}$. Since $d$ is weakly dominant, 
then there is no $d'
$ \st\
$\dmg(d) \subset \dmg(d')$. Hence, given any $d'
\neq d$, for each $g \in \dmg(d')$, either 
\begin{enumerate}
\item
$g \in \dmg(d)$ or

\item
$g \not\in \dmg(d)$, but there exists some $g' \in \goals$ such that
  $g' \in \dmg(d)$ and $g' \not\in \dmg(d')$. 
\end{enumerate}

In the first case, $notMet(d,g)$ 
is attacked by $\argu{\{\}}{met(d,g)}$; in the second case, 
$notMore(d,d')$ 
there is some $g'$ met by 
 $d$ but not $d'$ (so $more(d,d')$ can be ``proved'' from $\Delta$). Hence,
whichever the case, $\argu{\{wDom(d)\}}{wDom(d)}$ withstands
attacks towards it. 
Also, 
no $notMet(d',g)\in\Delta$ can be attacked, 
and 2. also holds.

It is easy to see that 
$\Delta$ does not attack itself, and 3. also holds.

({\bf Part II.}) We then show that if
admissibility implies weak dominance. 

Since $\argu{\{wDom(d)\}}{wDom(d)}$ is admissible, all of its
attackers are counterattacked
. Hence, arguments for
$notWDom(d)$ are counter-attacked
. 
By definition of $\mc{R}$ in the $AF$, 
if $notWDom(d)$ is attacked, then, for any $g$ and $d'\neq d$:
\begin{enumerate}
\item
it is not the case that $met(d',g)$ and $notMet(d,g)$ both hold or
\item
there is some $g'$ such that $met(d,g')$ and $notMet(d',g')$ (so
$more(d,d')$ holds and $notMore(d,d')$ is attacked).
\end{enumerate}
Whichever the case, $\dmg(d)$ is not a subset of $\dmg(d')$. Therefore
$d$ is weakly dominant. \qed

\paragraph{Proof of Proposition~\ref{prop:WDExp}}
Since $\atree$ is \leastAsm{}, for each goal $g$ met by $d$, there
must be a $\pro$ node $\pro:\argu{\{\}}{met(d,g)}$ in a leaf of
$\atree$. Moreover, for each node
$\pro:\argu{\{notMet(\_,g)\}}{more(d,\_)}$ in $\atree$, from 
Definition~\ref{dfn:wdABA}, we can see that $d$ meets $g$. Jointly,
the two collections of goals $g$ in these nodes are met by $d$. 

For each leaf node $\pro:\argu{\{notMet(d',g)\}}{more(d,d')}$ in
$\atree$, as it has no $\opp$ child, we can see that $d'$ does not
meet $g$ but $d$ meets. Thus, the collection of these $(d',g)$ pairs
gives the set of goals met by $d$ but not $d'$s.

Overall, we can see that for each $d'$, let $g' \in \dmg(d')$, if $d$
meets $g'$, then $\pro:\argu{\{\}}{met(d,g')}$ would be a leaf node in
$\atree$; otherwise, $\pro:\argu{\{notMet(d',g)\}}{more(d,d')}$ would
be a leaf node in $\atree$ for some $g, g \in \dmg(d)$ and $g \not\in
\dmg(d')$.

\paragraph{Proof of Proposition~\ref{prop:NWDExp}}
An explanation for some decision $d$ not being weakly dominant is the
set of decisions meeting strictly more goals than $d$ (\wrt{}
$\subset$). As $\atree$ is best effort, 
all nodes $N$ of the form $\opp:\argu{\{notMet(d,g),
  notMore(d,d')\}}{notWDom(d)}$ have children
$\pro:\argu{\{\}}{met(d,g)}$, if $d$ meets $g$. Only for the goals
$g'$, which $d$ does not meet, arguments of the form
$\argu{\{notMet(d',g')\}}{more(d,d')}$ are used to label \pro{} nodes
as children of $N$. If these arguments cannot be attacked, then indeed
$d$ meets more goals than $d'$. However, since $d$ is not weakly
dominant, some of these arguments must be attacked by arguments of the
form $\argu{\{\}}{met(d',g')}$. In these cases, $d'$ meets strickly
more goals than $d$ ($g'$ is one such example). It is easy to see
that nodes of the form $\opp:\argu{\{\}}{met(d',g')}$ are leaf nodes
of $\atree$ so the collection of $d'$s in these nodes give the
explanation for $d$.

When $d$ does not meet any goal, no node of the form
$\pro:\argu{\{notMet(d',g')\}}{more(d,d')}$ can be added to $\atree$,
(to construct these arguments, $d$ has to meet $g'$). In this case,
leaf nodes of $\atree$ would be in the form of $\opp:\argu{\{notMet(d,g),
  notMore(d,d')\}}{notWDom(d)}$, which shows that $d'$ meets $g$ and
$d$ does not. Thus, the collection of $d'$ in these leaf nodes
also constitutes the explanation for $d$ not being weakly dominant. 

\paragraph{Proof of Theorem~\ref{thmTran2}}
The proof of this theorem is similar to the proofs of
Theorems~\ref{SDABAThm}, \ref{DABAThm} and \ref{WDABAThm}. 
Let $AF = \lracF$.

({\bf Part I.}) We first prove if $d$ is \pfrSet, then
$\argu{\{\pS(d)\}}{\pS(d)}$ is in an admissible extension. To
show $\argu{\{\pS(d)\}}{\pS(d)}$ is admissible, we need to show:
\begin{enumerate}
\item
$\argu{\{\pS(d)\}}{\pS(d)}$ is an argument.

\item
With help of the set of arguments $\Delta$,
$\argu{\{\pS(d)\}}{\pS(d)}$ withstands its attacks.

\item
$\{\argu{\{\pS(d)\}}{\pS(d)}\} \cup \Delta$ is conflict-free.
\end{enumerate}
Since $\pS(d)$ is an assumption, $\argu{\{\pS(d)\}}{\pS(d)}$
is an argument. 

To show $\argu{\{\pS(d)\}}{\pS(d)}$ withstands its attacks, we show its
attackers are defeated. Since $\mc{C}(\pS(d)) = \{\notPS(d)\}$,
attackers of $\argu{\{\pS(d)\}}{\pS(d)}$ are arguments with claim
$\notPS(d)$. Since rules with head $\notPS(d)$ are of the 
form 
\begin{center}
$\notPS(X) \gets \metSet(X_1,S), \notMetSet(X,S), notBetter(X,X_1,S);$
\end{center}
attackers of $\argu{\{\pS(d)\}}{\pS(d)}$ are arguments
of the form 
\begin{center}
$\argu{\{\metSet(d',s), notMet(d,g), notBetter(d,d',s)\}}{\notPS(d)}$
  (for some $g \in s$).
\end{center}
Hence we need to show for all $s \in \cgset, d' \in \decisions$,
$\argu{\{\pS(d)\}}{\pS(d)}$ withstands (with help) attacks from 
$\argu{\{\metSet(d',s),notSet(d,g),notBetter(d,d',s)\}}{\notPS(d)}$. 

Because $\metSet(d',s)$, $notMet(d,g)$ and $notBetter(d,d',s)$ are
assumptions, if there are arguments $\arg$ for contraries of
them and $\arg$ is either not attackted or withstands its attacks,
then we conclude that the argument
$\argu{\{\metSet(d',s),notMet(d,g),notBetter(d,d',s)\}}{notPG(d)}$ 
is defeated and $\argu{\{\pS(d)\}}{\pS(d)}$ holds. We show such $\arg$  
exists when $d$ is a \pfrSet.

For all $d' \in \decisions, d' \neq d$, for any $s \in \cgset$ there
are two possibilities:
\begin{enumerate}
\item
it is the case that $s \not\subseteq \dmg(d)$ and $s \subseteq
\dmg(d')$; and 

\item
it is not the case that $s \not\subseteq \dmg(d)$ and $s \subseteq
\dmg(d')$, i.e. one of the following holds:
\begin{enumerate}
\item
$s \not\subseteq \dmg(d)$ and $s \not\subseteq \dmg(d')$,

\item
$s \subseteq \dmg(d)$ and $s \subseteq \dmg(d')$,

\item
$s \subseteq \dmg(d)$ and $s \not\subseteq \dmg(d')$.
\end{enumerate}
\end{enumerate}

In case 1, since $d$ is \pfrSet, by Definition~\ref{decFuncPSet},
there exists $s' \in \cgset$, such that
\begin{center}
(1) $s' \geq s  \in \preferenceSet$, (2) $s' \subseteq \dmg(d)$, and
  (3) $s' \not\subseteq \dmg(d')$. 
\end{center}
From the three conditions above, we see that:
\begin{itemize}
\item
Since $s' \geq s$ in \preferences, $pfr(s',s) \gets$ is in
$\mc{R}$.
\item
Since $s' \subseteq \dmg(d)$, $\argu{\{\}}{met(d,g)}$ is an argument
for all $g \in s'$.
\item
Since $s' \not \subseteq \dmg(d')$, there is no argument for
$met(d',g')$ for some $g' \in s'$, hence the assumption
$notMet(d',g')$ is not attacked (the contrary of $notMet(d',g')$ is
$met(d',g')$).
\end{itemize}
Jointly, we know that there is an argument
$\argu{\{notMet(d',g')\}}{better(d,d',s)}$, and this argument is not
attacked. Therefore, since the contrary of $notBetter(d,d',s)$ is
$better(d,d',s)$, the argument $\argu{\{notMet(d,g),
  notBetter(d,d',s)\}}{notPG(d)}$, cannot withstand its attacks.

In case 2(a), $s \not\subseteq \dmg(d')$, there is some $g \in s$
such that $g \notin \dmg(d')$. There is an argument
$\argu{\{notMet(d',g)\}}{\notMetSet(d',s)}$ defeating the argument
$$\argu{\{\metSet(d',s), notMet(d,g'), notBetter(d,d',s)\}}{notPG(d)}.$$

In case 2(b) and 2(c), $s \subseteq \dmg(d)$, hence for each $g' \in
s$, arguments of the form $\argu{\{\}}{met(d,g')}$ can be constructed
and not attacked. Such arguments attack 
$$\argu{\{\metSet(d',s),notMet(d,g'),notBetter(d,d',s)\}}{notPG(d)}.$$

In case 1 or 2, $\argu{\{pG(d)\}}{pG(d)}$ withstands its attacks.

It is easy to see that $\{\argu{\{pG(d)\}}{pG(d)}\} \cup \Delta$ is
conflict-free by checking through all assumption and contrary
mappings. 

Since $\argu{\{\pS(d)\}}{\pS(d)}$ is an argument and, with help from
a conflict-free set of arguments, withstands all of its attacks,
$\argu{\{\pS(d)\}}{\pS(d)}$ is admissible.

({\bf Part II.}) We show: if $\argu{\{\pS(d)\}}{\pS(d)}$ belongs to
an admissible set of arguments, then $d$ is a \pfrSet. To show
$d$ is \pfrSet, we need to show for all $d' \in \decisions
\setminus \{d\}$, the following holds: 

\begin{list}{$\star$}{}  
\item
for all $s \in \cgset$, if $s \not \subseteq \dmg(d)$ and $s \subseteq
\dmg(d')$, then there exists $s' \in \cgset$ such that:
(1) $s' \geq s$ in $\preferences$, (2) $s' \subseteq \dmg(d)$, and (3)
  $s' \not\subseteq \dmg(d')$.
\end{list}

Since there is $\Delta$ such that $\argu{\{pG(d)\}}{pG(d)}
\cup \Delta$ is an admissible set,
\begin{enumerate}
\item
$\argu{\{\pS(d)\}}{\pS(d)}$ is an argument;

\item
with help of $\Delta$, $\argu{\{\pS(d)\}}{\pS(d)}$ withstands all
attacks towards it. 
\end{enumerate}

With the reasoning shown in Part I in reverse, we can see that
$\argu{\{\pS(d)\}}{\pS(d)}$ being admissible implies $d$ being \pfrSet.
\qed

\paragraph{Proof of Proposition~\ref{prop:expPfrSet}}
To show $\expPair{G}{M}$ is an explanation for $d$ being \pfrSet, we
first need to show that $G \subset \dmg(d)$. This is easy to see as
$G$ is the set of goals $g$s in leaf \pro{} nodes labelled by
$\argu{\{\}}{\{met(d,g)\}}$. By Definition~\ref{dfn:pfrGoalSet}, $d$
meets these $g$s. Secondly, for each pair $(S,d')$ \st{}
$N = \pro:\argu{\{\metSet(d,S),notMet(d',\_)\}}{better(d,d',\_)}$ is a 
leaf node in $\atree$, we know that $d$ meets $S$ and $d'$ does not
meet anything more preferred than $S$ (as if it does, $N$ would not be
a leaf node). \qed

\paragraph{Proof of Proposition~\ref{prop:expNPfrSet}}
The proof of this proposition is similar to the one for
Proposition~\ref{prop:NWDExp}. Since $\atree$ is best effort, all
arguments labelling leaf \opp{} nodes in $\atree$ capture decisions
meeting more preferred goals than $d$ with $g$ being a goal in such a
goal set. \qed

\paragraph{Proof of Proposition~\ref{prop:DGinstanceADF}}
Given a DG $G$=\edgTuple\ with  \node = $\dNode \cup \iNode \cup \gNode$,
$\df = \nadf$
with, let, for $n_1 \in \decisions$ and $n_2 \in \goals$,  
$n_2 \in \dmg(n_1)$
\ifaf{}  $n_2$ is reachable from $\{n_1\}$. Then,   
it is immediate to see that $\df$
is an equivalent ADF  to $G$, 
\st{} decision $d \in \decisions$ meets a goal $g \in \goals$ in
$\df$ \ifaf{} $d$ meets $g$ in $G$. \qed

\paragraph{Proof of Proposition~\ref{prop:ADFinstanceDG}}
Given an ADF
$\df = \nadf$,
we can create an equivalent DG $\dg =
\dgPair$ where $\node = \{n|n \in \decisions \cup \goals\}$
and $\edge = \{\egPair{n_1}{n_2}|
n_2 \in \dmg(n_1)
\}$. It is easy to
see that a decision $d \in \decisions$ meets a goal $g \in \goals$ in
$\dg$ \ifaf{} $d$ meets $g$ in $AF$. \qed

\paragraph{Proof of Proposition~\ref{prop:DGinstancePDF}}
The proof of this proposition is similiar to the one for
Proposition~\ref{prop:DGinstanceADF}. For any PDG $G = \pdgTuple$ with
$\node = \dNode \cup \iNode \cup \gNode$, we can have $F
= \neadfps$ \st{} for $n_1 \in \decisions$ and $n_2 \in \goals$, 
$n_2 \in \dmg(n_1)$ \ifaf{}  $n_2$ is reachable from $\{n_1\}$. This 
ensures that a decision $d$ meets a goal $g$ in $G$ if and only if $d$ 
meets $g$ in $F$.

\paragraph{Proof of Proposition~\ref{prop:SDABAGD}}
The proof of this proposition is similar to the proof of
Theorems~\ref{SDABAThm}, \ref{DABAThm} and \ref{WDABAThm}. The
difference is that upon examining $met(d,g)$ for some $d \in
\decisions$ and $g \in \goals$, instead of checking
$\dmg$ directly,
it uses the rule $$met(D,S) \gets reach(D,S)$$
which further invokes rules
$$reach(X,Y) \gets edge(X,Y,T)$$
$$reach(X,Y) \gets reach(X,Z), edge(Z,Y,T), \neg
unreachableSib(Z,Y,T,X)$$
$$unreachableSib(Z,Y,T,X) \gets edge(W,Y,T), \neg reach(X,W), W \neq
Z$$
$$edge(n_1,n_2,\myTag{e}) \gets dEdge(n_1,n_2,\myTag{e})$$
$$edge(n_1,n_2,\myTag{e}) \gets $$

\noindent
and assumptions $\neg unreachableSib(Z,Y,T,X)$, $reach(X,W)$ and their
contraries.
The defeasibility is captured by assumptions of
the form $dEdge(n_1,n_2,\myTag{e})$, which have contraries $\neg
dEdge(n_1,n_2,\myTag{e})$ that can be derived from rules generated
from defeasible conditions. Thus, the condition of a decision $d$
meeting a goal $g$ (both $d$ and $g$ are nodes in the graph) is now
subject to the test on defeasibility on the path from $d$ to $g$.
The structure of the rest of the proof remains unchanged
and the conclusions hold. 
\qed

\paragraph{Proof of Proposition~\ref{prop:pfrSetDGABA}}
As in the proof of Proposition~\ref{prop:SDABAGD}, the proof of this
proposition is similiar to the one for 
Theorem~\ref{thmTran2}. Since DG generalises ADF and allows richer
description on the decision-meeting-goal relation, the rule
$$met(d,g) \gets$$
in ABA frameworks corresponding to ADFs are replaced by
$$met(d,g) \gets reach(d,g)$$
and additional rules for proving $reach(d,g)$, e.g.,
$$reach(n_1,n_2) \gets edge(n_1,n_2,\myTag{e})$$ and
$$reach(n_1,n_2) \gets reach(n_1,n_3), edge(n_3,n_2,\myTag{e}),
\neg unreachableSib(n_3,n_2,\myTag{e},n_1).$$
However, these rules do not affect the preference relations between
sets of goals, i.e., the rules, assumptions and contraries described
in the \pfrSet{} component. Thus, with reasoning similiar to
Theorem~\ref{thmTran2}, this proposition holds. \qed

\paragraph{Proof of Proposition~\ref{prop:expDGS}}
To prove $S = \{g|\pro:\argu{\{\_\}}{met(d,g)}$ is a leaf node in
$\atree\}$ is an explanation for $d$ being strongly dominant we need
to show that $S = \dmg(d)$. This is easy to see as for each goal $g$
in $\dg$, by Definition~\ref{dfn:sdABADG}, there is an \opp{} node of
the form $\opp:\argu{\{notMet(d,g\}}{notSDom(d)}$ attacking the root
node $\pro:\argu{\{sDom(d)\}}{sDom(d)}$. These \opp{} nodes can only
be attacked by $\pro$ nodes of the form
$\pro:\argu{\{\_\}}{met(d,g)}$. Thus, if these $\pro$ nodes are leaf
nodes, it means that $d$ meets $g$ is undisputed. Since $\atree$ is
admissible, all $g$s are met by $d$. Thus, $S$ is an explanation for
$d$ being strongly dominant.

With the same reasoning, we see that for decisions that are not
strongly dominant, $\{g|N=\opp:\argu{\{notMet(d,g)\}}{notSDom(d)}$ is
a leaf node in $\atree$ or $N$ is 
ancestor of an \opp{} leaf
node$\}$ is an explanation as either $N$ is a leaf node or $N$ is
ancestor of an $\opp$ leaf node, it means that $g$ is not met by
$d$. The collection of all such goals $g$s is an explanation for $d$
not being strongly dominant. \qed

\paragraph{Proof of Proposition~\ref{prop:expDGD}}
The proof of this proposition is similiar to the ones for
Proposition~\ref{prop:DExp} and \ref{prop:NDExp}. The differences are:
\begin{itemize}
\item
  Since the rule $met(d,g) \gets$ in the ABA framework corresponding
  to ADF is now replaced by $met(d,g) \gets reach(d,g)$ and rules with
  head $reach(d,g)$, arguments $\argu{\{\}}{met(d,g)}$ labelling leaf
  nodes are replaced by $\argu{\_}{met(d,g)}$ for admissible dispute
  trees.

\item
  Moreover, since $met(d,g)$ in the ABA framework corresponding to DG
  can be claims for arguments supported by assumptions, when $d$ does
  not meet $g$, it might be the case that
  $\argu{\{notMet(d,g)\}}{notDom(d)}$ is no longer a leaf node in
  $\atree$. Thus, the condition for $d$ not meeting $g$ for explaining
  $d$ not being dominant is changed to there is a path from an $\opp$
  leaf node to $\opp:\argu{\{notMet(d,g)\}}{notDom(d)}$. \qed
\end{itemize}

\paragraph{Proof of Proposition~\ref{prop:expDGW}}
The proof of this proposition is similiar to the ones for
Proposition~\ref{prop:WDExp} and \ref{prop:NWDExp}. As in the proof
for Proposition~\ref{prop:expDGD}, $\argu{\{\}}{met(d,g)}$ are
replaced by $\argu{\_}{met(d,g)}$. Since $\argu{\_}{met(d,g)}$ are now
subject to possible attacks, in order to establish its validity, we
impose that nodes labelled by these arguments must not be 
ancestors of $\opp$ leaf nodes. \qed

\paragraph{Proof of Proposition~\ref{prop:expDGP}}
The proof of this proposition is similiar to the ones for
Proposition~\ref{prop:WDExp} and \ref{prop:NWDExp} with the same
modification shown in proofs for Proposition~\ref{prop:expDGD} and
\ref{prop:expDGW}. \qed

\section{ABA Frameworks for some of the Examples in Section~\ref{sec:dg}}
\label{app:ABA}
\begin{example}
  \label{exp:dgCore-app}

  \begin{figure}
\begin{scriptsize}
\centerline{
\xymatrix@1@=12pt@R=8pt{
                 &  ic \ar@/_/[ld] \ar[d]   &       & ritz \ar@/_/[ld] \ar[d] \ar@/^/[rd] \\
inSK \ar[d]      & 50 \ar@/_1pc/[rrdd]|-{1} & inPic & 200 \ar[dd]|-{2} & discount \ar@/^/[ldd]|-{2}\\
near \ar@/_/[rd] &                          &                    \\
                 & convenient              && cheap             
}
}
\end{scriptsize}
\caption{DG for Example~\ref{exp:dgCore-app}.}
\label{fig:decGraph-app}
  \end{figure}
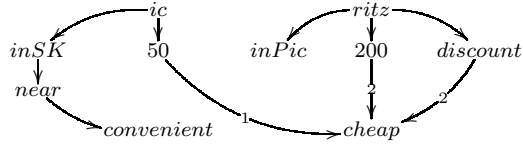

  Consider the DG without defeasible information 
	in Figure~\ref{fig:decGraph-app}.
  Let $uS$, $dt$ and $ct$ be short-hands for $unreachableSib$,
  $discount$ and $convenient$, \respectively, the core ABA framework
  corresponding to it is $\lracFcoreG$, where

  \begin{itemize}
  \item
    $\RcoreG$ consists of:

\noindent\begin{tabular}{lll}
$edge(ic,inSK,1) \gets$         & $edge(ic,50,1) \gets$         & $edge(ritz,inPic,1) \gets$ \\
$edge(ritz,dt,1) \gets$   & $edge(ritz,200,1) \gets$      & $edge(inSK,near,1) \gets$ \\
$edge(near,ct,1) \gets$ & $edge(50,cheap,1) \gets$      & $edge(200,cheap,2) \gets$ \\
$edge(dt,cheap,2) \gets$
\end{tabular}

\noindent\begin{tabular}{ll}
$reach(ic,inSK) \gets edge(ic,inSK,1)$       & $reach(ic,50) \gets edge(ic,50,1)$ \\
$reach(ritz,inPic) \gets edge(ritz,inPic,1)$ & $reach(ritz,200) \gets edge(ritz,200,1)$ \\
$reach(inSK,near) \gets edge(inSK,near,1)$   & $reach(50,cheap) \gets edge(50,cheap,1)$ \\
$reach(200,cheap) \gets edge(200,cheap,2)$  \\
\end{tabular}

\noindent\begin{tabular}{l}
$reach(near,ct) \gets edge(near,ct,1)$ \\
$reach(ritz,dt) \gets edge(ritz,dt,1)$ \\
$reach(dt,cheap) \gets edge(dt,cheap,2)$ \\
$reach(ic,near) \gets reach(ic,inSK),edge(inSK,near,1),\neg uS(inSK,near,1,ic)$\\
$reach(ic,ct) \gets reach(ic,near), edge(near,ct,1), \neg uS(near,ct,1,ic)$\\
$reach(ic,cheap) \gets reach(ic,50), edge(50,cheap,1), \neg uS(50,cheap,1,ic)$ \\
$reach(ritz,cheap) \gets reach(ritz,200), edge(200,cheap,2), \neg uS(200,cheap,2,ritz)$\\
$reach(ritz,cheap) \gets reach(ritz,dt), edge(dt,cheap,2), \neg uS(dt,cheap,2,ritz)$ \\
$uS(dt,cheap,2,ritz) \gets edge(200,cheap,2), \neg
reach(ritz, 200)$ \\
$uS(200,cheap,2,ritz) \gets edge(dt,cheap,2), \neg
reach(ritz, dt)$\\
$\{met(n_1,n_2) \gets reach(n_1,n_2) | n_1 \in \{ic, ritz\}, n_2 \in \{ct, cheap\}\}$.
\end{tabular}

\item
  $\AcoreG$ consists of:

  \noindent\begin{tabular}{lll}
  $\neg uS(inSK,near,1,ic)$ & $\neg uS(near,ct,1,ic)$ & $\neg uS(50,cheap,1,ic)$ \\
  $\neg uS(200,cheap,2,ritz)$ & $\neg uS(dt,cheap,2,ritz)$ & $\neg reach(ritz, 200)$ \\
  $\neg reach(ritz, dt)$ & \multicolumn{2}{l}{$\{notMet(d,g) | d \in \{ic, ritz\}, g \in \{ct, cheap\}\}$} \\
  \end{tabular}
  








  

  
\item
  Let $AN = \{ic, ritz, ct, cheap, inSK, near, 50, inPic, 200,
  dt\}$.

  For $n_1, n_2, n_3 \in AN$, $\mc{C}_g(\neg
  uS(n_3,n_2,t,n_1)) = \{uS(n_3,n_2,t,n_1)\}$;

  For $n_1, n_2 \in AN$, $\mc{C}_g(\neg reach(n_1,n_2)) =
  \{reach(n_1,n_2)\}$; 

  For $d \in \{ic, ritz\}$, $g \in \{ct, cheap\}$,
  $\mc{C}_g(notMet(d,g))=\{met(d,g)\}$.
  \end{itemize}

\end{example}

\begin{example}
  \label{exp:dgCoreD-app}
Consider the DG (with defeasible information) in
Figure~\ref{fig:expDGW}.
The core ABA framework corresponding to it is as shown
in Example~\ref{exp:dgCore-app} with the following modifications.

\begin{center}
\begin{tabular}{ll}
  $edge(ic,inSK,1) \gets$ & $edge(ic,50,1) \gets$
\end{tabular}
\end{center}

\noindent
replaced by

\begin{center}
\begin{tabular}{ll}
  $termTime \gets$ & $\neg dEdge(ic,50,1) \gets termTime$.
\end{tabular}
\end{center}

\noindent
$edge(ic,inSK,1)$ and $edge(ic,50,1)$ added to $\mc{A}_g$, \st{}
$\mc{C}_g(edge(ic,inSK,1))$ $= \neg edge(ic,inSK,1)$ and
$\mc{C}_g(edge(ic,50,1))$ $= \neg edge(ic,50,1)$, respectively.

\end{example}

\paragraph{ABA framework for Example~\ref{exp:expDGS}.}

The ABA framework 
	for this example is \lracF{} 
	where
\begin{itemize}
  \item
    \mc{R} consists of:
    
    \noindent\begin{tabular}{lll}
    $edge(d_1,g_1,1) \gets$ & $edge(d_2,a_1,1) \gets$ & $edge(d_2,a_2,1) \gets$ \\
    $edge(a_1,g_1,2) \gets$ & $edge(a_1,g_2,1) \gets$ & $edge(a_2,g_2,1) \gets$ \\
    \end{tabular}

    \noindent\begin{tabular}{lll}
    $reach(d_1,g_1) \gets edge(d_1,g_1,1)$ & $reach(d_2,a_1) \gets edge(d_2,a_1,1)$ \\
    $reach(d_2,a_2) \gets edge(d_2,a_2,1)$ & $reach(a_1,g_1) \gets edge(a_1,g_1,1)$ \\
    $reach(a_1,g_2) \gets edge(a_1,g_2,1)$ & $reach(a_2,g_2) \gets edge(a_2,g_2,1)$ \\
    \end{tabular}

    \noindent\begin{tabular}{l}
    $reach(d_1,g_2) \gets reach(d_1,a_1), edge(a_1,g_2,1), \neg unreachableSib(a_1,g_2,1,d_1)$ \\
    $reach(d_1,g_2) \gets reach(d_1,a_2), edge(a_2,g_2,1), \neg unreachableSib(a_2,g_2,1,d_1)$ \\
    $reach(d_2,g_1) \gets reach(d_2,a_1), edge(a_1,g_1,2), \neg unreachableSib(a_1,g_1,2,d_2)$ \\
    $reach(d_2,g_2) \gets reach(d_2,a_1), edge(a_1,g_2,1), \neg unreachableSib(a_1,g_2,1,d_2)$ \\
    $reach(d_2,g_2) \gets reach(d_2,a_2), edge(a_2,g_2,1), \neg unreachableSib(a_2,g_2,1,d_2)$ \\
    \end{tabular}
    
    \noindent\begin{tabular}{l}
    $unreachableSib(a_1,g_2,1,d_1) \gets edge(a_2,g_2,1), \neg reach(d_1,a_2)$ \\
    $unreachableSib(a_2,g_2,1,d_1) \gets edge(a_1,g_2,1), \neg reach(d_1,a_1)$ \\
    $unreachableSib(a_1,g_2,1,d_2) \gets edge(a_2,g_2,1), \neg reach(d_2,a_2)$ \\
    $unreachableSib(a_2,g_2,1,d_2) \gets edge(a_1,g_2,1), \neg reach(d_2,a_1)$ \\
    \end{tabular}

    \noindent\begin{tabular}{ll}
    $met(d_1,g_1) \gets reach(d_1,g_1)$ & $met(d_1,g_2) \gets reach(d_1, g_2)$ \\
    $met(d_2,g_1) \gets reach(d_2,g_1)$ & $met(d_2,g_2) \gets reach(d_2, g_2)$ \\
    \end{tabular}

    \noindent\begin{tabular}{ll}
    $notSDom(d_1) \gets notMet(d_1,g_1)$ & $notSDom(d_1) \gets notMet(d_1,g_2)$ \\
    $notSDom(d_2) \gets notMet(d_2,g_1)$ & $notSDom(d_2) \gets notMet(d_2,g_2)$ \\
    \end{tabular}

  \item
    $\mc{A}$ consists of: 

    \noindent\begin{tabular}{ll}
    $\neg unreachableSib(a_1,g_2,1,d_1)$ & $\neg unreachableSib(a_2,g_2,1,d_1)$ \\
    $\neg unreachableSib(a_1,g_1,2,d_2)$ & $\neg unreachableSib(a_1,g_2,1,d_2)$ \\
    $\neg unreachableSib(a_2,g_2,1,d_2)$ \\
    \end{tabular}
    
    \noindent\begin{tabular}{lllll}
    $\neg reach(d_1,a_2)$ & $\neg reach(d_1,a_1)$ & $\neg reach(d_2,a_2)$ & $\neg reach(d_2,a_1)$ & $sDom(d_1)$\\
    $notMet(d_1,g_1)$ & $notMet(d_1,g_2)$ & $notMet(d_2,g_1)$ & $notMet(d_2,g_2)$ & $sDom(d_2)$ \\
    \end{tabular}

  \item
    Let $AN = \{d_1, d_2, a_1, a_2, g_1, g_2\}$.

    For $n_1, n_2, n_3 \in AN$, $\mc{C}(\neg
    unreachableSib(n_3,n_2,t,n_1)) = \{unreachableSib(n_3,n_2,t,n_1)\}$.
    
    For $n_1, n_2 \in AN$, $\mc{C}(\neg reach(n_1,n_2)) = \{reach(n_1,n_2)\}$.

    For $d \in \{d_1,d_2\}$, $g \in \{g_1, g_2\}$, $\mc{C}(notMet(d,g))=\{met(d,g)\}$.

    For $d \in \{d_1,d_2\}$, $\mc{C}(sDom(d)) = \{notSDom(d)\}$.
\end{itemize}

\paragraph{ABA framework for Example~\ref{exp:expDGD}.}

The ABA framework for this example 
	is \lracF{}
where

\begin{itemize}
\item
  \mc{R} is all rules in the ABA framework for Example~\ref{exp:expDGS} with

  \noindent\begin{tabular}{ll}
  $notSDom(d_1) \gets notMet(d_1,g_1)$&$notSDom(d_1) \gets notMet(d_1,g_2)$\\
  $notSDom(d_2) \gets notMet(d_2,g_1)$&$notSDom(d_2) \gets notMet(d_2,g_2)$\\
  \end{tabular}
  
  replaced by
  
  \noindent\begin{tabular}{ll}
  $notDom(d_1) \gets notMet(d_1,g_1)$ & $notDom(d_1) \gets notMet(d_1,g_2)$\\
  $notDom(d_2) \gets notMet(d_2,g_1)$ & $notDom(d_2) \gets notMet(d_2,g_2)$\\
  $met(d_1,g_3) \gets reach(d_1,g_3)$ & $met(d_2,g_3) \gets reach(d_2,g_3)$\\
  \end{tabular}

\item
  \mc{A} is all assumptions in the ABA framework for Example~\ref{exp:expDGS} with

  \noindent\begin{tabular}{ll}
  $sDom(d_1)$ & $sDom(d_2)$ \\
  \end{tabular}

  replaced by

  \noindent\begin{tabular}{llll}
  $dom(d_1)$ & $dom(d_2)$ & $notMet(d_1,g_3)$   & $notMet(d_2,g_3)$ \\
  $noOthers(d_1,g_1)$     & $noOthers(d_1,g_2)$ & $noOthers(d_1,g_3)$ \\
  $noOthers(d_2,g_1)$     & $noOthers(d_2,g_2)$ & $noOthers(d_2,g_3)$ \\  
  \end{tabular}
  
\item
  For any \asm{} in $\mc{A}$, $\mc{C}(\asm)$ is as defined in
the ABA framework for   Example~\ref{exp:expDGS} along with

  \noindent\begin{tabular}{ll}
  $\mc{C}(dom(d_1)) = \{notDom(d_1)\}$ & $\mc{C}(dom(d_2)) = \{notDom(d_2)\}$ \\
  $\mc{C}(noOthers(d_1,g_1)) = \{met(d_2,g_1)\}$ & $\mc{C}(noOthers(d_1,g_2)) = \{met(d_2,g_2)\}$ \\
  $\mc{C}(noOthers(d_1,g_3)) = \{met(d_2,g_3)\}$ & $\mc{C}(noOthers(d_2,g_1)) = \{met(d_1,g_1)\}$ \\
  $\mc{C}(noOthers(d_2,g_2)) = \{met(d_1,g_2)\}$ & $\mc{C}(noOthers(d_2,g_3)) = \{met(d_1,g_3)\}$ \\
  \end{tabular}

  \hspace{5pt}For $d \in \{d_1,d_2\}, g \in \{g_1,g_2,g_3\}$,
  $\mc{C}(notMet(d,g)) = \{met(d,g), noOthers(d,g)\}$.
\end{itemize}

\paragraph{ABA framework for Example~\ref{exp:expDGW}.}

The ABA framework for this example 
	is
	$\lracF$, where

  \begin{itemize}
  \item
    \mc{R} is all rules in the ABA framework for Example~\ref{exp:expDGS} with 
    
    \noindent\begin{tabular}{ll}
    $notSDom(d_1) \gets notMet(d_1,g_1)$&$notSDom(d_1) \gets notMet(d_1,g_2)$\\
    $notSDom(d_2) \gets notMet(d_2,g_1)$&$notSDom(d_2) \gets notMet(d_2,g_2)$\\
    \end{tabular}

    replaced by

    \noindent\begin{tabular}{ll}
    $edge(d_3,a_3,1) \gets$ & $edge(a_3,g_3,1) \gets$ \\
    $reach(d_3,a_3) \gets edge(d_3,a_3,1)$ & $reach(a_3,g_3) \gets edge(a_3,g_3,1)$ \\
    $met(d_1,g_3) \gets reach(d_1, g_3)$ & $met(d_2,g_3) \gets reach(d_2, g_3)$ \\
    $met(d_3,g_1) \gets reach(d_3, g_1)$ & $met(d_3,g_2) \gets reach(d_3, g_2)$ \\
    $met(d_1,g_3) \gets reach(d_1, g_3)$ \\
    \end{tabular}

    \hspace{5pt}\vspace{-5pt}$reach(d_1,g_3) \gets reach(d_1,a_3), edge(a_3,g_3,1),
    \neg unreachableSib(a_3,g_3,1,d_1)$
    
    \hspace{5pt}\vspace{-5pt}$reach(d_2,g_3) \gets reach(d_2,a_3), edge(a_3,g_3,1),
    \neg unreachableSib(a_3,g_3,1,d_2)$
    
    \hspace{5pt}$reach(d_3,g_3) \gets reach(d_3,a_3), edge(a_3,g_3,1),
    \neg unreachableSib(a_3,g_3,1,d_3)$
    
    \hspace{5pt}For $d,d' \in \{d_1,d_2,d_3\}, d \neq d'$, $g,g' \in \{g_1,g_2,g_3\}$, 

    \hspace{15pt}$notWDom(d) \gets met(d',g), notMet(d,g), notMore(d,d')$,

    \hspace{15pt}$more(d,d') \gets met(d,g), notMet(d',g)$.

  \item
    \mc{A} is all assumptions in the ABA framework for Example~\ref{exp:expDGS} with

  \noindent\begin{tabular}{ll}
  $sDom(d_1)$ & $sDom(d_2)$ \\
  \end{tabular}

  replaced by

  \noindent\begin{tabular}{llll}
  $wDom(d_1)$ & $wDom(d_2)$ & $wDom(d_3)$ & $notMet(d_1,g_3)$\\
  $notMet(d_2,g_3)$  & $notMet(d_3,g_1)$ & $notMet(d_3,g_2)$ & $notMet(d_3,g_3)$\\
  $notMore(d_1,d_2)$ & $notMore(d_1,d_3)$ & $notMore(d_2,d_1)$ & $notMore(d_2,d_3)$ \\
  $notMore(d_3,d_1)$ & $notMore(d_3,d_2)$ & \multicolumn{2}{l}{$\neg unreachableSib(a_3,g_3,1,d_1)$}\\
  \multicolumn{2}{l}{$\neg unreachableSib(a_3,g_3,1,d_2)$} & \multicolumn{2}{l}{$\neg unreachableSib(a_3,g_3,1,d_3)$} \\
  \end{tabular}
  
\item
  For any \asm{} in $\mc{A}$, $\mc{C}(\asm)$ is as defined in the ABA framework for
  Example~\ref{exp:expDGS} along with

  \noindent\begin{tabular}{ll}
  $\mc{C}(wDom(d_1))=\{notWDom(d_1)\}$ & $\mc{C}(wDom(d_2))=\{notWDom(d_2)\}$ \\
  $\mc{C}(wDom(d_3))=\{notWDom(d_3)\}$ 
  \end{tabular}

  \hspace{5pt}For $d \in \{d_1,d_2,d_3\}$, $g \in \{g_1,g_2,g_3\}$,
  $\mc{C}(notMet(d,g)) = \{met(d,g)\}$

  \hspace{5pt}For $d,d' \in \{d_1,d_2,d_3\}, d \neq d'$,
  $\mc{C}(notMore(d,d')) = \{more(d,d')\}$

  \hspace{5pt}For $d \in \{d_1,d_2,d_3\}$,
  $\mc{C}(\neg unreachableSib(a_3,g_3,1,d)) = \{unreachableSib(a_3,g_3,1,d)\}$

  \end{itemize}

  \paragraph{ABA framework for Example~\ref{exp:expDGP}.}

The ABA framework for this example 
  $\lracF$ where

  \begin{itemize}
  \item
    \mc{R} is all rules in the ABA framework for Example~\ref{exp:expDGW} with

    for all $d,d' \in \{d_1,d_2,d_3\}, d \neq d'$, $g
    \in \{g_1,g_2,g_3\}$
    
    \noindent\begin{tabular}{l}
    $notWDom(d) \gets met(d',g), notMet(d,g), notMore(d,d')$ \\
    $more(d,d') \gets met(d,g), notMet(d',g)$\\
    \end{tabular}
    
    replaced by

    \begin{itemize}
    \item    
      \begin{tabular}{l}       
        $pfr(s3, s12) \gets$
      \end{tabular}
      
    \item
      for all $d \in \{d_1,d_2,d_3\}$,
 
      \begin{tabular}{ll}    
        $\notMetSet(d,s3) \gets notMet(d,g_3)$ &
        $\notMetSet(d,s12) \gets notMet(d,g_1)$ \\
        $\notMetSet(d,s12) \gets notMet(d,g_2)$ \\
      \end{tabular}
    
    \item
      for all $d,d' \in \{d_1,d_2,d_3\}, d\neq d', s,s' \in
      \{s3,s12\}, s\neq s'$,
      
      \begin{tabular}{l}
        $better(d,d',s) \gets \metSet(d,s'), \notMetSet(d',s'), pfr(s',s)$\\
        $\notPS(d) \gets \metSet(d',s), \notMetSet(d,s),notBetter(d,d',s)$\\
      \end{tabular}
    \end{itemize}

  \item
    \mc{A} is all assumptions in the ABA framework for Example~\ref{exp:expDGW} with for $d,d' \in
    \{d_1,d_2,d_3\}, d \neq d'$ 

      \noindent\begin{tabular}{ll}
      $wDom(d)$ & $notMore(d,d')$ 
      \end{tabular}

      replaced by 
          
      \noindent\begin{tabular}{lllll}
      $notBetter(d,d',s3)$ & $notBetter(d,d',s12)$ &
      $\metSet(d,s3)$ & $\metSet(d,s12)$ & $\pS(d)$
      \end{tabular}

    \item
      For all $\asm \in \mc{A}$, $\mc{C}(\asm)$ is as given in
		  the ABA framework for 
      Example~\ref{exp:expDGW}, along with
      \begin{itemize}
      \item
        for $d \in \{d_1,d_2,d_3\}, s \in \{qs3, q1q2S\}$,

        \noindent\begin{tabular}{ll}        
        $\mc{C}(\pS(d))= \{\notPS(d)\}$ &
        $\mc{C}(\metSet(d,s))=\{\notMetSet(d,s)\}$\\  
        \end{tabular}

      \item
        for $d,d' \in \{d_1,d_2,d_3\}, d \neq d'$

        \noindent\begin{tabular}{ll}
        $\mc{C}(notBetter(d,d',s3))=\{better(d,d',s3)\}$ \\
        $\mc{C}(notBetter(d,d',s12))=\{better(d,d',s12)\}$
        \end{tabular}        
      \end{itemize}    
  \end{itemize}

\end{document}